%% file: paper.tex
\documentclass[sigconf]{acmart}
\usepackage{amsmath,amsfonts,dsfont,multirow,multicol,epsfig,url,array,makecell,balance,color,epstopdf}
\usepackage{algorithmic}
\usepackage[ruled, vlined, linesnumbered]{algorithm2e}
\usepackage{hhline}
\usepackage{array}
\usepackage{enumerate}
\usepackage{enumitem}
\usepackage{subfigure}
\usepackage{booktabs}
\usepackage{xcolor,colortbl}
\usepackage{bm}
\usepackage{lipsum}
\usepackage{tabularx}
\usepackage{diagbox}
\setlist[itemize]{leftmargin=*}

\input{symbol-def.tex}
\AtBeginDocument{%
  \providecommand\BibTeX{{%
    \normalfont B\kern-0.5em{\scshape i\kern-0.25em b}\kern-0.8em\TeX}}}

\setcopyright{acmcopyright}
\copyrightyear{2022}
\acmYear{2022}

\acmConference[KDD '22] {Proceedings of the 28th ACM SIGKDD Conference on Knowledge Discovery and Data Mining}{August 14--18, 2022}{Madrid, Spain.}
\acmBooktitle{Proceedings of the 28th ACM SIGKDD Conference on Knowledge Discovery and Data Mining (KDD '22), August 14--18, 2022, Madrid, Spain}
\acmPrice{15.00}
\acmISBN{978-1-4503-9385-0/22/08}
\acmDOI{10.1145/XXXXXX.XXXXXX}

\newcommand{\liu}[1]{\textcolor{blue}{[LIU: #1]}}
\newcommand{\eat}[1]{}
\newtheorem{proposition}{Proposition}
\newtheorem{invariant}{Invariant}

\begin{document}

\title{Instant Graph Neural Networks for Dynamic Graphs}
\subtitle{[technical report]}

\author{Yanping Zheng}
\affiliation{%
  \institution{Renmin University of China}
  \city{Beijing}
  \country{China}
}
\email{zhengyanping@ruc.edu.cn}

\author{Hanzhi Wang}
\affiliation{%
  \institution{Renmin University of China}
  \city{Beijing}
  \country{China}
}
\email{hanzhi\_wang@ruc.edu.cn}

\author{Zhewei Wei}
\authornote{Zhewei Wei is the corresponding author. The work was partially done at Gaoling School of Artificial Intelligence, Peng Cheng Laboratory, Beijing Key Laboratory of Big Data Management and Analysis Methods and MOE Key Lab of Data Engineering and Knowledge Engineering.}
\affiliation{%
  \institution{Renmin University of China}
  \city{Beijing}
  \country{China}
}
\email{zhewei@ruc.edu.cn}

\author{Jiajun Liu}
\affiliation{%
  \institution{Data 61, CSIRO}
  \city{Queensland}
  \country{Australia}
}
\email{jiajun.liu@csiro.au}

\author{Sibo Wang}
\affiliation{%
  \institution{The Chinese University of Hong Kong}
  \city{Hong Kong}
  \country{China}
}
\email{swang@se.cuhk.edu.hk}

\input{abstract}

%
%




\setcopyright{none}
\pagestyle{plain}

\maketitle

\input{intro}
\input{pre}

\input{related_work}
\input{algorithm}
\input{experiments}
\input{conclusion}

\allowdisplaybreaks[4]

\begin{acks}
This research was supported in part by Beijing Natural Science Foundation (No. 4222028), by National Natural Science Foundation of China (No. 61972401, No. 61932001, No. 61832017, No. U1936205), by the major key project of PCL (PCL2021A12), by Beijing Outstanding Young Scientist Program No. BJJWZYJH012019100020098, by Alibaba Group through Alibaba Innovative Research Program, by CCF-Baidu Open Fund (No.2021PP15002000), by China Unicom Innovation Ecological Cooperation Plan and by the Huawei-Renmin University joint program on Information Retrieval. Additionally, Jiajun Liu was supported in part by CSIRO's Science Leader project R-91559. Sibo Wang was supported by Hong Kong RGC ECS (No. 24203419), Hong Kong RGC CRF (No. C4158-20G), Hong Kong ITC ITF (No. MRP/071/20X), CUHK Direct Grant (No. 4055181). We also wish to acknowledge the support provided by Intelligent Social Governance Interdisciplinary Platform, Major Innovation \& Planning Interdisciplinary Platform for the “Double-First Class” Initiative, Public Policy and Decision-making Research Lab, Public Computing Cloud, Renmin University of China.
\end{acks}

\begin{small}
\bibliographystyle{plain}
\bibliography{paper}
\end{small}

\appendix

\input{batch_update}
\input{error_analysis}
\input{details_exp}

\input{details_exp_more}
\input{proof}

\end{document}

%% file: symbol-def.tex
\newtheorem{lemma}{Lemma}
\newtheorem{theorem}{Theorem}

\def\r{\bm{r}}

\def\Z{\mathbf{Z}}

\def\I{\mathbf{I}}
\def\P{\mathbf{P}}
\def\A{\mathbf{A}}
\def\D{\mathbf{D}}

\def\Q{\mathbf{Q}}

\def\Y{\mathbf{Y}}
\def\X{\mathbf{X}}

\def\instgnn{{\em InstantGNN}\xspace}

\definecolor{darkpurple}{RGB}{102,0,255} 

%% file: abstract.tex
\begin{abstract}
\label{abstract}
Graph Neural Networks (GNNs) have been widely used for modeling graph-structured data. With the development of numerous GNN variants, recent years have witnessed groundbreaking results in improving the scalability of GNNs to work on static graphs with millions of nodes. \eat{However, how the \liu{continuous changes} in large-scale dynamic graphs can be instantly represented by GNNs remains an open problem. }However, how to instantly represent continuous changes of large-scale dynamic graphs with GNNs is still an open problem. Existing dynamic GNNs focus on modeling the periodic evolution of graphs, often on a snapshot basis. Such methods suffer from two drawbacks: first, there is a substantial delay for the changes in the graph to be reflected in the graph representations, resulting in losses on the model's accuracy; second, repeatedly calculating the representation matrix on the entire graph in each snapshot is predominantly time-consuming and severely limits the scalability. In this paper, we propose Instant Graph Neural Network (\instgnn), an incremental computation approach for the graph representation matrix of dynamic graphs. Set to work with dynamic graphs with the edge-arrival model, our method avoids time-consuming, repetitive computations and allows instant updates on the representation and instant predictions. Graphs with dynamic structures and dynamic attributes are both supported. The upper bounds of time complexity of those updates are also provided. Furthermore, our method provides an adaptive training strategy, which guides the model to retrain at moments when it can make the greatest performance gains. We conduct extensive experiments on several real-world and synthetic datasets. Empirical results demonstrate that our model achieves state-of-the-art accuracy while having orders-of-magnitude higher efficiency than existing methods.

\end{abstract}


%% file: intro.tex
\vspace{-2mm}
\section{Introduction} 
\label{sec:intro}
Graph Neural Networks (GNNs) are at the frontier of graph learning research. They have been widely employed in various applications, such as recommendation systems~\cite{fan2019social_recommendation}, molecules synthesis and manipulation~\cite{you2018molecular}, etc. Traditional GNN algorithms, with their un-optimized messaging techniques and storage mechanics, struggle to handle large-scale graph data. Numerous methods~\cite{wu2019simplifying,wang2021agp,chen2020scalable,bojchevski2020pprgo,zeng2019graphsaint,hamilton2017inductive} have been proposed to improve the scalability of GNNs and optimize the efficiency in processing large-scale graphs, by exploiting the computational characteristics of GNNs. These algorithms are designed for static graphs and use static computation strategies. However, graphs in the real-world generally change their graph structure and properties over time.\eat{ The evolutionary information is proved to be helpful in improving GNNs’ performance in some graph analysis tasks such as traffic forecast. Therefore,} Several dynamic Graph Neural Networks~\cite{panagopoulos2021mpnnlstm,zhao2019tgcn,pareja2020evolvegcn,zhou2018DynamicTriad,goyal2020dyngraph2vec} have been proposed in recent years to model the temporal information from graph evolution. Nevertheless, they focus on the prediction of the properties of nodes in the future and are limited in scalability. How to instantaneously compute the representation matrix on large-scale dynamic graphs remains an open problem.

\noindent \textbf{Motivation.} A GNN's working mechanism can be characterized by two operations: propagation and prediction. The former digests the message passing while the latter learns the weighted matrix. After a series of GNNs~\cite{kipf2016semi,velivckovic2017graph}  with both operations combined and being inseparable, SGC~\cite{wu2019simplifying} and PPNP\hspace{-0.5mm}~\cite{klicpera2018predict} signaled the finding that desirable properties could be introduced by separating the propagation process from the prediction process. By convention, a well-designed message propagation mechanism can be substantially beneficial to GNN models~\cite{zhu2021interpreting}. 

Let $\Z$ be the graph representation matrix, a generalized propagation equation which has been used in SGC~\cite{wu2019simplifying}, AGP~\cite{wang2021agp} and GBP~\cite{chen2020scalable} can be formulated as:
\vspace{-1mm}
\begin{equation}
\label{equ:eq1}
    \Z=\sum\limits_{\ell=0}^{\infty} \alpha (1-\alpha)^\ell\P^\ell \X =\sum\limits_{\ell=0}^{\infty} \alpha (1-\alpha)^\ell\hspace{-0.5mm}\left( \D^{-a}\A\D^{-b} \right)^\ell \X,
\vspace{-1mm}
\end{equation}
where $\X$ denotes the feature matrix of the input graph, $\A$ and $\D$ represent the adjacent matrix and the degree matrix, respectively. The propagation matrix $\P$, defined as $\D^{-a}\A\D^{-b}$, can easily generalize existing methods, where $a$ and $b$ are Laplacian parameters ranging from 0 to 1. By setting $a\hspace{-0.5mm}=\hspace{-0.5mm}b\hspace{-0.5mm}=\hspace{-0.5mm}1/2$, $\P$ represents the symmetric normalization adjacency matrix used in~\cite{kipf2016semi, klicpera2018predict, wu2019simplifying}. Also, the transition probability matrix $\A\D^{-1}$~\cite{chiang2019cluster,zeng2019graphsaint} and the reverse  transition probability matrix $\D^{-1}\A$~\cite{xu2018jknet} are obtained by setting $a\hspace{-0.5mm}=\hspace{-0.5mm}0,b\hspace{-0.5mm}=\hspace{-0.5mm}1$ and $a\hspace{-0.5mm}=\hspace{-0.5mm}1,b\hspace{-0.5mm}=\hspace{-0.5mm}0$, respectively. After deriving $\Z$, we can utilize Multi-Layer Perceptrons (MLPs) to make the prediction with mini-batch training. Therefore, a GNN model for the multi-class classification task can be written as $\Y=softmax(MLP(\Z))$. 

It is known that the calculation of $\Z$ is computationally expensive. To handle dynamic graphs, static or snapshot-based GNN models have to recompute to obtain the updated representation matrix $\Z$ periodically, which increases computational complexity dramatically. A natural question worth investigating is ``\emph{can we reuse results from previous computations to accelerate the calculation process for the current moment?}''. This paper aims to propose a dynamic GNN algorithm that incrementally updates the representation matrix in Equation~\ref{equ:eq1} as the graph changes over time, with necessary updates on affected nodes only, avoiding the time-consuming computation on the entire graph.

\noindent \textbf{Contributions.} In this paper, we present InstantGNN, an efficient approximate GNN on dynamic graphs. The intuition behind our method is to incrementally update the representation matrix based on the generalized propagation framework. Specifically, we adopt the computing framework of Personalized PageRank to obtain an approximate version of the representation matrix $\Z$, which is inspired by \cite{bojchevski2020pprgo,wang2021agp}. When the graph changes, we adjust the representation vectors of the affected nodes locally, and the impact of the change on remaining nodes is calculated by propagation operation in the next step. As a result, our method provides instantaneous updates for node representations while guaranteeing the same error as the recalculation. The basis of the update strategy is based on the underlying mechanics of Personalized PageRank, i.e., we always maintain that the invariant holds throughout the evolution of dynamic graphs, as described in Lemma~\ref{lemma1}. Extensive experimental results on real-world and synthetic datasets demonstrate the effectiveness and efficiency of InstantGNN. Especially, InstantGNN improves the computing efficiency of large graphs with millions of nodes by one-to-several orders of magnitude, while keeping comparable accuracy.
Our contributions can be summarised as follows:
\begin{itemize}[leftmargin = *]
  \item We propose an Instant Graph Neural Network model for dynamic graphs, namely \emph{InstantGNN}. Our method is able to compute the node representation matrix incrementally based on the result of previous steps, avoiding unnecessary re-computation and with guaranteed accuracy. Instead of just the dynamic graph structure, the method deals with dynamic node features too.
  \item We provide strong guarantees on the total running time of computing and updating the approximate propagation matrix on undirected graphs, as shown in Theorem~\ref{theorem:3}\&\ref{theorem:4}.
  \item Most previous work is exclusively evaluated on graphs with static node labels. We construct a real-life dataset from Aminer Network coauthor data, which contains both dynamic structure and dynamic node labels.
  \item Our method can create an adaptive forecast for the optimal time to retrain the model, which is useful for maximizing performance gains with a fixed budget for computational costs.
\end{itemize}

%% file: pre.tex
\begin{table}[t]
  \caption{Table of notations}
  \vspace{-3mm}
  \label{tab:notation}
  \begin{tabular}{c|l}
    \hline {\bf Notation} & {\bf Description}\\ \hline
    $G=(V, E)$ & the graph with vertex set $V$ and edge set $E$ \\
    $\bm{x}$ & the graph signal vector \\
    $\bm{\pi}$, $\bm{\hat{\pi}}$ & the true and estimated propagation vectors \\
    $\bm{r}$ & the residual vector \\
    $\bm{e}_s$ & the one-hot vector with $\bm{e}_s(s)=1$ \\
    $d(s)$ & the degree of node $s$ \\
    $N(s)$ & the neighbor set of node $s$ \\
    $M$ & the graph event set, $M=\{Event_1,\dots,Event_k\}$ \\
    $G_i$, $E_i$ & the graph and edge set at time $i$ $(0\leq i\leq k)$ \\
    $\A_i$, $\D_i$ & the adjacency matrix and degree matrix of $G_i$ \\
\bottomrule
\end{tabular}
\end{table}

\vspace{-5mm}
\section{Preliminaries} 
\label{sec:pre}
We consider an undirected graph $G=(V,E)$, where $V$ is the vertex set with $n=|V|$ nodes, $E$ denotes the edge set with $m=|E|$ edges. \eat{The interconnection relationship between nodes is defined by the adjacency matrix $\A$. The degree matrix $\D$ is a diagonal matrix in which each element is defined as $\D_{ii}=\sum_{j=1}^n{\A_{ij}}$. }Detailed notations are shown in Table~\ref{tab:notation}.

\vspace{-3mm}
\subsection{Dynamic Graphs}
\label{sec:dynamic_graphs}
In general, dynamic graphs can be broadly classified into two categories: Continuous-Time Dynamic Graph (CTDG) and Discrete-Time Dynamic Graph (DTDG). Given an initial graph $G_0$, a CTDG describes a continuous graph evolving process which can be formulated as sequential events $\{ Event_1, \dots, Event_k\}$, where $k$ is the number of graph events. Specifically, $Event_i \hspace{-0.5mm} = \hspace{-0.5mm} (type_i, e_i)$ denotes an edge operation of $type_i$ conducting on edge $e_i\in E_i$ at time $i$. The types of edge operations are restricted as \emph{InsertEdge} and \emph{DeleteEdge}. For the sake of simplicity, we assume the graph evolves from time $i=0$ and only one edge operation is conducted on one edge at each time step. More precisely, let $G_i$ denote the graph topology at the time step $i$. Then $G_i$ can be derived after $i$ subsequent edge operations (i.e. $\{ Event_1, \dots, Event_i\}$) from $G_0$. A DTDG is a sequence of static graph snapshots taken at the time intervals, represented as $S=\{G_0, \dots, G_{|S|}\}$. A graph snapshot $G_i$ in a DTDG can be treated as the graph captured at time $i$ of a CTDG, while the detailed evolving process (e.g. $\{Event_1, \ldots, Event_i\}$) is omitted. Therefore, CTDGs are more general than DTDGs~\cite{kazemi2020representation}, and we aim to design GNN algorithms for CTDGs.

\noindent
\textbf{Dynamic Attributes.} We note that for real-world graphs, both graph topologies and node attributes change over time. For example, consider the paper collaboration networks with authors as graph nodes and the community each author belongs to as the node attribute. The changes of authors' research interests result in dynamic node attributes. Thus, in Section~\ref{sec:dynamic_attribute} and ~\ref{sec:cost_analysis}, we consider the CTDG with dynamic node attributes (DA-CTDG). Specifically, we use $\X_i$ to denote the node feature matrix at the time step $i$. And $\X_i(s)$ corresponds to the attribute vector of node $s$. 

\noindent\textbf{Problem Definition.} We consider a CTDG, a dynamic attribute graph\eat{ if it has} with a fixed node set $V$, a dynamic edge set, and a dynamic feature matrix.\eat{ For simplicity, we assume that the structure and features of the graph do not change at the same time.} Let $\Z_i$ denote the graph representation matrix of the graph $G_i$\eat{ at time $i$}, $\Z_{i+1}$ is the matrix after operating $Event_{i+1}$ or updating node features to $\X_{i+1}$. The problem of\eat{ instant graph neural network for dynamic graphs} InstantGNN is to calculate the representation matrix $\Z_{i+1}$ incrementally and instantly.

\vspace{-3mm}
\subsection{Personalized PageRank (PPR) based GNNs}
The mainstream methods of propagation in GNNs are implemented based on Personalized PageRank (PPR). PageRank (PR)~\cite{page1999pagerank} is a web ranking algorithm used by Google, and PPR is its personalized version. PPR measures the importance of nodes from the point of view of a given node, and it has been widely used in web search engines~\cite{page1999pagerank,jeh2003ppr_web} and social recommendations~\cite{gupta2013ppr_social}. Given a graph $G$ and a starting node $s$, the PPR score reflects the relative importance of node $t\in V$ with respect to node $s$. Formally, the PPR is calculated as follows:
\begin{equation}
  \label{equ:eq_pprdefinition}
  \bm{\pi}_s=\alpha \bm{e}_s + (1-\alpha)\A\D^{-1}\bm{\pi}_s,
\end{equation}
where $\alpha$ is the teleport probability and $\alpha \hspace{-0.5mm}\in\hspace{-0.5mm} (0,1)$, $\bm{e}_s$ is a one-hot vector with $\bm{e}_s(s)=1$\eat{ with $\bm{e}_s(s)=1$ the $s$-th entity $\bm{e}_s(s)$ set to 1 and others $\bm{e}_s(t), (t\hspace{-1mm}\in\hspace{-1mm} V, t \hspace{-0.5mm}\neq\hspace{-0.5mm} s)$ set to 0}. By solving Equation~\ref{equ:eq_pprdefinition} and using a power series expansion, an equivalent form is given by $\sum_{\ell=0}^\infty \alpha(1-\alpha)^\ell (\A\D^{-1})^\ell \bm{e}_s$.

Efficient methods for calculating PPR have been extensively studied. A commonly used approximate version is known as Forward Push~\cite{andersen2007forwardpush}, which focuses on the region close to the source nodes and avoids traversing the entire graph. Based on the significant performance of PPR in applications such as recommendations, researchers found that PPR is incredibly beneficial for understanding and improving graph neural networks~\cite{klicpera2018predict} recently. Many methods, including PPRGo~\cite{bojchevski2020pprgo}, GBP~\cite{chen2020scalable}, and AGP~\cite{wang2021agp}, provide enhanced propagation strategies based on approximation PPR, in order to accelerate the propagation process and achieve efficient computing of GNNs.
Similarly, the method proposed in this paper adopts and extends the approximate dynamic PPR~\cite{zhang2016approximate} to solve the GNN propagation problem in dynamic graphs.

%% file: related_work.tex
\vspace{-2mm}
\section{Related Work}
\subsection{Scalable GNNs}
Due to the high complexity of traditional full-batch GNN approaches, they have difficulty supporting large-scale graphs with millions of nodes. Numerous methods have been introduced to tackle this problem, which can be summarized into two categories: sampling-based methods and linear models. 

\noindent\textbf{Sampling-based methods} divide a large graph into smaller segments by sampling subsets of neighbors or sub-graphs of the original graph. GraphSage~\cite{hamilton2017inductive} is the first to consider the scalability of graph neural networks, in which each node only aggregates the information from its sampled neighbors. Another node sampling approach, VR-GCN~\cite{chen2018VRGCN}, further reduces the sampling size but still does not solve the problem of exponential growth in the number of neighboring nodes. Layer sampling approaches, such as FastGCN~\cite{chen2018fastgcn} and ASGCN~\cite{zhang2019ASGCN}, successfully expand to large-scale graphs by sampling a fixed number of nodes at each layer of the GCN. However, overlapping nodes in different batches introduce significant computational redundancy, reducing the method's efficiency. Graph sampling methods represented by ClusterGCN~\cite{chiang2019cluster} and GraphSAINT~\cite{zeng2019graphsaint} obtain a series of subgraphs of the original graph by exploiting the graph clustering structure or subgraphs sampler. The problem of exponential growth of neighbors can be remedied by restricting the size of subgraphs.

\noindent\textbf{Linear models} eliminate the nonlinearity between layers in the forward propagation, allowing us to combine different neighboring aggregators in a single layer and design scalable models without relying on samplings.
SGC~\cite{wu2019simplifying} and PPNP~\cite{klicpera2018predict} discover that the propagation and neural network (activations) can be separated in GNNs and therefore open the door for simpler or approximate methods.
GDC~\cite{klicpera2019diffusion} and GBP~\cite{chen2020scalable} capture information about multi-hop neighborhoods using PPR and heat kernel PageRank and convert the initial feature matrix to a representation matrix for subsequent tasks. PPRGo~\cite{bojchevski2020pprgo} calculates the approximate PPR matrix by precomputation and then applies the PPR matrix to the feature matrix, deriving the final propagation matrix. The precomputation deals with all message passing operations on the graph topology. 

\vspace{-4mm}
\subsection{Dynamic Graph Representation Learning}
The straightforward way to handle dynamic graphs is to apply the traditional static GNN models, for example, GCN~\cite{kipf2016semi}, GAT~\cite{velivckovic2017graph} and GraphSage~\cite{hamilton2017inductive}, to learn the node representation on every graph snapshot. However, generating new representation at each time step is computationally costly, since most traditional graph neural network methods learn node representation by retraining the whole model. Some researchers proposed to combine temporal modeling with node representation learning. DySAT~\cite{sankar2020dysat} learns node representations on each graph snapshot and employs attention mechanisms to learn structural and temporal attention to integrate useful information into node representations adaptively. To this end, a common solution is to use graph learning algorithms to collect structural information and utilize sequence models, such as Recurrent Neural Networks (RNNs), to capture temporal dependencies and add consistency constraints between adjacent time steps~\cite{goyal2018dyngem, goyal2020dyngraph2vec, manessi2020dynamic, pareja2020evolvegcn, taheri2019learning}. \eat{Within this framework, WD-GCN~\cite{manessi2020dynamic}, GC-LSTM~\cite{chen2018gclstm}, and DyGGNN~\cite{taheri2019learning} employ GCN~\cite{kipf2016semi} to learn node representations and use Long Short Term Memory (LSTM)~\cite{hochreiter1997lstm} to capture temporal properties. }These methods view a dynamic graph as a discrete evolution process. The graph is sampled with an interval into snapshots which consist of a large number of graph changes instead of a continuous stream of atomic graph events that consistently change the graph.

\eat{To reduce the loss of information, }Nguyen et al.~\cite{nguyen2018continuous} suggest modeling the spatio-temporal network as a continuous model. They add a temporal constraint to random walk so that both the structural and temporal patterns of the nodes can be captured. Similarly, HTNE~\cite{zuo2018htne} introduces the neighborhood formation to describe the evolutionary process of nodes and applies the Hawkes process to capture the influence of historical neighbors on nodes. Through an attention mechanism, DyRep~\cite{trivedi2019dyrep} creates an RNN-like structure, embedding structural information about temporal evolution into node representations. And JODIE~\cite{kumar2019jodie} proposes a coupled recurrent neural network architecture and applies two independent RNNs to update the embeddings of users and items.
\eat{ This paper mainly focuses on modeling the continuous evolution progress of graphs, with an efficient way to compute the latest node representations upon any atomic graph changes. Our architecture is based on a GNN, which enables us to include node attributes into the learning process automatically.}

\noindent\textbf{Limitation.} To the best of our knowledge, the problem defined in Section~\ref{sec:dynamic_graphs} has not been addressed in the existing literature. Existing scalable graph neural networks are unfortunately inefficient for our problem as they are optimized for static graphs~\cite{wang2021agp,bojchevski2020pprgo}. They assume static graphs and compute representation matrixes from scratch whenever a graph is updated. Since the graph is updated frequently in dynamic environments, it is computationally expensive or even unviable to employ such techniques. On the other hand, temporal GNNs focus on modeling evolution rather than enhancing the efficiency of generating node representations, and they are challenging to scale to large graphs. \eat{As mentioned above, it is still unclear how to integrate node properties with embeddings.}


%% file: algorithm.tex
\vspace{-2mm}
\section{Instant Graph Neural Network} 
\label{sec:algorithm}
In real-world graph-like structures, the relations between nodes may change at any time, which means that graphs evolve continuously and, in certain cases, are never-ending. It is sub-optimal to model these changes with discretely sampled graph snapshots, as firstly changes will not be reflected in the representations in time, and secondly the full model needs to be retrained from the entire graph at each time step, raising efficiency and scalability issues. \eat{In those methods, predictions are made using the updated model at a fixed time interval, i.e. model retain at each snapshot. This approach consumes a large number of computational resources and yet does not guarantee the timeliness of the node representational nor the prediction results.} To address this problem, we use incremental, approximate learning for GNN, which effectively handles the continuous graph changes and make predictions with instantly updated representations. 
Our approach is essentially a scalable and online graph neural network. Scalability is achieved by decoupling the propagation from the prediction progress, allowing the model to be trained in small batches and scaled to large graphs. Then we update graph representation vectors for necessary nodes only to attain real-time.\eat{ The static propagation method is used to obtain the initial representation matrix and maintain results meeting the error tolerance.} 

In this section, we present an instant propagation method, which can be recognized as an extension for dynamic graphs based on the static method. For ease of understanding, we first introduce the basic propagation method. In Section~\ref{sec:instant2structural}, we describe details of the instant propagation method, especially how to calculate increments for CTDGs. Later Section~\ref{sec:dynamic_attribute} introduces the extension of our algorithm to DA-CTDGs. A theoretical guarantee of the running time is provided in Section~\ref{sec:cost_analysis}. Finally, we propose an adaptive training strategy based on InstantGNN in Section~\ref{sec:Adaptive_Training}.

\vspace{-2mm}
\subsection{Basic Propagation Method}
\label{sec:static}
Although our method is naturally applicable to general circumstances, for the sake of simplicity, we restrict $a+b\hspace{-0.5mm}=\hspace{-0.5mm}1$ in Equation~\ref{equ:eq1} and set $\beta\hspace{-0.5mm}=\hspace{-0.5mm}a,1\hspace{-0.5mm}-\hspace{-0.5mm}\beta\hspace{-0.5mm}=\hspace{-0.5mm}b$ in this paper.\eat{ setting $a\hspace{-0.5mm}=\hspace{-0.5mm}\beta,b\hspace{-0.5mm}=\hspace{-0.5mm}1\hspace{-0.5mm}-\hspace{-0.5mm}\beta$ and $\P=\D^{-\beta}\A\D^{\beta-1}$.} Treating $\X\hspace{-0.5mm}\in\hspace{-0.5mm} \mathbb{R}^{(n\times d)}$ as $d$ independent information vectors $\bm{x}$, $\Z$ can be split into $d$ segments, with each segment denoted by $\bm{\pi}$. Therefore, we can write Equation~\ref{equ:eq1} in a vector form: $\bm{\pi}\hspace{-0.5mm}=\hspace{-0.5mm}\sum_{\ell=0}^{\infty}\alpha(1-\alpha)^\ell \P ^\ell \bm{x}$. Theoretically, $\bm{\pi}$ has the effect of aggregating the information of all nodes on the graph using infinite hops (GNN layers), but computation time and training parameters are reduced. However, the summation goes to infinity, making it computationally infeasible to calculate the result exactly on large graphs. Following \cite{wang2021agp, bojchevski2020pprgo}, we consider an approximate version of $\bm{\pi}$ as vector $\bm{\hat{\pi}}$, which is the approximate solution under a given error limit $\varepsilon$. Meanwhile, we maintain a vector $\bm{r}$ with each element indicating the accumulated mass of information the node has received but not yet propagated. This enables us to design an approximate propagation process that achieves a speed-accuracy trade-off by adjusting $\bm{r}$. Algorithm~\ref{alg:1} describes the pseudo-code of the basic propagation progress. The algorithm takes the initial $\bm{\hat{\pi}}$ and $\bm{r}$ as input. Then, we iteratively select the node $s$ that exceeds the allowed error. Its $\alpha$-fraction of the residual mass is converted into the estimated mass, and the remaining ($(1\hspace{-0.5mm}-\hspace{-0.5mm}\alpha)\hspace{-0.5mm}\cdot\hspace{-0.5mm}\bm{r}(s)$) is distributed evenly among its neighbors using the transition matrix $\P$. For instance, let $\bm{\hat{\pi}}=\bm{0}$ and $\bm{r}=\bm{x}$ be the initial parameters, where $\bm{0}$ is a zero vector, Algorithm 1 outputs the propagation result corresponding to the current graph structure. As a result, we obtain $\bm{\hat{\pi}}$ as an estimator for the graph propagation vector $\bm{\pi}$ which satisfies the condition $|\bm{\hat{\pi}}(s) - \bm{\pi}(s)| \leq \varepsilon\hspace{-0.5mm}\cdot\hspace{-0.5mm} d(s)^{1-\beta}$ for each node $s\in V$. We add self-loops for all nodes, following the setting in GCN~\cite{kipf2016semi}, which guarantees that the degree of node $s\in V$ is at least 1. \eat{It also implies that each node has at least one neighbor, i.e. the node itself. }Therefore, if node $s$ is a dangling node, it only propagates and aggregates information of its own, such that $|\bm{\hat{\pi}}(s) - \bm{\pi}(s)| \leq \varepsilon\cdot d(s)^{1-\beta}$ holds.
 
Therefore, the relationship between $\bm{x}$, $\bm{\pi}$, $\bm{\hat{\pi}}$ and $\bm{r}$ can be described as Lemma~\ref{lemma1}, which is the basic premise for calculating the increment when the graph changes, and we shall discuss it later. Due to the space limit, we only present the conclusion and defer detailed proof to the technical report~\cite{technical_report}. 
\vspace{-2mm}
\begin{lemma}
\label{lemma1}
For every graph signal vector $\bm{x}$, let $\bm{\pi}$ be the true propagation vector, the estimated vector $\bm{\hat{\pi}}$ and residual vector $\bm{r}$\eat{, as defined and updated above, always} satisfy the following invariant property during the propagation process:
\vspace{-1mm}
\begin{equation}
  \label{equ:invariant_vector}
   \bm{\hat{\pi}} + \alpha \bm{r} = \alpha \bm{x} + (1-\alpha) \P\bm{\hat{\pi}}.
\end{equation}
\end{lemma}
\begin{algorithm}[t]
\setlength{\abovecaptionskip}{-2mm}
\setlength{\belowcaptionskip}{-2mm}
\caption{Basic Propagation Algorithm.}\label{alg:1}
\SetKwInOut{Input}{Input}
\SetKwInOut{Output}{Output}
\Input{Graph $G$\eat{$\hspace{-0.5mm}=\hspace{-0.5mm}(V, E)$}, error threshold $\varepsilon$, teleport probability $\alpha$, convolutional coefficient $\beta$, initial estimated propagation vector $\bm{\hat{\pi}}$, initial residual vector $\bm{r}$
}
\Output{Updated estimated propagation vector $\bm{\hat{\pi}}$ and residual vector $\bm{r}$}
\While{exist $s \in V$ with $|\bm{r}(s)|>\varepsilon \cdot d(s)^{1-\beta}$}
{
 $\bm{\hat{\pi}}(s)\leftarrow \bm{\hat{\pi}}(s)+\alpha\cdot \bm{r}(s)$\;
 \For{each $t\in N(s)$}
 {
    $\bm{r}(t)\leftarrow \bm{r}(t)+ \frac{(1-\alpha)\cdot\bm{r}(s)}{d(s)^{\beta} d(t)^{1-\beta}}$ \;
 }
 $\bm{r}(s)\leftarrow 0$\;
}
return $\bm{\hat{\pi}}$, $\bm{r}$\;
\end{algorithm}
\vspace{-2mm}
\subsection{Instant Propagation Method}
\label{sec:instant2structural}
Given a continuous-time dynamic graph $(G, M)$ with the static graph signal $\bm{x}$, we first apply Algorithm~\ref{alg:1} to obtain $\bm{\hat{\pi}}_0$ and $\bm{r}_0$ for the graph at time $i=0$. The approximate dynamic propagation algorithm aims to maintain a pair of $\bm{\hat{\pi}}$ and $\bm{r}$ while keeping errors within a limited range, i.e. $\bm{r}(s)\leq  \varepsilon\cdot d(s)^{1-\beta}$ for any node $s\in V$.\eat{ For simplicity, we regard an undirected edge as two directed edges with opposite directions. Let $d(s)$ be the out-degree of node $s$, $N(s)$ and $N^{in}(s)$ denote out-neighbor set and in-neighbor set of node $s$, respectively.} The key to solving the challenge is to maintain Equation~\ref{equ:invariant_vector} in the dynamic graph setting\eat{, after each graph event occurs}. By locally adjusting the affected nodes (by each graph event) to keep the equation holding, we can obtain feasible estimates and residuals corresponding to the updated graph. We start from the invariant equation of Lemma~\ref{lemma1} and rewrite it as an equivalent formulation that for each node $s\in V$, we have:
\begin{equation}
\label{equ:invariant_each_node}
    \bm{\hat{\pi}}(s) + \alpha \bm{r}(s) = \alpha \bm{x}(s) + \sum\limits_{t\in N(s)}\gamma(s,t) 
\end{equation}
where $\gamma(s,t)=\frac{(1-\alpha) \bm{\hat{\pi}}(t)}{d(s)^{\beta}d(t)^{1-\beta}}$ refers to the element related to node $t$ in the equation of node $s$.
 
Without loss of generality, let us suppose that we add a new edge to graph $G$ in the next time step. \eat{The situation of removing an edge can be solved in a similar process. }Considering the new edge as $(u, v)$, we found that only nodes $u$, $v$, $w\hspace{-1mm}\in \hspace{-1mm} N(u)$, and $y\hspace{-1mm}\in \hspace{-1mm}  N(v)$ do not satisfy Equation~\ref{equ:invariant_each_node}, since no other variables change except the degrees of node $u$ and $v$. 
By substituting the $s$ in Equation~\ref{equ:invariant_each_node} with $u$, it is easy to obtain $\bm{\hat{\pi}}(u) + \alpha \bm{r}(u) = \alpha \bm{x}(u) + \sum_{t\in N(u)}\gamma(u,t)$, which is failed after inserting edge $(u,v)$\eat{updating $d(u)$ to $d(u)+1$}. This is caused by two changes on the right-hand side of the equation:
\begin{itemize}[leftmargin = *]
  \item In denominator of the summation term, $d(u)$ changes to $d(u)+1$. 
  An intuitive explanation is that the mass $u$ pushed out would always be divided into several equal proportions for all neighbors of $u$. Since $v$ is added to be a new neighbor, the proportion should change to $\frac{1}{(d(u)+1)^\beta}$ accordingly.
  \item A new term, which refers to $\gamma(u^\prime,v^\prime)\hspace{-0.5mm}=\frac{\hspace{-0.5mm}(1-\alpha) \bm{\hat{\pi}}(v)}{ (d(u)+1)^{\beta} (d(v)+1)^{1-\beta}}$, is added, where $u^\prime$ and $v^\prime$ are the updated nodes $u$ and $v$ with degree $d(u)+1$ and $d(v)+1$, respectively.
\end{itemize}
The updated equation of node $u$ implies: $\bm{\hat{\pi}}(u) + \alpha \bm{r}^\prime(u) \hspace{-0.5mm}=\hspace{-0.5mm} \alpha \bm{x}(u) + \sum_{t\in N(u)} \gamma(u^\prime,t)+\gamma(u^\prime,v^\prime)$, where $\bm{r}^\prime(u)$ is the adjusted residual, also known as $\bm{r}(u) + \Delta \bm{r}(u)$. Such that we can obtain an accurate increment for residual of node $u$ by subtracting $\bm{r}(u)$ from $\bm{r}^\prime(u)$. Furthermore, we can use $\bm{\hat{\pi}}(u)+\alpha \bm{r}(u)-\alpha \bm{x}(u)$ to equivalently substitute $\sum_{t\in N\hspace{-0.5mm}(u)}\hspace{-0.5mm}\gamma(u,t)$. Then we\eat{ substitute it back into the equation of node $u$ and} finally obtain the result: 
\vspace{-2mm}
\begin{equation}
\label{equ:delta_r(u)}
    \hspace{-0.5mm}\Delta \bm{r}(u)\hspace{-0.5mm}=\hspace{-0.5mm} 
    \left(\bm{\hat{\pi}}(u)\hspace{-1mm}+\hspace{-1mm}\alpha \bm{r}(u)\hspace{-1mm}-\hspace{-1mm}\alpha \bm{x}(u)\right)
    \cdot \frac{d(u)^{\beta}\hspace{-0.5mm}-\hspace{-0.5mm} (d(u)\hspace{-1mm}+\hspace{-1mm}1)^{\beta}}{\alpha\cdot(d(u)\hspace{-1mm}+\hspace{-1mm}1)^{\beta}}\hspace{-0.5mm}+\hspace{-0.5mm}\frac{\gamma(u^\prime,v^\prime)}{\alpha}.
\end{equation}
Note that we only update $u$’s residual here, and its estimate $\bm{\hat{\pi}}(u)$ will be adjusted in the next step. 

For each node $w\hspace{-0.5mm}\in\hspace{-0.5mm} N(u)$, we observe that the equation $\bm{\hat{\pi}}(w) + \alpha \bm{r}(w) = \alpha \bm{x}(w) + \sum_{t\in N(w)}\gamma(w,t)$ turns to be invalid. It is also caused by the fact that the degree of node $u$ has changed. Specifically, the element containing node $u$ has changed to $\gamma(w,u^\prime)$. And we obtain the precise increment of node $w$ by altering its residual $\bm{r}(w)$ in a same fashion.

Updates to the affected nodes $v$ and $y\hspace{-1mm}\in\hspace{-1mm} N(v)$ are almost the same as above. As a result, we have made adjustments for all affected nodes to keep Equation~\ref{equ:invariant_each_node} holding to these nodes. However, the residuals of a few nodes may be outside of the error range after the correction, i.e., above $(\varepsilon\hspace{-0.5mm}\cdot\hspace{-0.5mm} d(s)^{1-\beta})$ or below $(-\varepsilon\hspace{-0.5mm}\cdot\hspace{-0.5mm} d(s)^{1-\beta})$. We then invoke Algorithm~\ref{alg:1} to help guarantee that the error of results is within the error tolerance. Due to the fact that majority of nodes already have residuals that meet the requirements of the given error, the running time is significantly less than the initialization time.

\begin{algorithm}[t]
\setlength{\intextsep}{-2mm}
\caption{Dynamic Propagation Algorithm.}\label{alg:2}
\SetKwInOut{Input}{Input}
\SetKwInOut{Output}{Output}
\Input{Dynamic graph $\{G_0, M=\{Event_1,\dots,Event_k \}\}$, graph signal vector $\bm{x}$, error threshold $\varepsilon$, teleport probability $\alpha$, convolutional coefficient $\beta$}
\Output{estimated propagation vector $\bm{\hat{\pi}}$,\newline residual vector $\bm{r}$}
$\bm{\hat{\pi}}\leftarrow \bm{0}, \bm{r}\leftarrow\bm{x}$\;
$\bm{\hat{\pi}}, \bm{r}\leftarrow Basic Propagation(G_0, \varepsilon, \alpha, \beta, \bm{\hat{\pi}}, \bm{r})$\;
\For{$Event_i=(e_i=(u,v), type_i) \in M$ and $i=\{1,\dots,k\}$}
{
  Generate $G_i$ from updating $G_{i-1}$ by $Event_i$\;
  UPDATE($u$, $v$, $i$, $type_i$)\;
  UPDATE($v$, $u$, $i$, $type_i$)\;
  $\bm{\hat{\pi}},\bm{r}\leftarrow Basic Propagation(G_i, \varepsilon, \alpha, \beta, \bm{\hat{\pi}}, \bm{r})$\;
}
return $\bm{\hat{\pi}}$, $\bm{r}$\;
\SetKwFunction{UPDATE} {UPDATE}
    \SetKwProg{Fn}{Function}{:}{}
    \Fn{\UPDATE{\emph{node} $u$, \emph{node} $v$, \emph{time} $i$, $type$}} {
        $\Delta \bm{r}(u) \leftarrow(\bm{\hat{\pi}}(u)+\alpha \bm{r}(u)-\alpha\bm{x}(u)) \cdot \frac{d_{i-1}(u)^{\beta}-d_{i}(u)^{\beta}}{d_{i}(u)^{\beta}}$\;
        \If{$type$ is \emph{InsertEdge}}
        {
        $\Delta \bm{r}(u) \leftarrow \Delta \bm{r}(u)+\frac{(1-\alpha) \bm{\hat{\pi}}(v)}{d_{i}(u)^{\beta} d_{i}(v)^{1-\beta}}$ \;
        }\
        \ElseIf{$type$ is \emph{DeleteEdge}}
        {
        $\Delta \bm{r}(u) \leftarrow \Delta \bm{r}(u)-\frac{(1-\alpha) \bm{\hat{\pi}}(v)}{d_{i}(u)^{\beta} d_{i}(v)^{1-\beta}}$ \;
        }
        $\Delta \bm{r}(u) \leftarrow \Delta \bm{r}(u)/\alpha $\;
        $\bm{r}(u) \leftarrow \bm{r}(u)+\Delta \bm{r}(u)$\;
        \For{each $w\in N_{i-1}(u)$}
        {
        $\Delta \bm{r}(w)\leftarrow\frac{(1-\alpha) \bm{\hat{\pi}}(u)}{\alpha \cdot d_{i}(w)^{\beta}} \cdot\left(\frac{1}{d_{i}(u)^{1-\beta}}-\frac{1}{d_{i-1}(u)^{1-\beta}}\right)$ \;
        $\bm{r}(w) \leftarrow \bm{r}(w)+\Delta \bm{r}(w)$\;
        }
    }
\end{algorithm}

According to the procedure described above, the situation of removing an edge can be solved in a similar process. The incremental update procedure differs from the insertion case only in the update of the degree, i.e. it should be $d(u)-1$ instead of $d(u)+1$. 
Let $G_i$ be the graph updated from $G_0$ with $\{ Event_1,\dots, Event_i\}, i=1,\dots,k$ indicates the time that $Event_i$ arrives. Algorithm~\ref{alg:2} describes the pseudo-code of generating the estimated propagation vector $\bm{\hat{\pi}}$ and residual vector $\bm{r}$ for a CTDG in the case that graph events arrive one by one in order. Each step in the outer loop of Algorithm~\ref{alg:2} proceeds as follows, where $e_i=(u, v)$ denotes the edge arrived at the current time step and $\bm{\hat{\pi}}$ denotes the node representation of a dimension at this step. First, each affected node ($u$, $v$, $w\hspace{-0.5mm}\in\hspace{-0.5mm} N_{i-1}(u)$ and $y\hspace{-0.5mm}\in\hspace{-0.5mm} N_{i-1}(v)$) updates the residual to maintain its own equation, line 5-6 in Algorithm~\ref{alg:2}. Note that this update step may cause the error on some nodes to surpass the bounded range. Then, starting with the affected nodes, we invoke Algorithm~\ref{alg:1} to reduce the error and propagate the effect of edge $e_i=(u, v)$ to the other nodes in the graph, line 7 in Algorithm~\ref{alg:2}. 

\noindent \textbf{Batch update.} To extend Algorithm~\ref{alg:2} to the batch update setting, given a set of graph events, we first update the graph and then calculate increments of residuals for the affected nodes. However, instead of iterating over all graph events, we compute only the final increments that are caused by the occurrence of $k$ graph events on each affected node. Parallel processing of the affected nodes can further improve the efficiency of the algorithm. Appendix contains the complete \eat{batch update }pseudo-code.

\vspace{-3mm}
\subsection{Extension to Dynamic-attribute Graphs}
\label{sec:dynamic_attribute}
This section discusses the extension of the proposed algorithm to handle DA-CTDGs, which are defined in Section~\ref{sec:dynamic_graphs}. The challenge is that the structure and attribute of the graph are both changing over time. For ease of implementation, we first deal with the dynamic attributes, then employ Algorithm~\ref{alg:2} to treat structural changes and obtain final results at each time step. The basic idea is to make local adjustments to ensure Equation~\ref{equ:invariant_each_node} still holds for every node, which is similar to Section~\ref{sec:instant2structural}. The main difference is that every node could be an influenced node, since each element of the graph signal vector $\bm{x}$ may have changed.

We presume that the underlying graph structure is static while dealing with dynamic attributes. Let $G_i$ be the graph at time $i$, where $0\leq i\leq k$ and $k$ is the number of graph events. We have $\A_0\hspace{-0.5mm}=\hspace{-0.5mm}\A_1\hspace{-0.5mm}=\hspace{-0.5mm}\dots\hspace{-0.5mm}=\hspace{-0.5mm}\A_k$, where $\A_i$ is the adjacent matrix of $G_i$. For each node $s\hspace{-0.5mm}\in\hspace{-0.5mm} V$, let $d_i(s)$ be the degree of node $s$ at time $i$. Recall the definition, it is easy to observe that $d_0(s)\hspace{-0.5mm}=\hspace{-0.5mm}\dots\hspace{-0.5mm}=\hspace{-0.5mm}d_k(s)$. Let $\bm{x}_i$ be the graph signal vector of $G_i$, an intuition is to add the increment of the graph signal $\Delta\bm{x}_i \hspace{-0.5mm}=\hspace{-0.5mm} \bm{x}_i \hspace{-0.5mm}-\hspace{-0.5mm} \bm{x}_{i-1} $ to the residual vector $\bm{r}_{i-1} $, i.e. $\Delta\bm{r}_i \hspace{-0.5mm}=\hspace{-0.5mm} \Delta\bm{x}_i $. From Equation~\ref{equ:invariant_each_node}, for each node $s\hspace{-0.5mm}\in\hspace{-0.5mm} V$, the updated equation implies:
\begin{equation*}
    \bm{\hat{\pi}}(s) + \alpha (\bm{r}(s) + \Delta\bm{r}(s)) = \alpha (\bm{x}(s) + \Delta\bm{x}(s)) + \sum\limits_{t\in N(s)} \frac{(1-\alpha)\bm{\hat{\pi}}(t)}{d(s)^{\beta }d(t)^{1-\beta}}.
\end{equation*}
As a result, setting $\Delta\bm{r}_i(s) = \Delta\bm{x}_i(s) $ is efficient to ensure that the equation holds for the graph $G_i$. After updating residuals for all impacted nodes, we can use the basic propagation algorithm to eliminate residuals exceeding the threshold until the desired error tolerance is reached. Following that, we can easily use Algorithm~\ref{alg:2} by temporarily keeping the graph signal static.

\vspace{-3mm}
\subsection{Analysis}
\label{sec:cost_analysis}
In this section, we provide a theoretical analysis of the computational complexity of the proposed\eat{dynamic propagation} method. All the proofs are deferred to the technical report~\cite{technical_report} due to space constraints. 
We begin by providing an expected upper bound on the running time of CTDGs based on the method discussed in Section~\ref{sec:instant2structural}. Following that, we examine the time required to manage dynamic attributes. Therefore, the total running time for DA-CTDGs can be easily calculated by adding these two components. For illustration purposes, we define a vector $\bm{\Phi}_i\hspace{-0.5mm}=\hspace{-0.5mm} \bm{1}^\top\hspace{-0.5mm} \cdot\hspace{-0.5mm}(\D_i^\beta\Q_i)$, where $\bm{1}$ is a vector with all elements set to 1, and $\D_i$ is the degree matrix of $G_i$. $\Q_i=\alpha [\I-(1\hspace{-0.5mm}-\hspace{-0.5mm}\alpha)\D_i^{-\beta}\A_i\D_i^{\beta-1}]^{-1}$ is a matrix in which $(s,t)$-th element is equal to $ d_i(s)^{1-\beta}\cdot\bm{\pi}_{si}(t)\cdot d_i(t)^{\beta -1}$. 

\noindent \textbf{Cost for CTDGs.} 
The total time cost is calculated by adding the initialization and the update time. We first present a theorem to bound the running time of step 2 of Algorithm~\ref{alg:2}.
\begin{theorem}[Cost of Initialization]
\vspace{-1mm}
\label{theorem:1}
Given an error threshold $\varepsilon$, the time of generating an estimate vector $\bm{\hat{\pi}}_0$ for the initial graph $G_0$, such that $|\bm{\hat{\pi}}_0(s)-\bm{\pi}_0(s)| \leq \varepsilon\cdot d_0(s)^{1-\beta}$, for any $s \hspace{-0.8mm}\in\hspace{-0.8mm} V$, using Algorithm~\ref{alg:1}, is at most $T_{init}=\frac{1}{\alpha \varepsilon}\cdot\left(\lVert \D_0^{\beta}\cdot \bm{\pi}_0 \rVert_1 - \lVert \bm{\Phi}_0^\top \cdot \bm{r}_0 \rVert_1\right)$.
\vspace{-1mm}
\end{theorem}
Next, we will show that the time to maintain a pair of estimated propagation and residual vector meeting the error tolerance in the evolution is only related to the increment. 
\begin{theorem}[Cost of Updates]
\vspace{-1mm}
\label{theorem:2}
Let $G_i$ denote the updated graph from $G_{i-1}$ guided by $Event_i$. The time for updating $\bm{\hat{\pi}}_{i-1}$ and $\bm{\r}_{i-1}$, such that $|\bm{\hat{\pi}}_i(s)-\bm{\pi}_i(s)| \leq \varepsilon\cdot d_i(s)^{1-\beta}$, for any node $s\hspace{-0.8mm}\in\hspace{-0.8mm} V$, is at most $T_i=\frac{1}{\alpha\varepsilon}\cdot\left(2\varepsilon+\lVert \bm{\Phi}_i^\top \cdot \Delta\bm{r}_i \rVert_1+\lVert \bm{\Phi}_{i-1}^\top \cdot \bm{r}_{i-1} \rVert_1-\lVert \bm{\Phi}_{i}^\top \cdot \bm{r}_{i} \rVert_1\right)$.
\vspace{-1mm}
\end{theorem}
Note that $\Delta\bm{r}_i$ is a vector with each element $\Delta\bm{r}_i(s), s\hspace{-0.5mm}\in\hspace{-0.5mm} V$, corresponding to the precise increment of node $s$’s residual. We only adjust the residuals of the affected nodes, such that $\Delta\bm{r}_i$ is sparse with values only at $u$, $v$, $w\hspace{-0.5mm}\in\hspace{-0.5mm}N(u)$, $y\hspace{-0.5mm}\in\hspace{-0.5mm} N(v)$ when $e_i=(u,v)$. With the assumption that the edges arrive randomly and each node has an equal probability of being the endpoint, we have $E[\lVert \bm{\Phi}_i^\top \cdot \Delta\bm{r}_i \rVert_1]=\frac{(6-4\alpha) \lVert \bm{\pi}_{i-1}\rVert_1}{\alpha n}+\frac{6\varepsilon}{\alpha}-2\varepsilon$, and details can be found in the technical report~\cite{technical_report}. Now we consider the expected time required for running Algorithm~\ref{alg:2}.

\begin{theorem}[Total Cost]
\vspace{-1mm}
\label{theorem:3}
Let $M=\{Event_1,…,Event_k\}$ be the graph event sequence. Given an error threshold $\varepsilon$, the time of generating an estimated vector $\bm{\hat{\pi}}_k$ for the graph $G_k$, such that $|\bm{\hat{\pi}}_k(s)-\bm{\pi}_k(s)| \leq \varepsilon\cdot d_k(s)^{1-\beta}$, for any node $s\in V$, is at most $T=T_{init}+\sum_{i=1}^k T_i=\frac{\lVert \D_0^{\beta}\cdot \bm{\pi}_0 \rVert_1}{\alpha \varepsilon} + \frac{6k}{\alpha^2} + (6\hspace{-0.5mm}-\hspace{-0.5mm}4\alpha)\sum_{i=1}^k \frac{\lVert \bm{\pi}_{i-1} \rVert_1}{\alpha^2 n\varepsilon }$.
\vspace{-1mm}
\end{theorem}
Thus, for the special case of $\beta=0, \bm{x}=\bm{e}_s$, the worst-case expected running time is $O(\frac{1}{\varepsilon}+k+\frac{k}{n\varepsilon})$. Typically $\varepsilon=\Omega(\frac{1}{n})$, we can achieve $O(\frac{1}{\varepsilon}+k)$, where $O(\frac{1}{\varepsilon})$ is the expected time for initialization. Therefore, we can maintain estimates using $O(1)$ time per update.

\noindent \textbf{Cost of handling dynamic attributes.} Time complex analysis in this case is almost the same, the main difference is that the graph structure is static. Such that we have $\lVert (\bm{\Phi}_i^\top - \bm{\Phi}_{i-1}^\top) \cdot \bm{r}_{i-1} \rVert_1 = \varepsilon\hspace{-0.5mm}\cdot\hspace{-0.5mm} \sum_{s\in V} (d_i(s) - d_{i-1}(s))\hspace{-0.5mm}= \hspace{-0.5mm} 0$.
Note that $\Delta\bm{r}_i$ may not be a sparse vector here. Applying the increment mentioned above, we have $\lVert \bm{\Phi}_i \Delta \bm{r}_i \rVert_1 = \sum_{s\in V} |\bm{\Phi}_i(s)\Delta \bm{x}_i(s)| = \sum_{s\in V} |d_i(s)^\beta \Delta \bm{x}_i(s)|$. Therefore, we have Theorem~\ref{theorem:4} to bound the time cost of the proposed method at the dynamic setting that the graph signal is changing over time. It is clear that the expected time required to update estimates is proportional to the signal increment.
\begin{theorem}[Total Cost for Handling Dynamic Attributes]
\vspace{-1mm}
\label{theorem:4} 
Let $\Delta \bm{x}_i$ be the increment of graph $G_i$ updated from $G_{i-1}$. Given an error threshold $\varepsilon$, the time of generating an estimated vector $\bm{\hat{\pi}}_k$ for the graph $G_k$, such that $|\bm{\hat{\pi}}_k(s)-\bm{\pi}_k(s)| \leq \varepsilon\cdot d_k(s)^{1-\beta}$, for any node $s\in V$, is at most $T=\frac{\lVert \D_0^{\beta}\cdot \bm{\pi}_0 \rVert_1}{\alpha \varepsilon} + \sum_{i=1}^k \frac{ \lVert \D_i^\beta \Delta \bm{x}_i \rVert_1}{\alpha \varepsilon}$.
\vspace{-1mm}
\end{theorem}
Since the propagation process on each dimension is independent, we perform the dynamic propagation on multiple features in parallel, which further accelerates the execution of algorithm.\eat{ Moreover, we explore a ``lazy'' variant of InstantGNN in our experiments, which calculates the precise increment for each graph event but only updates the representation matrix before generating predictions, achieving a further acceleration. Implement details can be found in the technical report~\cite{technical_report}.}

\vspace{-2mm}
\subsection{Adaptive Training}
\label{sec:Adaptive_Training}
Typically, we retrain models at fixed intervals on continuous-time dynamic graphs, which is known as \emph{Periodic training} and has been used in experiments of Sections~\ref{sec:static_label} and~\ref{sec:dynamic_label}. Let $k$ and $|S|$ be the total number of graph events and times allowed retraining, respectively. \emph{Periodic training} retrains the model $|S|$ times with an interval of $(k/|S|)$ during evolution, resulting in resource waste and sub-optimal results from retraining. Once a series of graph events have been evaluated, our method can detect instantly the degree of influence they have on the graph. Defining $\Delta \Z\hspace{-0.5mm}=\hspace{-0.5mm}\left \| \Z_i-\Z_{i-1} \right\|_F, 1\leq i\leq k$, we can track the degree of change for the current feature matrix compared to the previous one. Therefore, we propose an \emph{Adaptive training} strategy based on InstantGNN to control the interval width and obtain better performance in budget-constrained model retraining. Specifically, we consider that we should retrain the model only when a significant accumulative change is present, in other words, when the combined impact of numerous events surpasses a certain threshold, denoted as $\theta$.

\textbf{For graphs that satisfy specific graph change patterns}, we can easily predict when to retrain models next time by fitting their curves. Specifically, for a graph that exhibits changes that follow a certain distribution, after finding the first $r$ times for retraining under the guidance of $\theta$, we can fit the curve to predict remaining $(|S|-r)$ times when the model should be retrained. As a result, we can retrain the model frequently when the graph changes dramatically and often, while increasing the interval between retraining when the graph changes gently.
\textbf{For generic graphs}, since there is no prior knowledge of the graph change patterns, the problem turns into choosing the appropriate $\theta$. In the real-world, graph evolution is a long-term process. A feasible solution is to compute the average $\Delta \Z$ over the first few small intervals (e.g., first ten intervals of 10,000 graph events) and use it as the threshold $\theta$ to determine when the model needs to be retrained in the future.


%% file: experiments.tex
\begin{table}[t]
  \caption{Statistics of experimental datasets.}
  \label{tab:datasets}
  \vspace{-3mm}
  \begin{tabular}{l|l|l|l|l|l} \hline
    {\bf Dataset} & {$\boldsymbol{n}$} & $\boldsymbol{m}$ & $\boldsymbol{d}$ & $\boldsymbol{|C|}$ & $\boldsymbol{|S|}$ \\ \hline
    Arxiv & 169,343 & 1,157,799 & 128 & 40 & 17 \\
    Products & 2,449,029 & 61,859,012 & 100 & 47 & 16 \\
    Papers100M & 111,059,956 & 1,615,685,872 & 128 & 172 & 21 \\ 
    Aminer & 1,252,095 & 4,025,865 & 773 & 9 & 31 \\ \hline
    SBM-500$K$ & 500,000 & 6,821,393 & 256 & 50 & 10 \\
    SBM-1$M$ & 1,000,000 & 13,645,722 & 256 & 50 & 10 \\
    SBM-5$M$ & 5,000,000 & 68,228,223 & 1024 & 100 & 10 \\
    SBM-10$M$ & 10,000,000 & 136,469,493 & 1024 & 100 & 10 \\ \hline
\end{tabular}
\vspace{-3mm}
\end{table}
\section{Experiments} 
\label{sec:exp}
In this section, we conduct experiments on both real-world and synthetic datasets to verify the effectiveness and efficiency of InstantGNN. 
First, we compare the accuracy and efficiency of InstantGNN with the state-of-the-art methods on graphs with static labels and dynamic labels, respectively. Then, we conduct an experiment to demonstrate that InstantGNN is capable of adaptive training and achieving superior results while maintaining the same budget with periodic training.

\vspace{-1mm}
\subsection{Datasets and Baselines}
\label{sec:datasets_baseline}
{\bf Datasets.} We use three public datasets in different sizes, provided by~\cite{hu2020ogb}, for our experiments: a citation network Arxiv, a products co-purchasing network Products and a large-scale citation network Papers100M. To imitate the graphs' dynamic nature, we remove a certain number of edges from the original graphs and replace them later. Due to the scarcity of publicly available datasets for dynamic graph node classification, particularly those with dynamic labels, we construct a real-world dataset and four synthetic datasets with dynamic labels, to better evaluate the effectiveness of our method and facilitate the evolution of dynamic GNNs. Details of data pre-processing are deferred to the appendix.

\begin{itemize}[leftmargin = *]
  \item {\bf Aminer.} Following\eat{ a similar idea of}~\cite{zhou2018DynamicTriad}, we derive a co-authorship network from the academic social networks of Aminer~\cite{tang2008arnetminer}. It consists of 1,252,095 authors as vertices, and the coauthor relationships are considered as edges. \eat{Furthermore, we say an author belongs to a particular community if over half of his or her papers were published in corresponding journals this year. }The information about authors' communities is derived from \emph{Australian CORE Conference and Journal Ranking}~\cite{core} and used as labels in this dataset.
  \item {\bf SBM datasets.} We generate graphs with community structure using stochastic block model (SBM)~\cite{snijders1997estimation}. Based on a similar dynamic strategy with~\cite{yang2011dsbm}, we inherit nodes from the graph at the previous time step and change their connectivity to simulate the evolution of communities in the real world. The features of nodes are generated from sparse random vectors.
\end{itemize}
\begin{figure}[t]
\setlength{\abovecaptionskip}{2mm}
\setlength{\belowcaptionskip}{-2mm}
	\begin{small}
		\centering
		\begin{tabular}{cc}			 \hspace{-2mm}\includegraphics[height=27mm]{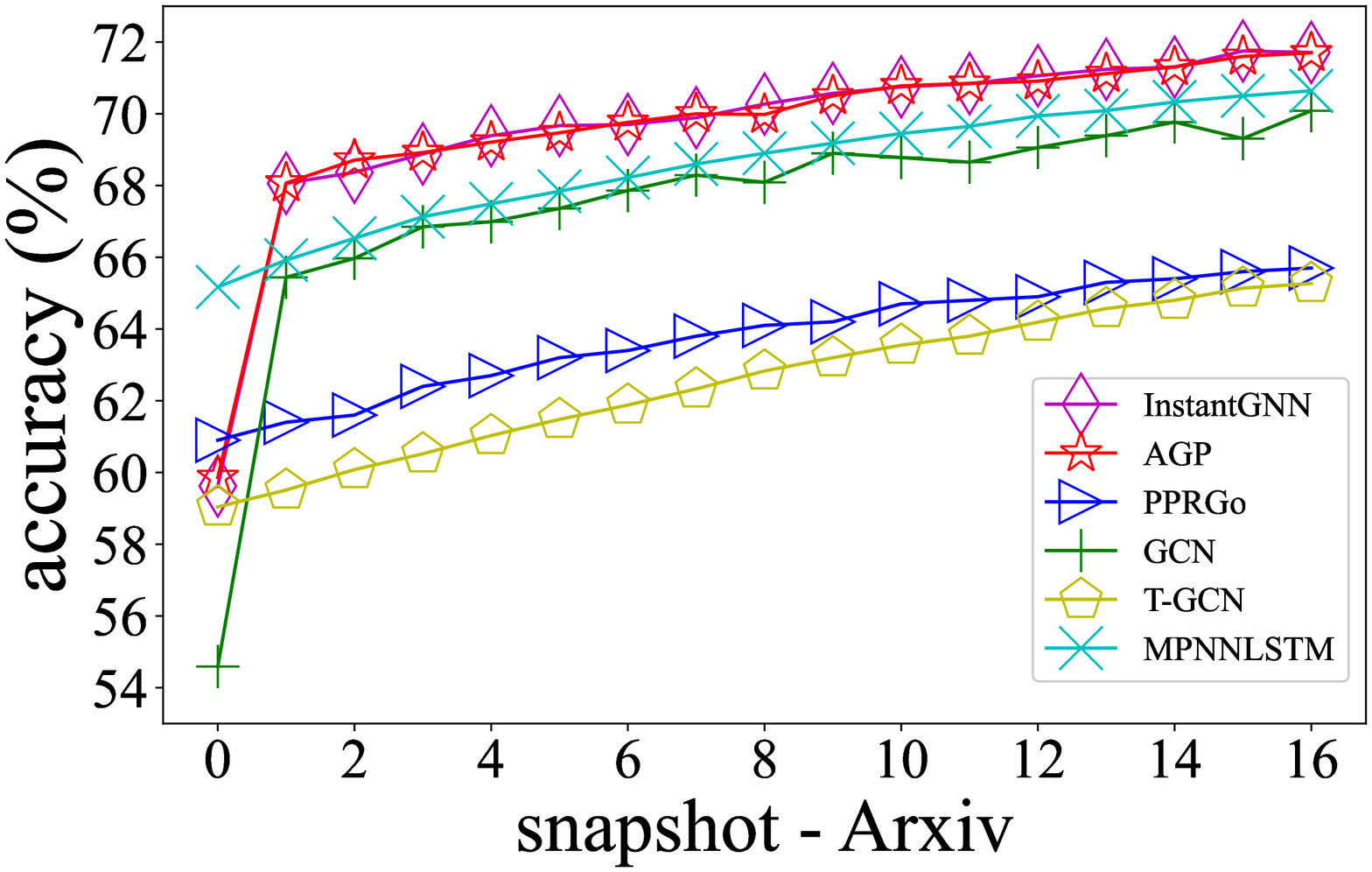} &			 \hspace{-4mm}\includegraphics[height=27mm]{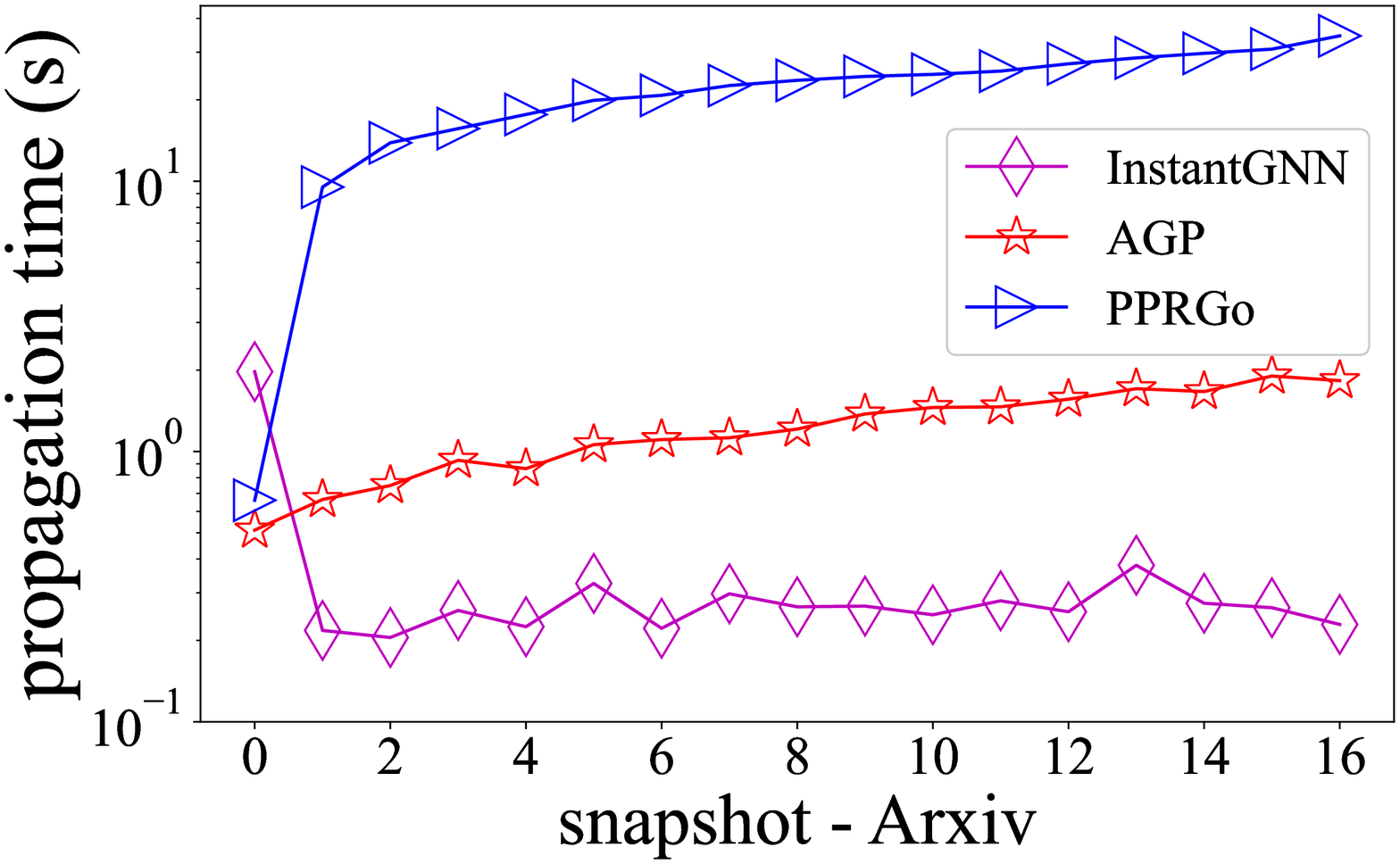}
		\end{tabular}
		\vspace{-4mm}
		\caption{Comparison of different methods on Arxiv.}
		\label{fig:arxiv_acc_time}
		\vspace{-1mm}
	\end{small}
\end{figure}

\begin{figure}[t]
\setlength{\abovecaptionskip}{2mm}
\setlength{\belowcaptionskip}{-2mm}
	\begin{small}
		\centering
		\vspace{-1.5mm}
		\begin{tabular}{cc}			 \hspace{-2mm}\includegraphics[height=27mm]{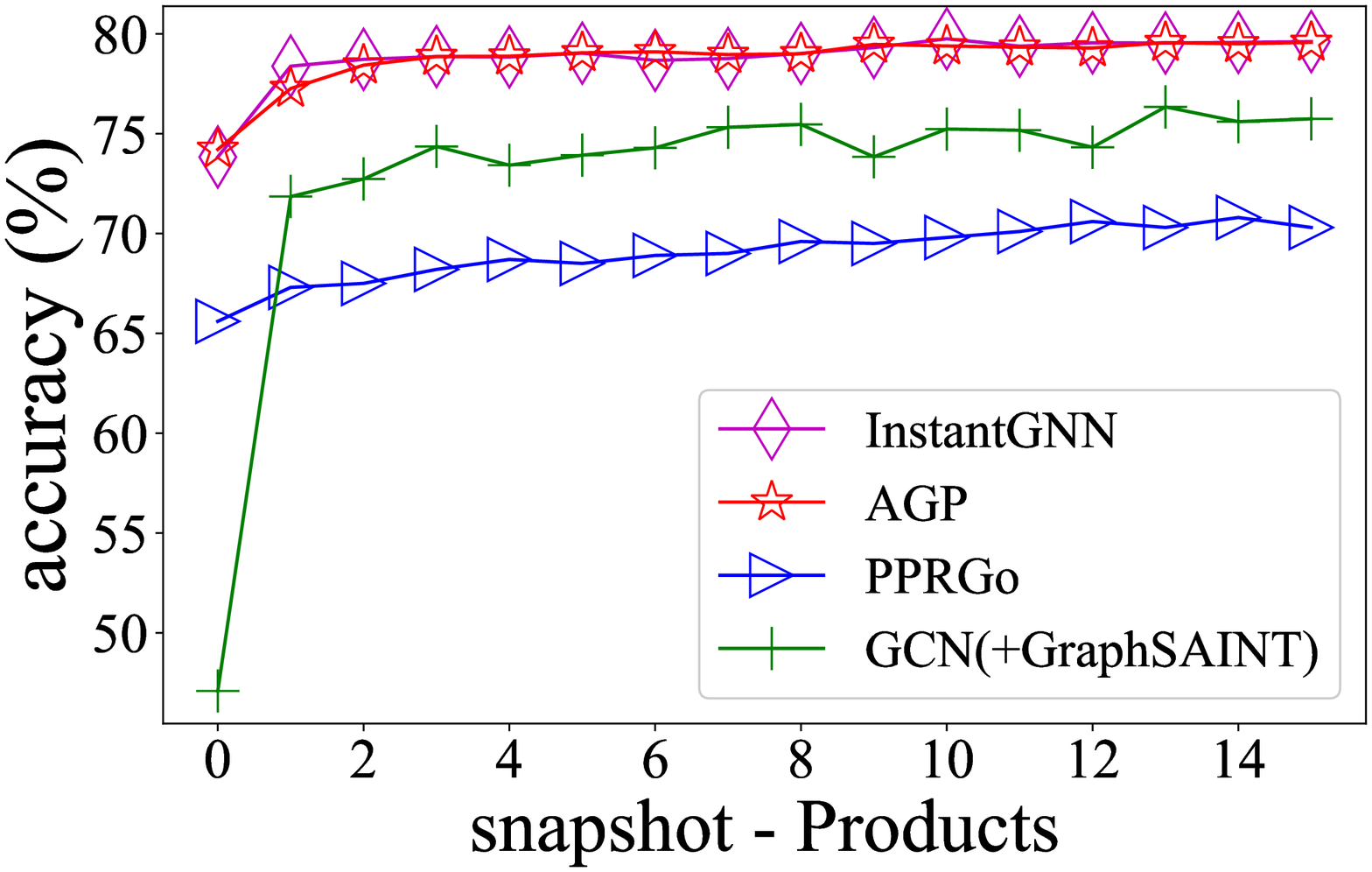} &			\hspace{-4mm}\includegraphics[height=27mm]{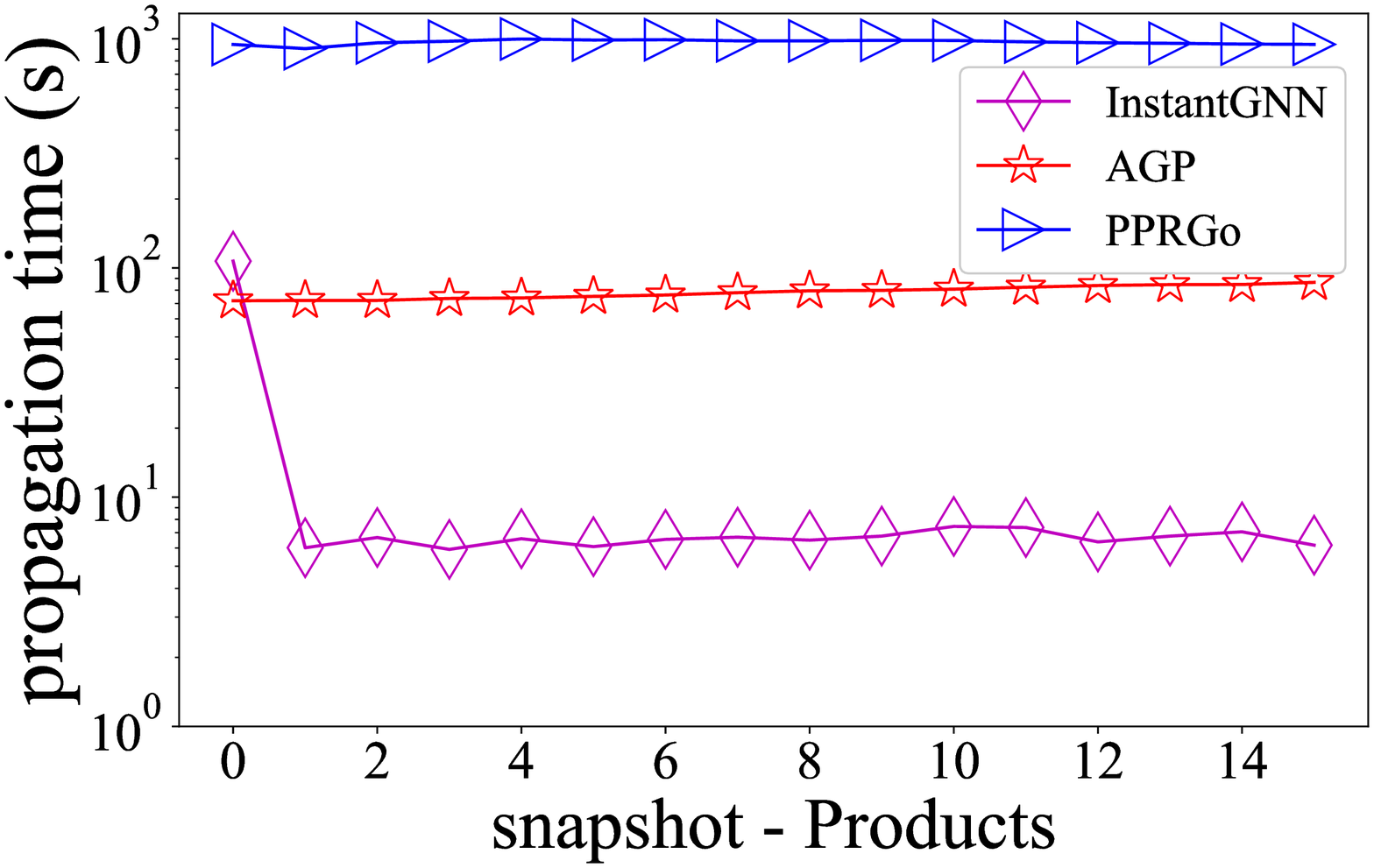}
		\end{tabular}
		\vspace{-4mm}
		\caption{Comparison of different methods on Product.}
		\label{fig:products_acc_time}
	\end{small}
\end{figure}

\begin{figure}[t]
\setlength{\abovecaptionskip}{2mm}
	\begin{small}
		\centering
		\vspace{-1.5mm}
		\begin{tabular}{ccc}
			 \hspace{-2mm}\includegraphics[height=27mm]{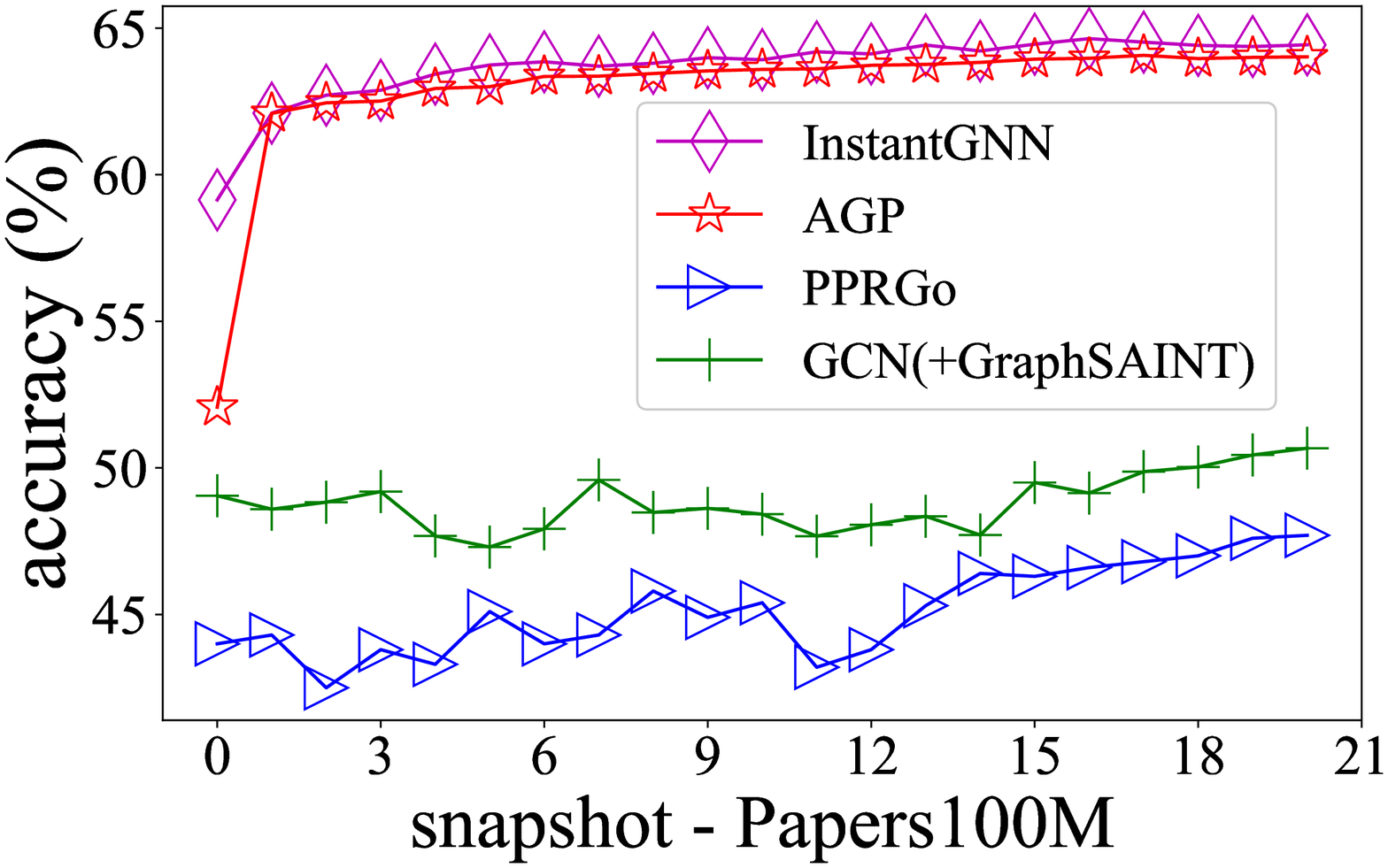} &
			 \hspace{-4mm}\includegraphics[height=27mm]{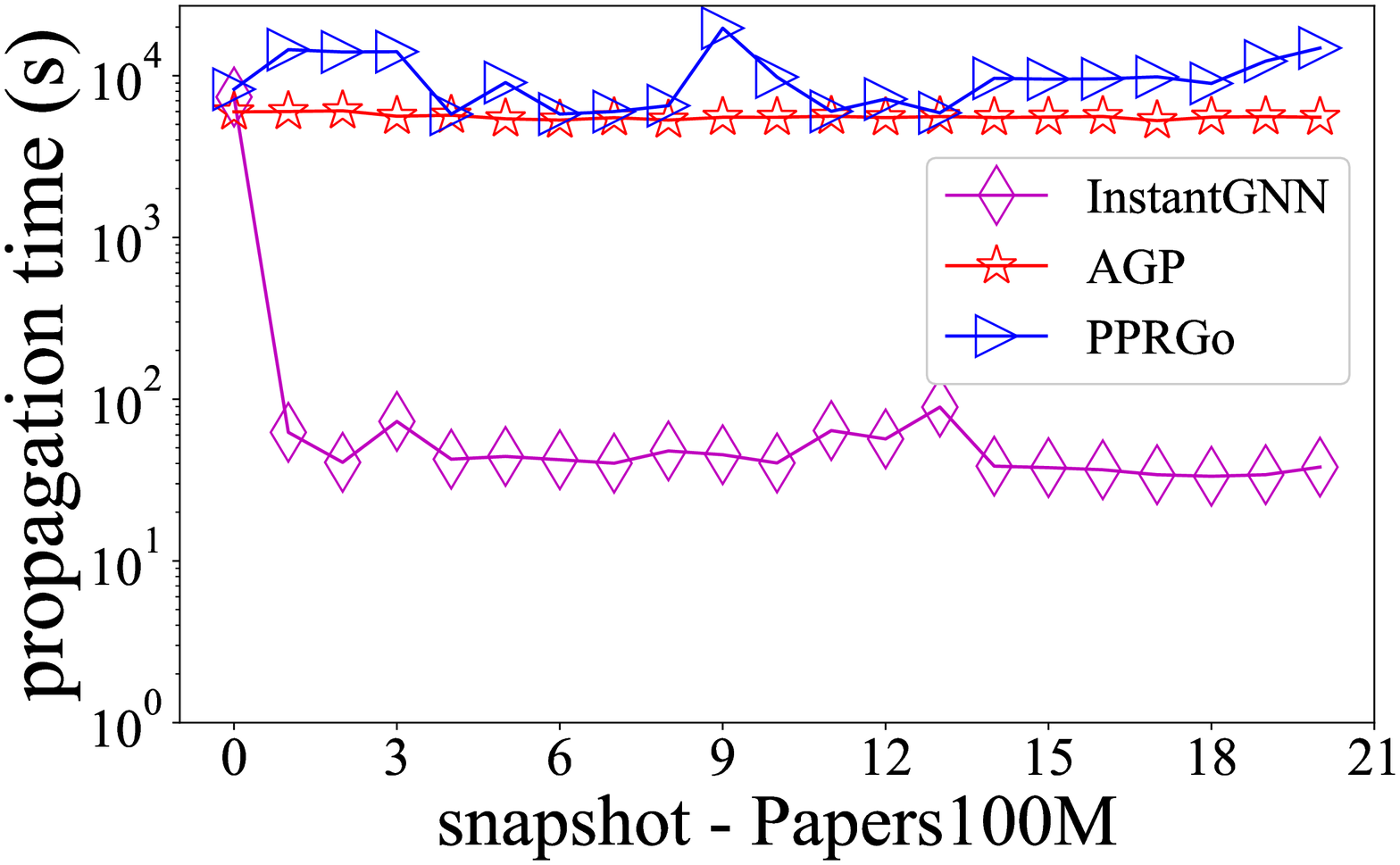}
		\end{tabular}
		\vspace{-4mm}
		\caption{Comparison of different methods on Papers100M.}
		\label{fig:papers100_acc_time}
		\vspace{-3mm}
	\end{small}
\end{figure}
\begin{figure}[t]
\setlength{\abovecaptionskip}{1mm}
\setlength{\belowcaptionskip}{-4mm}
	\begin{small}
		\centering
		\vspace{-2mm}
		\hspace{-2mm}\includegraphics[height=32mm]{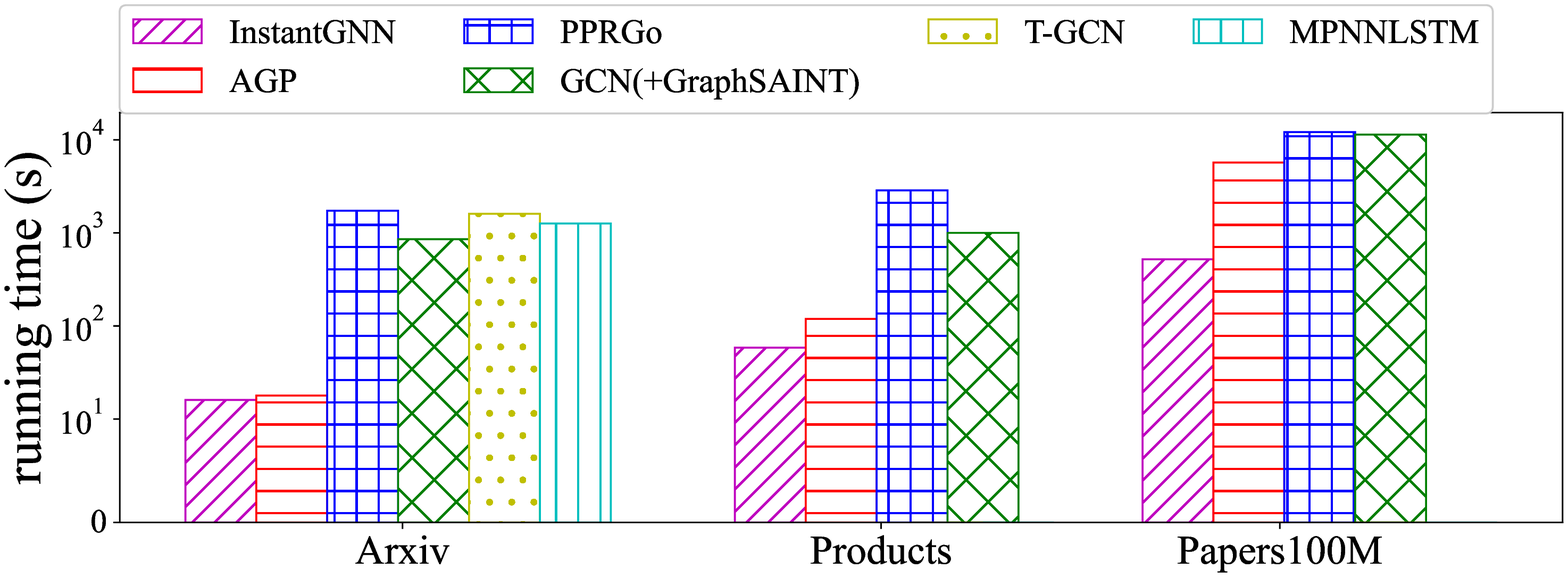}
		\vspace{-2mm}
		\caption{Average Running Time.}
		\label{fig:avg_running_time}
	\end{small}
\end{figure}
Table~\ref{tab:datasets} summarizes the properties and scales of the datasets used in the paper. For the number of edges of these datasets, we report the value at the last snapshot. Note that $d$ denotes the dimension of node features, $|C|$ is the number of categories, and $|S|$ represents the number of snapshots. All graphs in those datasets are transformed into unweighted and undirected graphs. 

We compare the proposed method with several state-of-the-art methods, including static graph learning methods and dynamic graph learning methods. Static methods include GCN~\cite{kipf2016semi} and two scalable GNN models, PPRGo~\cite{bojchevski2020pprgo} and AGP~\cite{wang2021agp}, and we train these models at each snapshot. In particular, we adopt GraphSAINT~\cite{zeng2019graphsaint} to sample subgraphs so that GCN can be trained on sampled subgraphs and scale to large graphs. We also compare InstantGNN with two temporal graph neural networks, including T-GCN~\cite{zhao2019tgcn} and MPNNLSTM~\cite{panagopoulos2021mpnnlstm}.\eat{ These three methods have similar structures that temporal layers and GNN layers are combined to learning the temporal information and the graph structure, and the only difference between is the specific temporal models and GNN models.} More details and parameter settings of baseline methods can be found in the appendix.\eat{ For baseline methods, we employ their published codes or the codes implemented with Pytorch Geometric Temporal~\cite{rozemberczki2021pytorch}.} For a fair comparison, we preserve\eat{ $\theta$ in AGP, $r_{max}$ in PPRGo, $\varepsilon$ in InstantGNN to keep} representation matrices of AGP, PPRGo and InstantGNN at the same approximation level, unless specifically stated.
\vspace{-1mm}
\subsection{Evaluation on CTDGs with Static Labels}
\label{sec:static_label}
\begin{figure}[t]
\setlength{\abovecaptionskip}{1mm}
\setlength{\belowcaptionskip}{-4mm}
	\begin{small}
		\centering
		\begin{tabular}{cc}
			 \hspace{-4mm}\includegraphics[height=27mm]{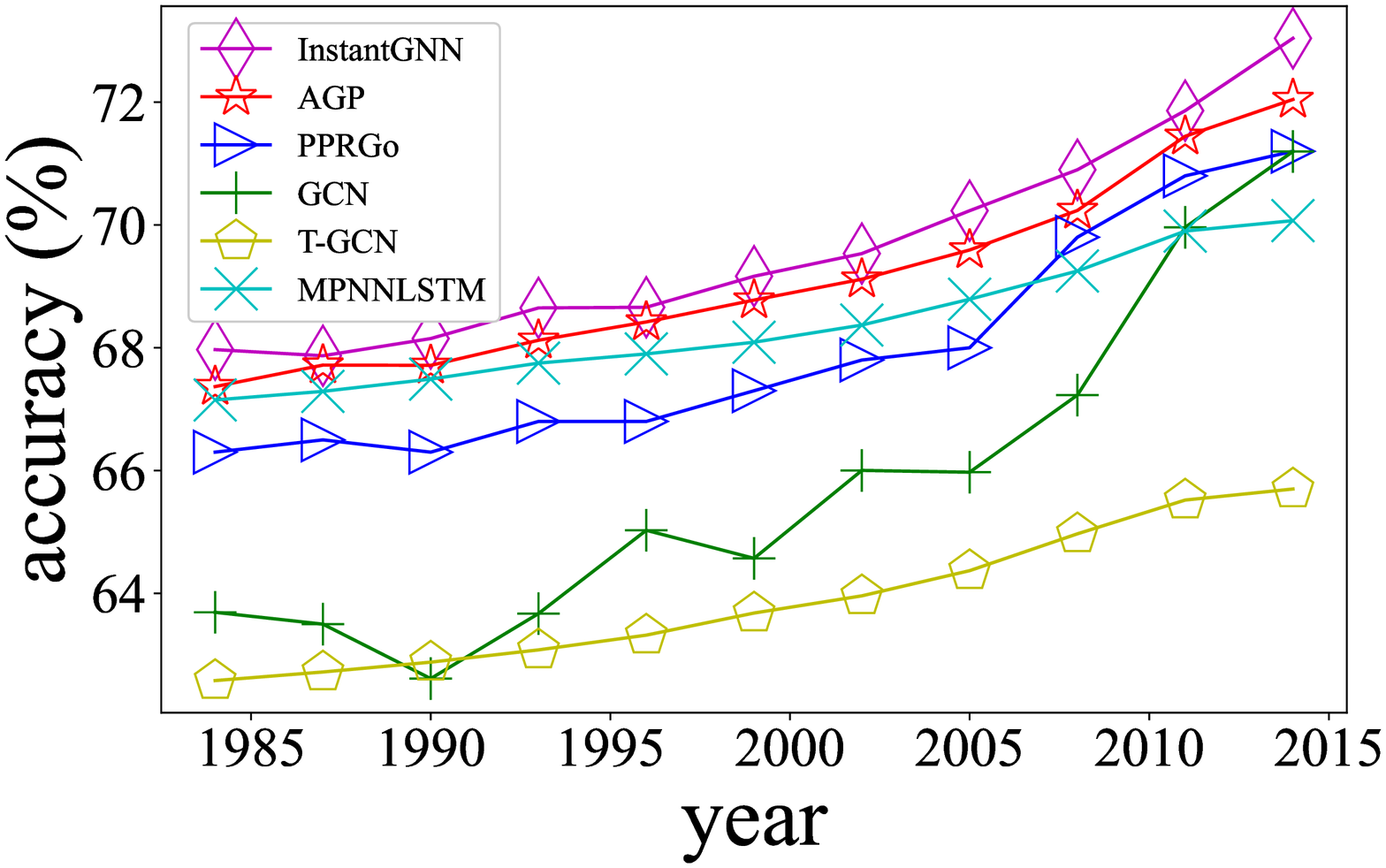} &
			 \hspace{-4mm}\includegraphics[height=27mm]{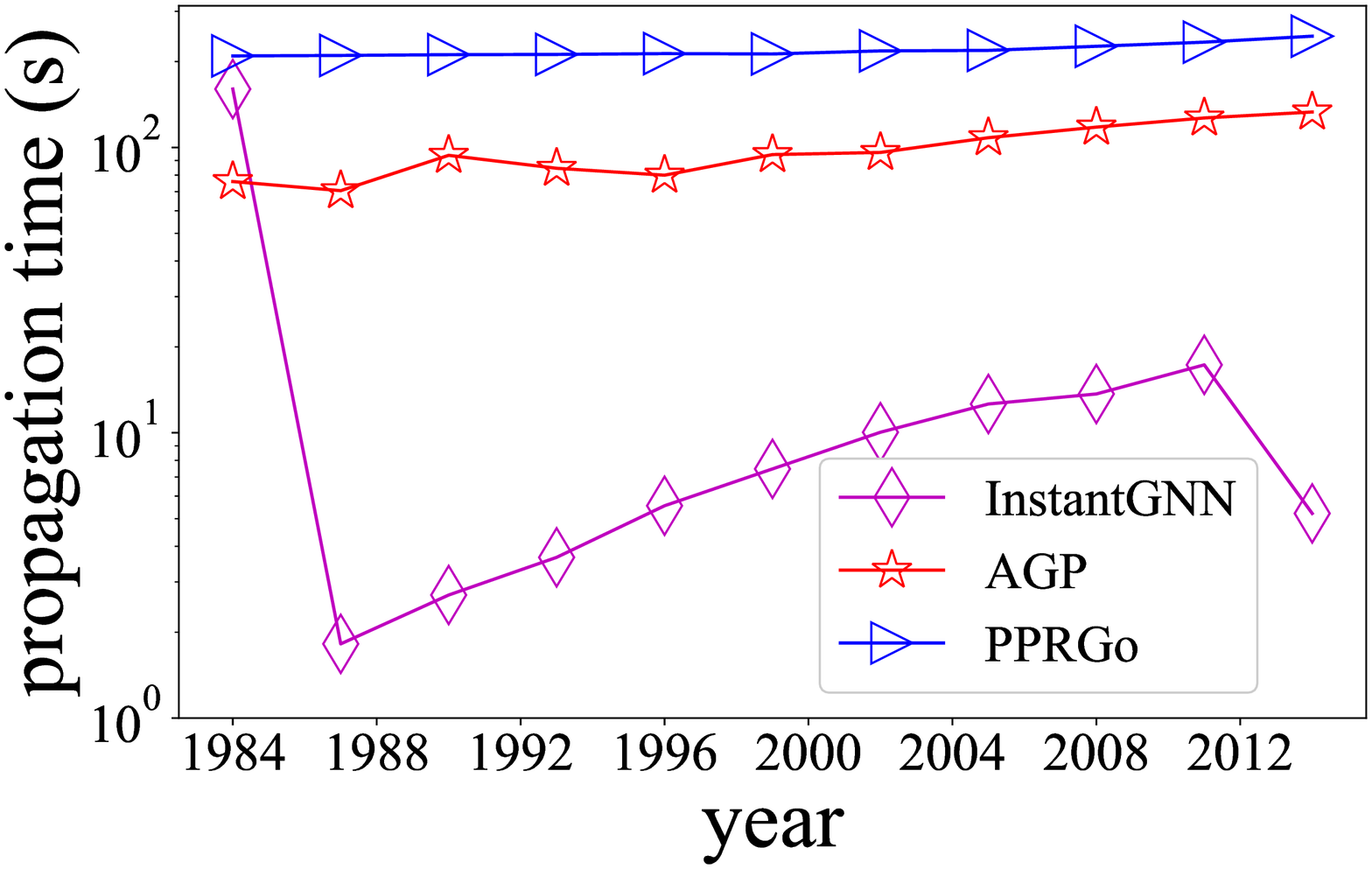}
		\end{tabular}
		\vspace{-3mm}
		\caption{Comparison on Aminer.}
		\label{fig:aminer_acc_time}
		\vspace{-3mm}
	\end{small}
\end{figure}
\begin{figure*}[t]
\setlength{\abovecaptionskip}{1mm}
\setlength{\belowcaptionskip}{-2mm}
	\begin{small}
		\centering
		\vspace{-4mm}
		\begin{tabular}{cccc}
			 \hspace{-2mm}\includegraphics[height=28mm]{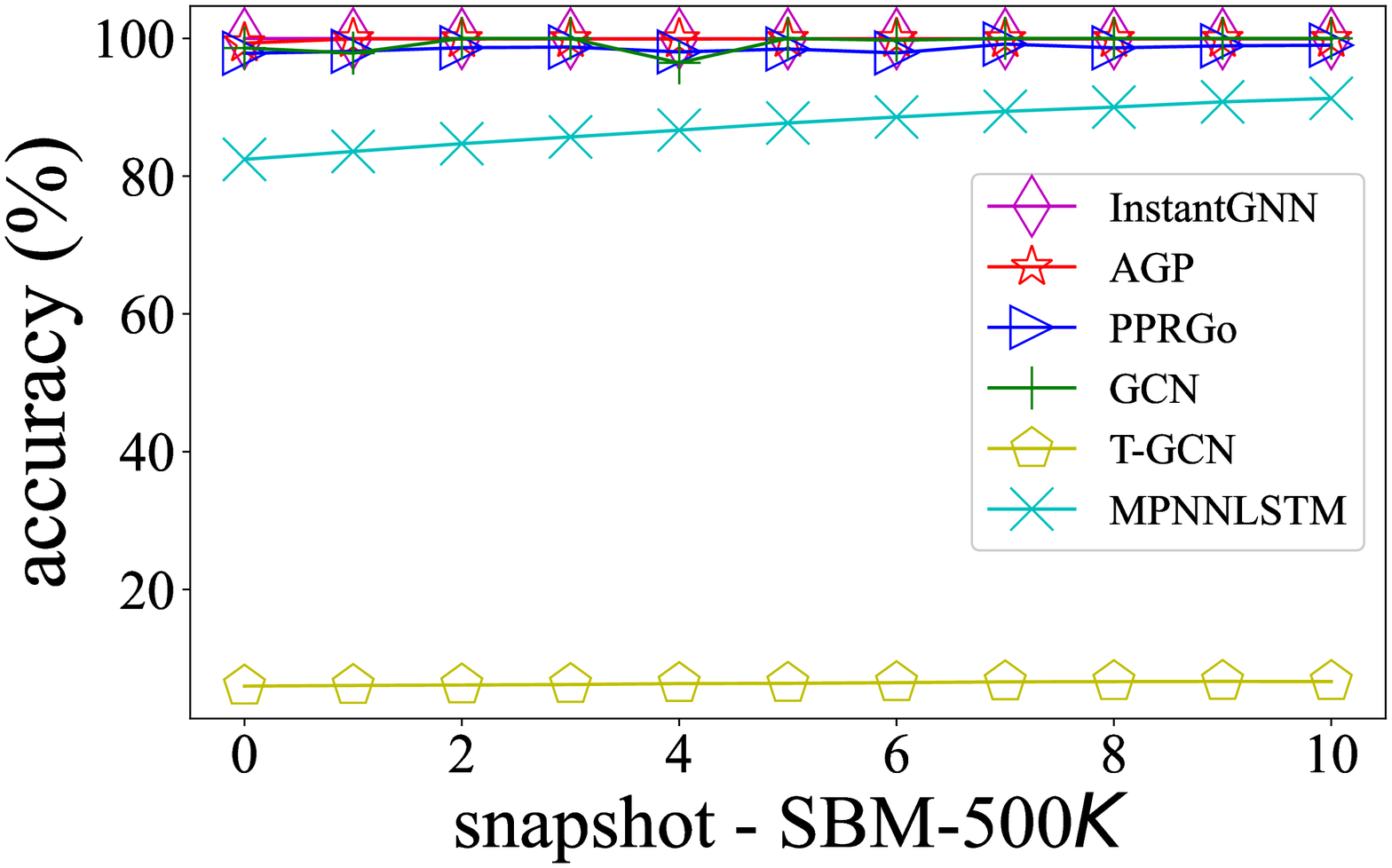} &
			 \hspace{-3mm}\includegraphics[height=28mm]{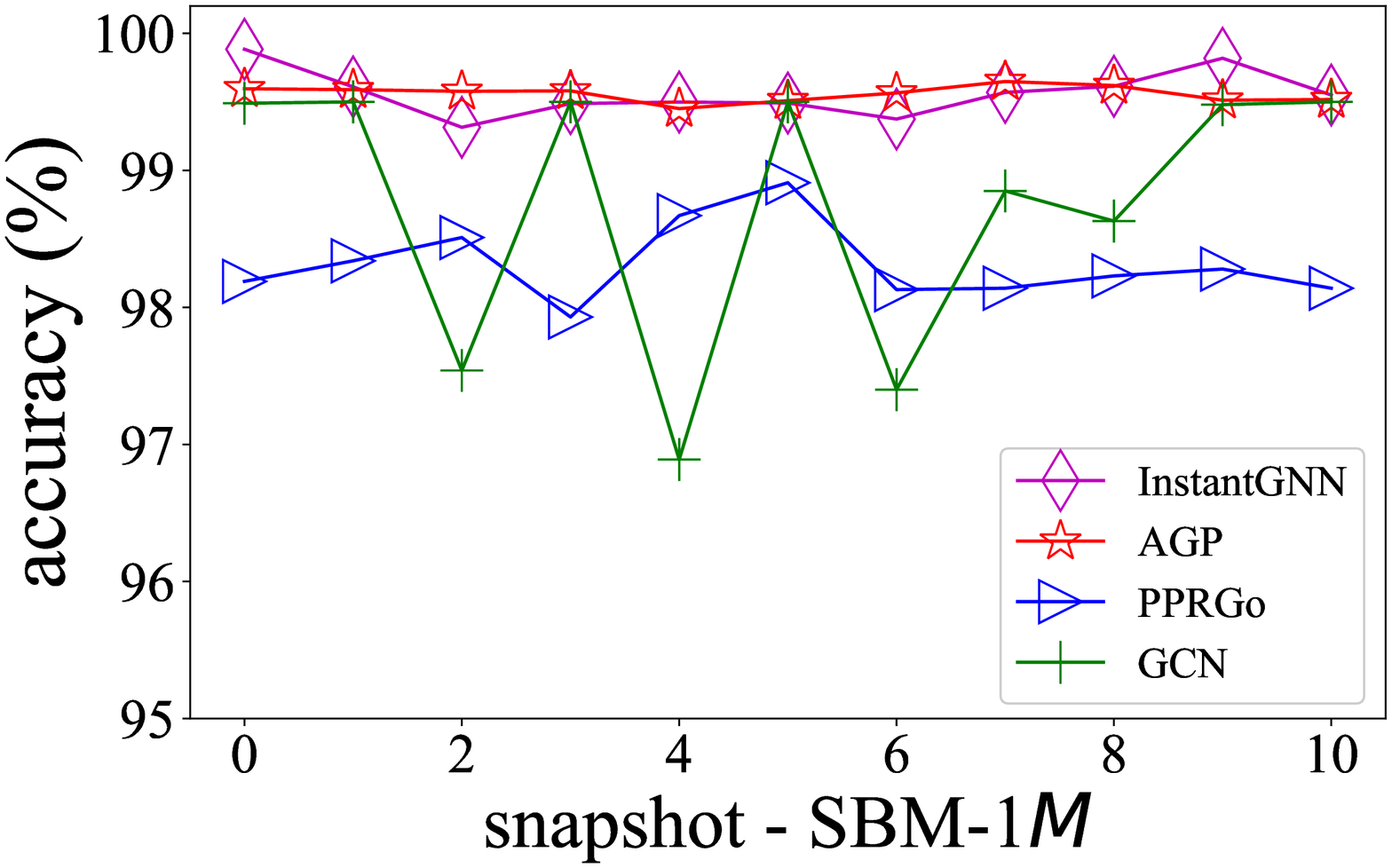} &
			 \hspace{-3mm}\includegraphics[height=28mm]{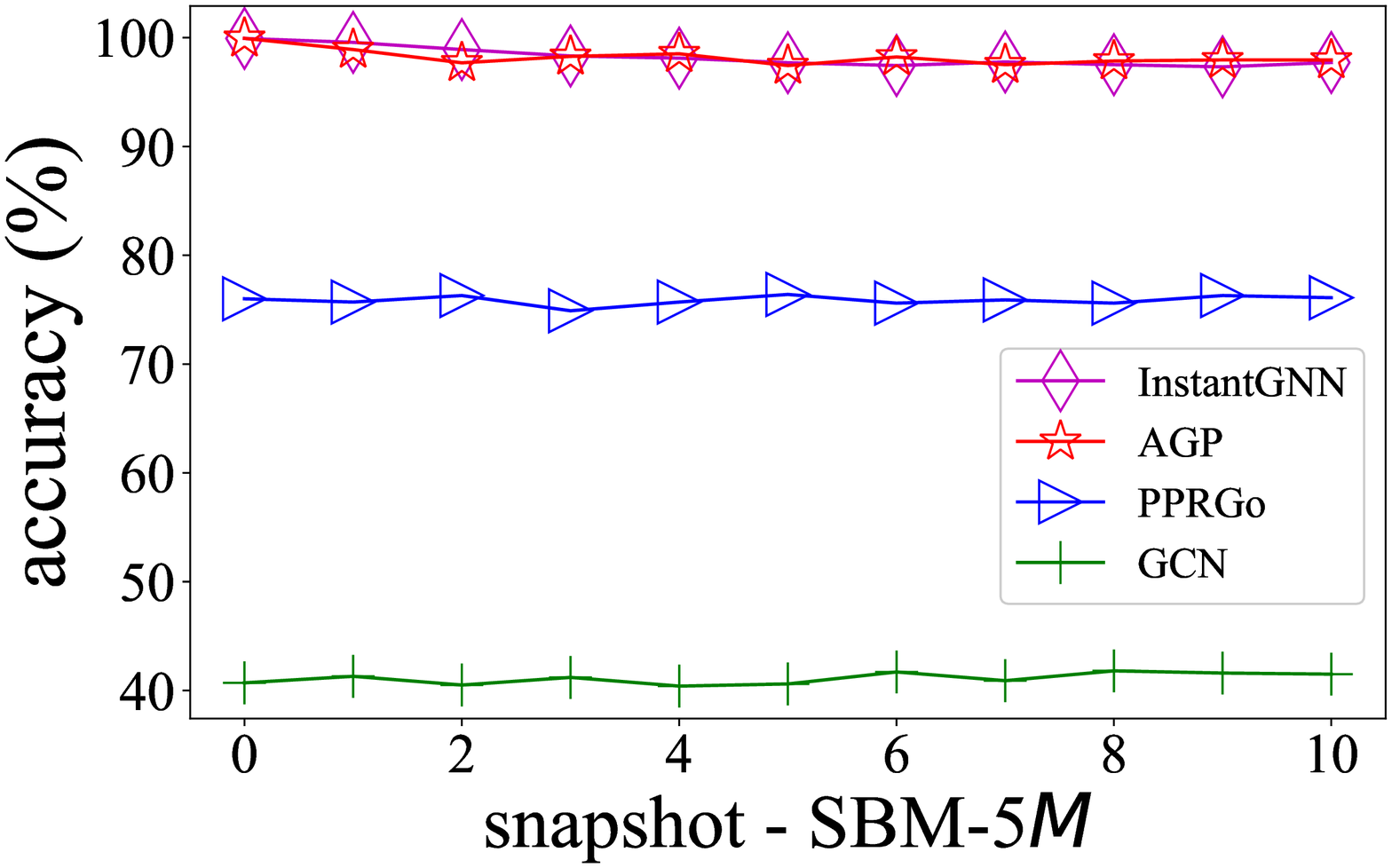} &
			 \hspace{-3mm}\includegraphics[height=28mm]{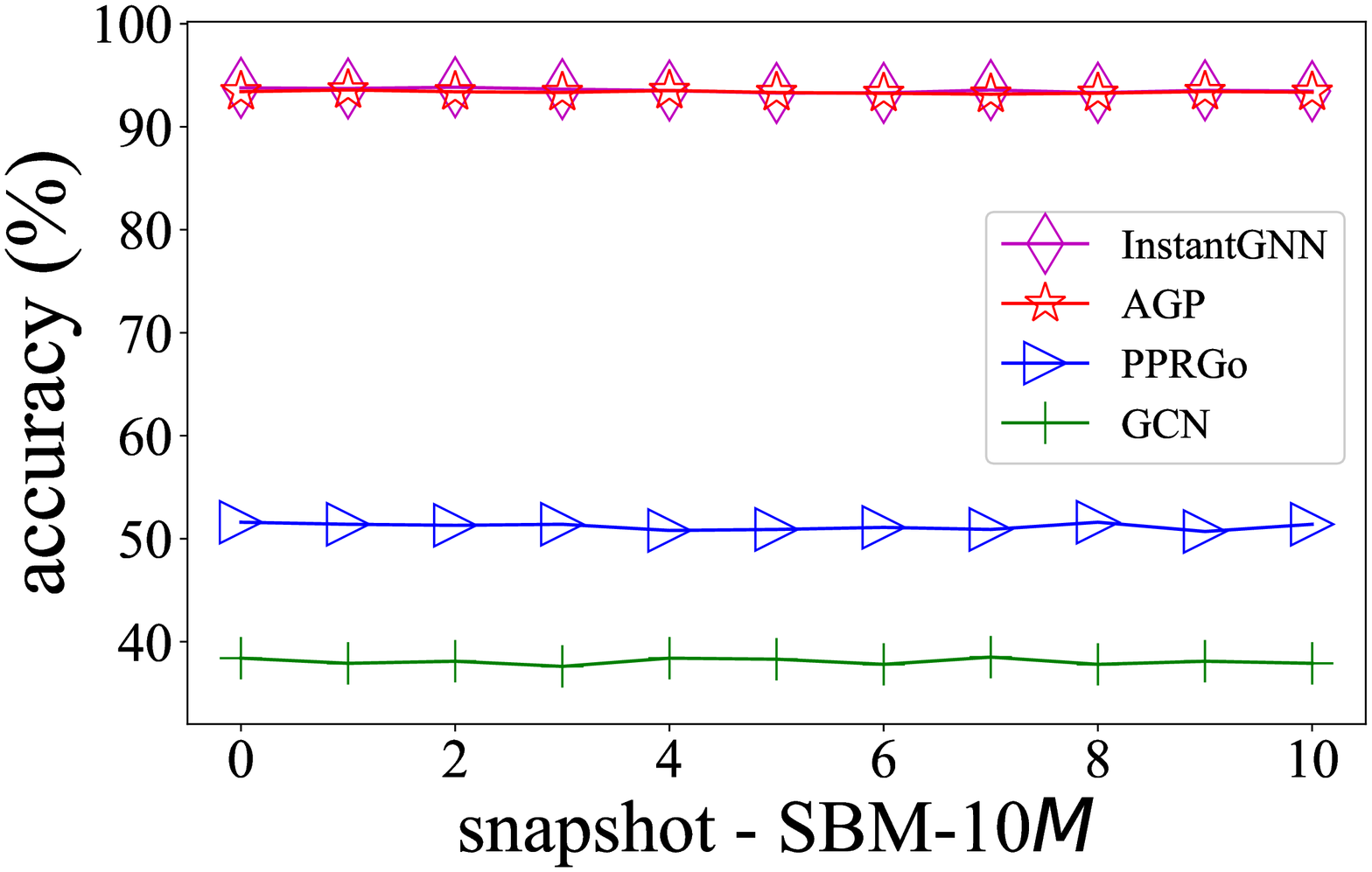}
		\end{tabular}
		\vspace{-2mm}
		\caption{Comparison of accuracy on SBM datasets.}
		\label{fig:sbm_acc}
	\end{small}
\end{figure*}
\begin{figure*}[t]
\setlength{\abovecaptionskip}{1mm}
\setlength{\belowcaptionskip}{-2mm}
	\begin{small}
		\centering
		\vspace{-2mm}
		\begin{tabular}{cccc}
			 \hspace{-2mm}\includegraphics[height=28mm]{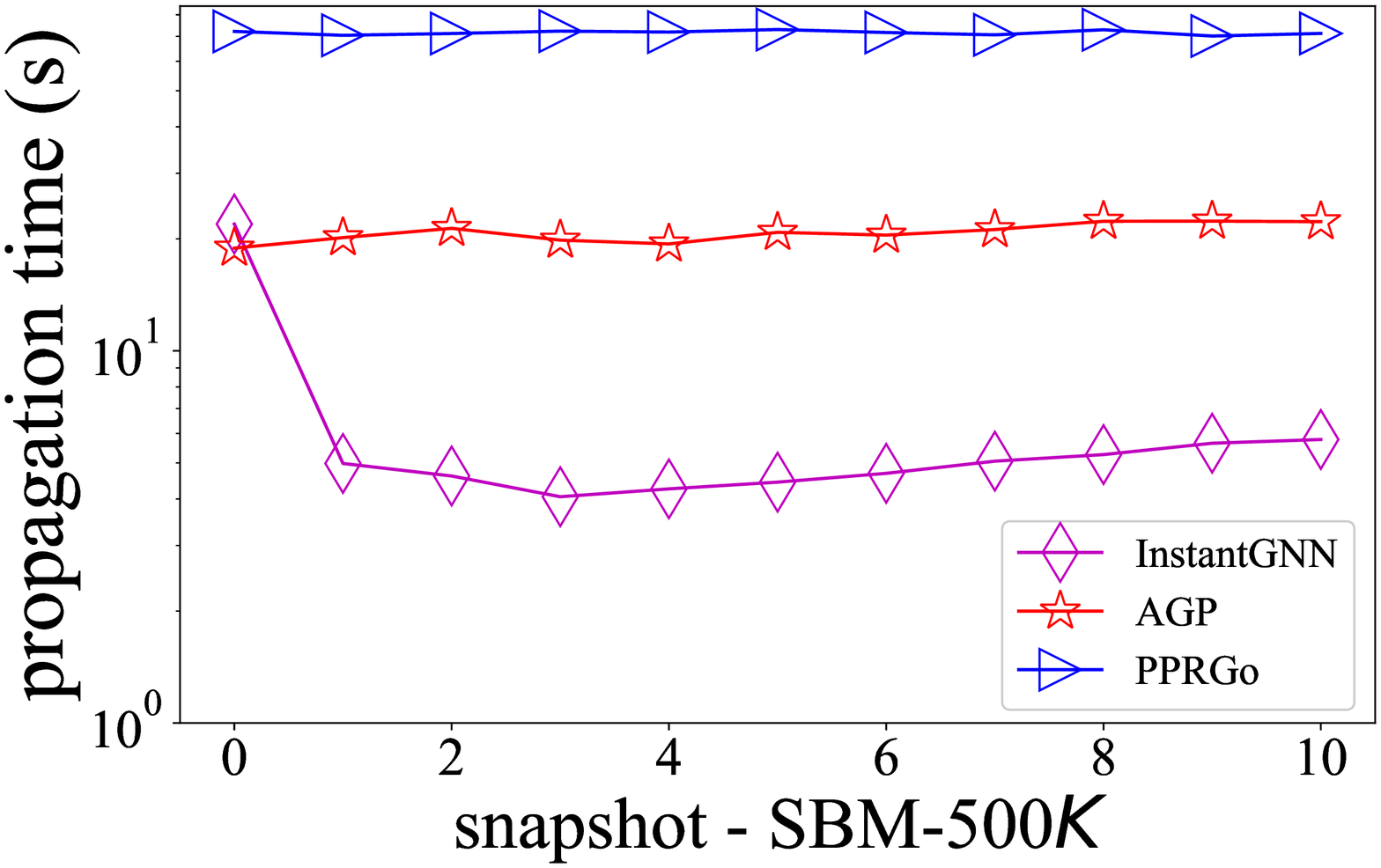} &
			 \hspace{-2.5mm}\includegraphics[height=28mm]{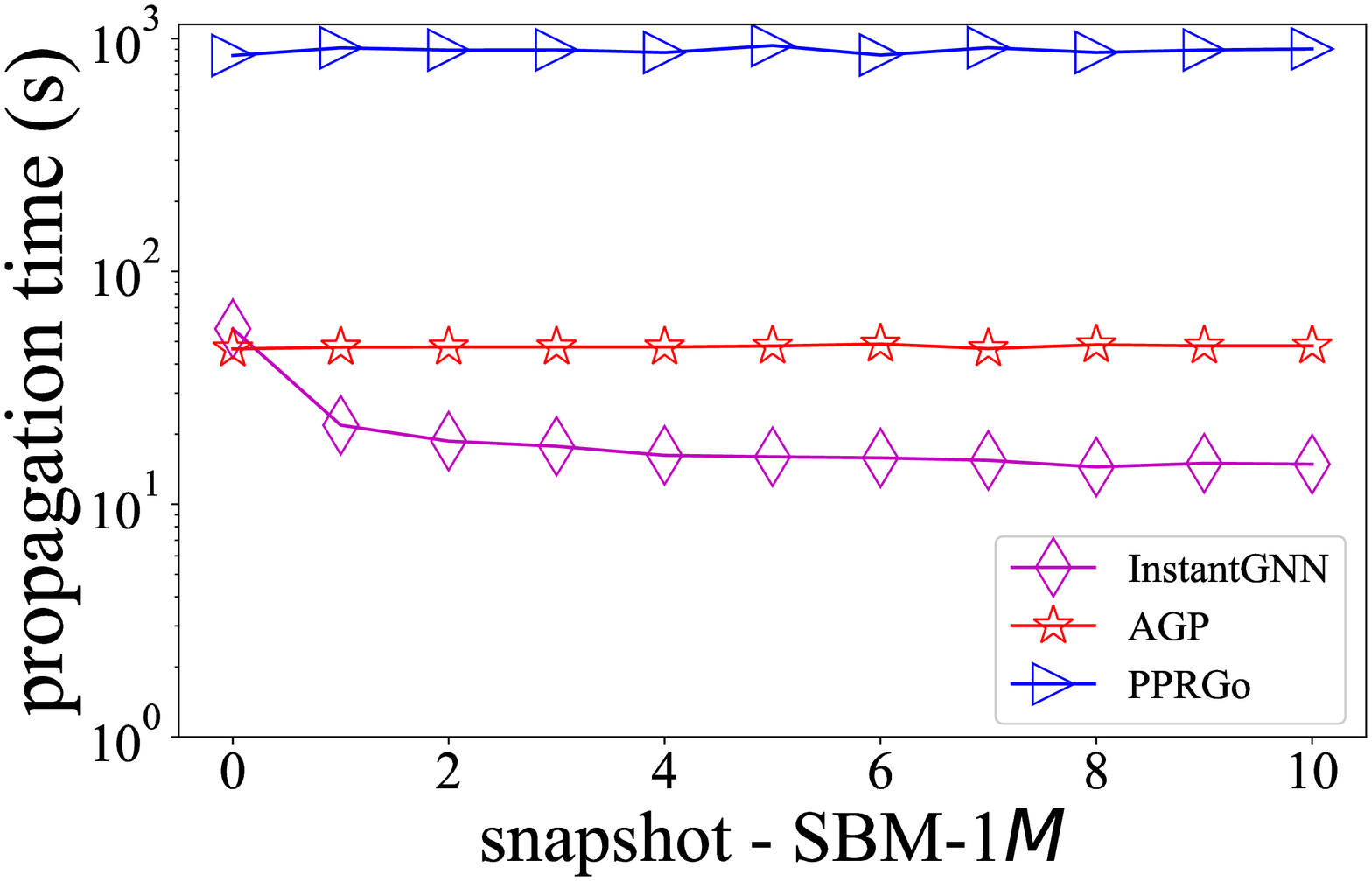} &
			 \hspace{-2.5mm}\includegraphics[height=28mm]{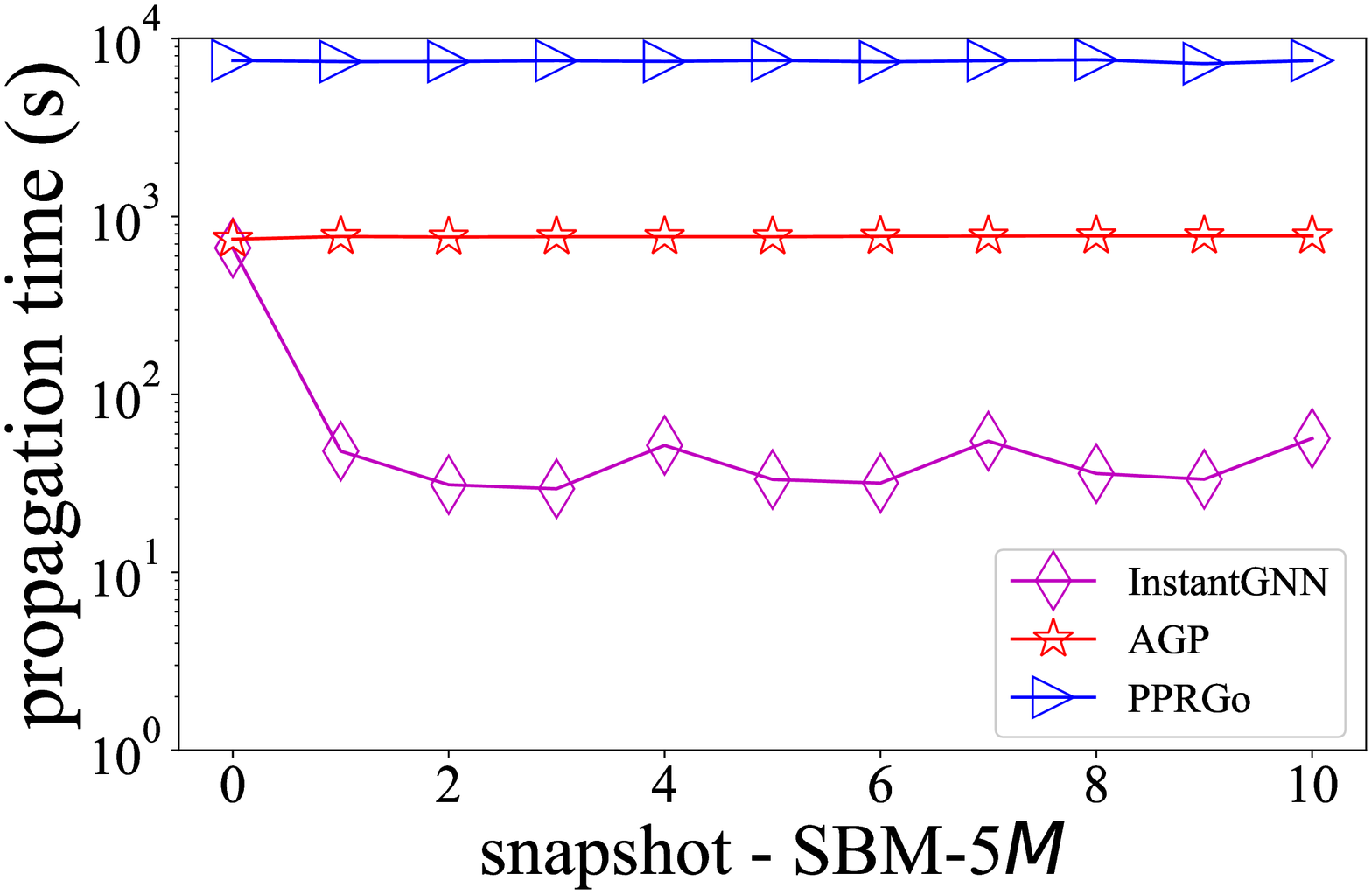} &
			 \hspace{-2.5mm}\includegraphics[height=28mm]{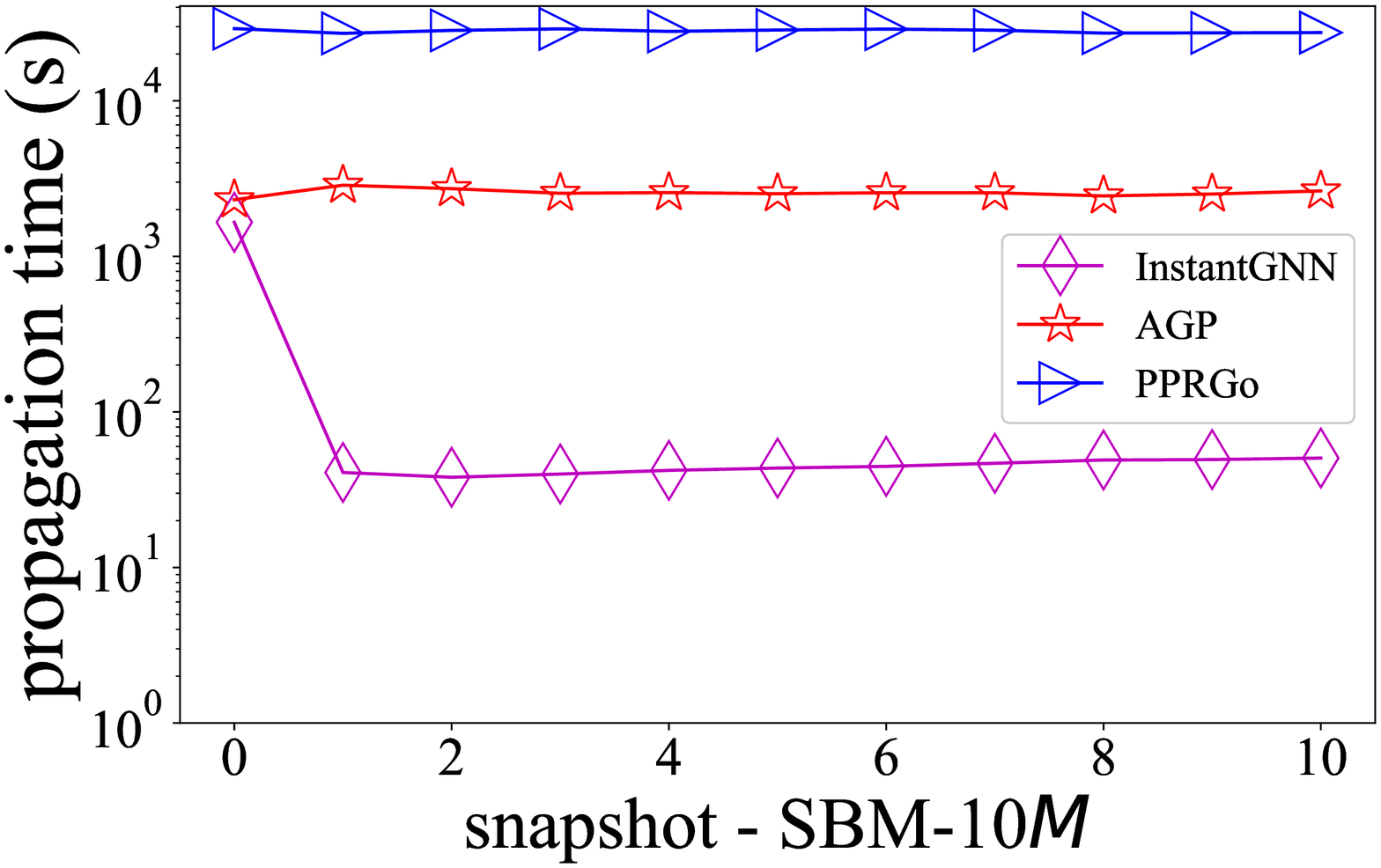}
		\end{tabular}
		\vspace{-2mm}
		\caption{Comparison of propagation time on SBM datasets.}
		\label{fig:sbm_proptime}
		\vspace{-2mm}
	\end{small}
\end{figure*}
On Arxiv, Products, and Papers100M, we run experiments using the basic experimental configuration of OGB~\cite{hu2020ogb} and the dynamic setting mentioned in Section~\ref{sec:datasets_baseline}. Particularly, nodes in these three datasets have static labels that would not change over time. 
We compare the accuracy and running time of different methods on each dataset. The time of InstantGNN, AGP and PPRGo can be divided into two parts: the time to obtain the propagation matrix (also known as preprocessing) and the time to train the neural network. Since we focus on reducing the propagation time on dynamic graphs, Figure~\ref{fig:arxiv_acc_time}-\ref{fig:papers100_acc_time} show the comparison results of accuracy and propagation time on Arxiv, Products and Papers100M, respectively. Overall, InstantGNN achieves orders-of-magnitude acceleration without compromising training accuracy, and advantages become even more apparent on large graphs. The results suggest that InstantGNN needs drastically less time to obtain the propagation matrix for graph at the current time, based on the empirical results. For instance, after obtaining the initialized propagation matrix, InstantGNN only spends about 60 seconds updating the propagation matrix for the next snapshot on Papers100M. In comparison, AGP has to calculate the result from scratch and needs more than an hour. We also observe that it is challenging for T-GCN and MPNNLSTM to scale to large-scale graphs such as Products and Papers100M. To be more specific, they cannot complete one training epoch within 48 hours on Products and run out of 768GB memory on Papers100M. Figure~\ref{fig:avg_running_time} shows the average total running time on three datasets, and we can come to a similar conclusion as above.

\vspace{-2mm}
\subsection{Evaluation on CTDGs with Dynamic Labels}
\label{sec:dynamic_label}
In this section, we conduct experiments on Aminer and four synthetic datasets. The label of each node in these datasets at time $i$ is heavily influenced by its present attributes, which are contained in the current graph $G_i$. As a result, labels evolve over time, more closely resembling the situation in real-world applications. We split nodes into training, validation, test sets in a 70:20:10 ratio.
\noindent{\bf Results on Aminer.} 
Figure~\ref{fig:aminer_acc_time} shows the comparison in accuracy and propagation time of different methods on Aminer. We observe that InstantGNN can significantly reduce propagation time while achieving better performance. InstantGNN generates updated propagation matrices based on previous computation results, allowing for a 10$\times$ speedup over static approaches that require starting from scratch for graph signal propagation and neural network training. We also note that there is an upward trend of propagation time. Understandably, the raw statistical data is collected from 1984, when authors began to collaborate with each other. As the amount of collaboration grows over time, the number of edges grows as well, resulting in an increase in propagation time.

\noindent{\bf Results on SBM datasets.} We also evaluate InstantGNN on four SBM datasets with 500$K$, 1$M$, 5$M$, and 10$M$ nodes ($1K\hspace{-0.5mm}=\hspace{-0.5mm}10^3, 1M\hspace{-0.5mm}=\hspace{-0.5mm}10^6$), respectively. The graph events, which include insertions and deletions of edges, occur at random in these datasets, and the labels of nodes change correspondingly, as discussed in Section~\ref{sec:datasets_baseline}.\eat{ Since node features are noisy and graph structures are more beneficial for distinguishing nodes, we set the teleport probability $\alpha$ to be a small value, e.g., 0.01, for InstantGNN, AGP and PPRGo.} Figure~\ref{fig:sbm_acc} compares the accuracy of different methods. We observe that InstantGNN, AGP, PPRGo, GCN and MPNNLSTM can capture structural information of graphs even from random features, while T-GCN fails to convergence. Figure~\ref{fig:sbm_proptime} plots propagation time on every snapshot. A critical observation is that InstantGNN can achieve comparable accuracy to the strongest baseline AGP while requiring significantly less propagation time. More specifically, on the SBM-10$M$ dataset and snapshot $t>1$, InstantGNN reduces the propagation time by more than 50 times over AGP without sacrificing accuracy. Due to the excessive running time of PPRGo and GCN, on both SBM-5$M$ and SBM-10$M$ datasets, we tune their parameters to optimize the performance and ensure that propagation and training processes could be completed within 48 hours.

\vspace{-1mm}
\subsection{Evaluation of Adaptive Training} 
\label{sec:dynamictrain}
\begin{figure}[t]
\setlength{\abovecaptionskip}{1mm}
	\begin{small}
		\centering
		\begin{tabular}{cc}
			 \hspace{-4.3mm} \includegraphics[height=27.5mm]{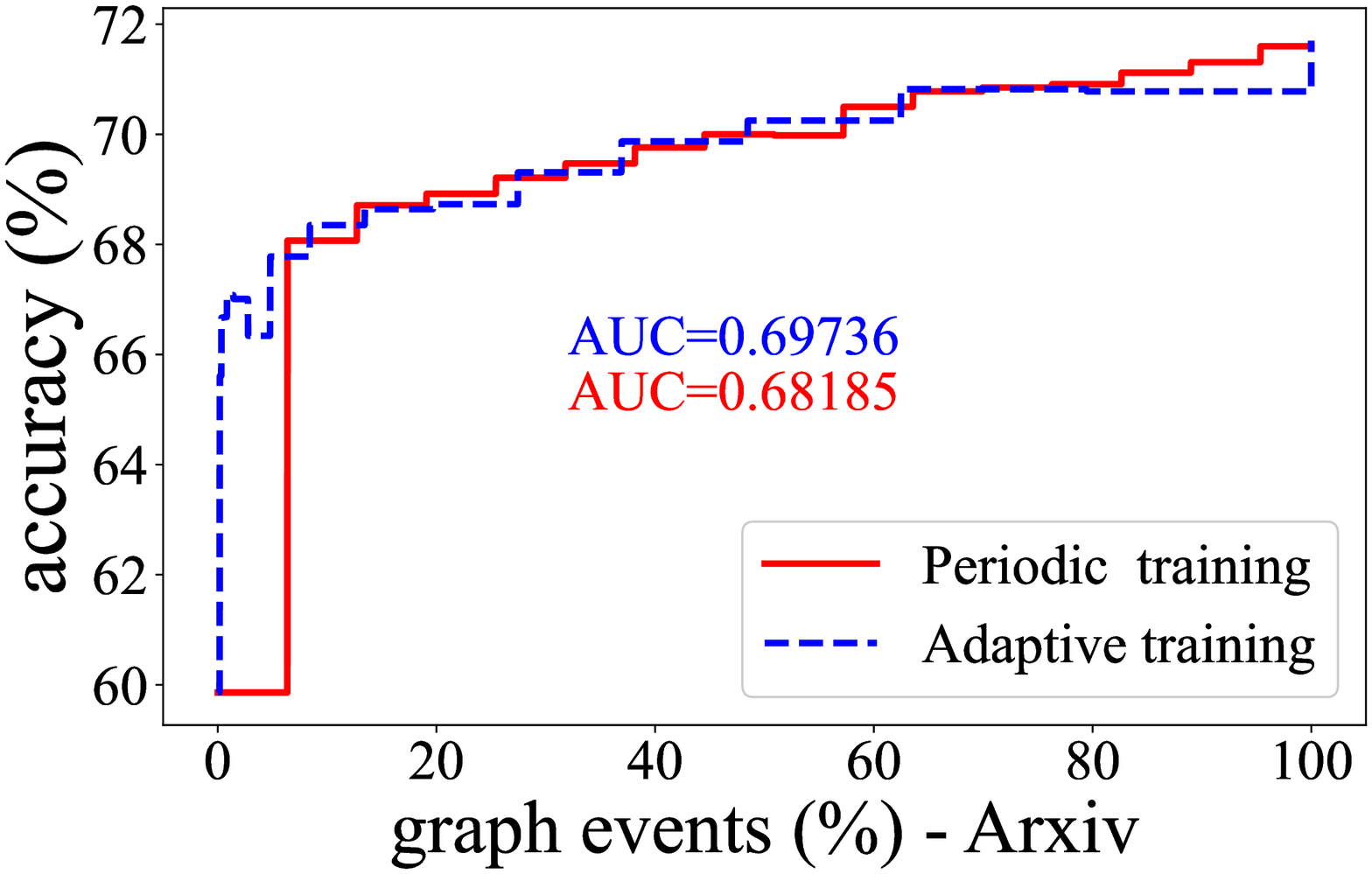} &
			 \hspace{-3.3mm} \includegraphics[height=27.5mm]{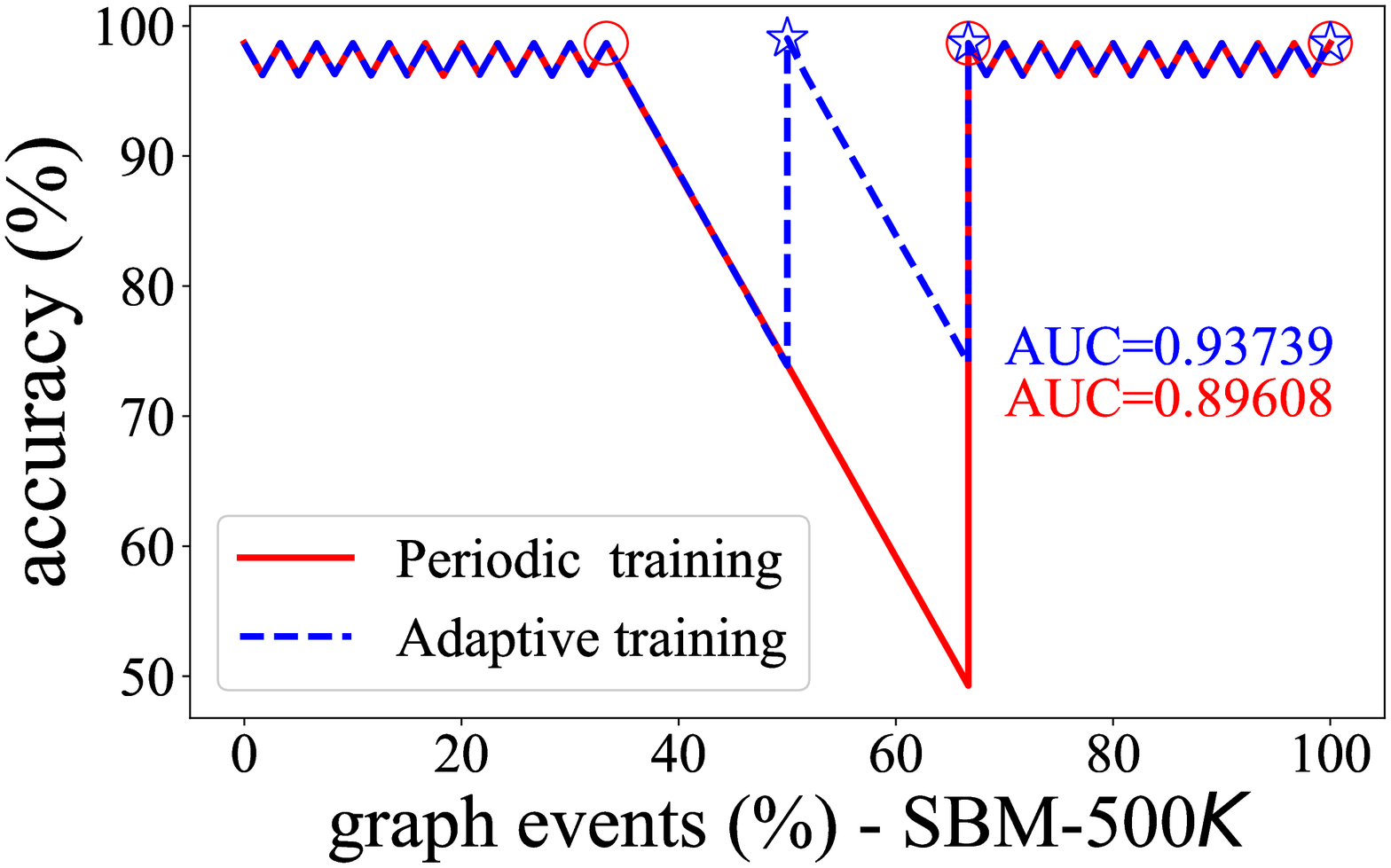} \\
		\end{tabular}
		\vspace{-2mm}
		\caption{Periodic training vs. Adaptive training.}
		\label{fig:adaptive_train}
		\vspace{-5mm}
	\end{small}
\end{figure}
We conduct experiments to demonstrate the effectiveness of \emph{Adaptive training} mentioned in Section~\ref{sec:Adaptive_Training}, details of the experimental setup can be found in the appendix. Setting the number of retraining times $|S|=16$, Figure~\ref{fig:adaptive_train} (left) shows the performance of \emph{Adaptive training} compared with \emph{Periodic training} on Arxiv, which has only insertion during the whole graph evolution and has been used in Section~\ref{sec:static_label}. It is evident that from the results, our method effectively identifies when it should retrain the model to obtain overall better results. As a result, quantitatively, we achieve a 1.5\% higher Area Under Curve (AUC) than \emph{Periodic training} over the entire course of the graph evolution.
We also evaluate Adaptive training on SBM-500$K$ with both inserted and deleted edges during the evolution, which better simulates real networks. \eat{We consider a special case where several nodes change their communities and recover immediately, and only a fraction of the changes are permanent. }Setting $|S|=3$, Figure~\ref{fig:adaptive_train} (right) plots the accuracy curves of \emph{Periodic training} and \emph{Adaptive training}, where hollow markers denote the moment of retraining. We observe that \emph{Adaptive training} significantly outperforms \emph{Periodic training} with a 4.1\% higher AUC.

%% file: conclusion.tex
\vspace{-2mm}
\section{Conclusion} 
\label{sec:conclusion}
This paper presents an instant graph neural network for dynamic graphs, which enables the incremental computation of the graph feature matrix, based on that from the previous step and a graph event. Compared with existing methods, InstantGNN avoids repeated and redundant computations when the graph is updated and enables instant prediction of node properties at the current moment. Furthermore, InstantGNN achieves adaptive training to maximize the model’s performance gains with a given training budget. It also deals with both CTDGs and DA-CTDGs. Extensive empirical evaluation demonstrates that InstantGNN is efficient for training on dynamic graphs while providing better performance. It truly enables online or instant graph neural networks for large-scale dynamic graphs with frequent updates.


%% file: batch_update.tex

\section{Batch Update}
\label{sec:batch_update}
We show how to construct estimated vectors for dynamic graphs with one-by-one updated edges in Algorithm~\ref{alg:2}. We adapt our approach in this section to support parallel processing and batch update settings. We begin with an equivalent approach of updating a single node. That is, we can modify the estimate of a node and so avoid repeatedly updating residuals on its neighbors. Taking the insertion of edge $e=(u,v)$ as an example, as described in Section~\ref{sec:instant2structural}, changes in the degrees of nodes $u$ and $v$ have an effect on their neighbors to maintain their equations hold. The primary cause is the transition of $\frac{\bm{\hat{\pi}}(u)}{d(u)^{1-\beta}}$ to $\frac{\bm{\hat{\pi}}(u)}{(d(u)+1)^{1-\beta}}$. As stated in line 5-6 and 9-19 of Algorithm~\ref{alg:2}, we add the increment $\bm{\hat{\pi}}(u)\cdot(\frac{1}{(d(u)+1)^{1-\beta}} - \frac{1}{d(u)^{1-\beta}})$ and $\bm{\hat{\pi}}(v)\cdot(\frac{1}{(d(v)+1)^{1-\beta}} - \frac{1}{d(v)^{1-\beta}})$ to the residuals of each node $w\in N(u)$ and $y\in N(v)$. To avoid modifying all neighbors of nodes $u$ and $v$, we let $\bm{\hat{\pi}}^{\prime}(u) = \bm{\hat{\pi}}(u)\cdot \frac{(d(u)+1)^{1-\beta}}{d(u)^{1-\beta}}$, where $\bm{\hat{\pi}}^{\prime}(u)$ denotes the updated estimate of node $u$. As a result, we have $\frac{\bm{\hat{\pi}}^{\prime}(u)}{(d(u)+1)^{1-\beta}}=\frac{\bm{\hat{\pi}}(u)}{d(u)^{1-\beta}}$, and the equation for node $w\in N(u)$ remains valid. Additionally, the residual of node $u$ needs to be adjusted to account for the change in $\bm{\hat{\pi}}(u)$, i.e., $\Delta \bm{r}(u)=-\frac{\bm{\hat{\pi}}^{\prime}(u)-\bm{\hat{\pi}}(u)}{\alpha}=\bm{\hat{\pi}}^{\prime}(u)\cdot \frac{d(u)^{1-\beta}-(d(u)+1)^{1-\beta}}{\alpha\cdot(d(u)+1)^{1-\beta}}$. Let $\P(w, u)$ be the element of the $w$-th row and $u$-th column of transition matrix $\P$. Note that $\Delta \bm{r}(u) \cdot (1-\alpha)\P(w, u) = \Delta \bm{r}(w)$ (Equation 6), indicating that $\Delta \bm{r}(u)$ can be transformed into $\Delta \bm{r}(w)$ following another push operation. This demonstrates the equivalent nature of these two operations from another perspective. We omit details about updating node $v$ since it is handled similarly. The advantage of modifying $\bm{\hat{\pi}}$ is that we need only update the nodes $u$ and $v$ when edge $e=(u,v)$ arrives, simplifying subsequent parallel processing.

Here we focus on the dynamic propagation algorithm under batch update setting, and Algorithm~\ref{alg:batch_update} gives the pseudo-code. The main idea is to simultaneously calculate the increments caused by multiple edges. Given a set of graph events $M=\{Event_1,\dots,Event_k\}$, line 3 corresponds to the graph updating stage, where the graph in time step $k$, $G_k$, is generated from the initial graph guided by $M$, and $k$ denotes the batch size. Lines 5-9 represent the first step of updating the affected nodes, in which we update the estimates and residuals of these nodes in the way aforementioned. Recall that a portion of an affected node's increment is caused by its added or deleted neighbors. Therefore, we compute the increment by iterating over all new and vanishing neighbors of each affected node, as in lines 10-17. Line 18 invokes the basic propagation algorithm to reduce the error, which is the same as line 7 in Algorithm~\ref{alg:2}.

\begin{algorithm}[t]
\setlength{\textfloatsep}{2cm}
\setlength{\floatsep}{2cm}
\caption{Parallel Dynamic Propagation Algorithm for Batch Update.}\label{alg:batch_update}
\SetKwInOut{Input}{Input}
\SetKwInOut{Output}{Output}
\Input{Dynamic graph $\{G_0, M=\{Event_1,\dots,Event_k \}\}$, graph signal vector $\bm{x}$, error threshold $\varepsilon$, teleport probability $\alpha$, convolutional coefficient $\beta$}
\Output{estimated propagation vector $\bm{\hat{\pi}}$,\newline residual vector $\bm{r}$}
$\bm{\hat{\pi}}\leftarrow \bm{0}, \bm{r}\leftarrow\bm{x}$\;
$\bm{\hat{\pi}}, \bm{r}\leftarrow Basic Propagation(G_0, \varepsilon, \alpha, \beta, \bm{\hat{\pi}}, \bm{r})$\;
Generate $G_k$ from updating $G_0$ by $M$\;
$V_{_{affected}} \leftarrow \{u|$ the degree of node $u$ has changed$\}$\; 
\textbf{parallel} \For{each $u\in V_{_{affected}}$}
{
  $\Delta d(u) \leftarrow$ the degree to which node $u$ has changed \;
  $\bm{\hat{\pi}}(u) \leftarrow \bm{\hat{\pi}}(u) \cdot \frac{d(u)^{1-\beta}}{(d(u)-\Delta d(u))^{1-\beta}}$ \;
  $\Delta \bm{r}_{_1}(u) \leftarrow \bm{\hat{\pi}}(u) \cdot \frac{(d(u)-\Delta d(u))^{1-\beta} - d(u)^{1-\beta}}{\alpha \cdot d(u)^{1-\beta}} $ \;
  $\bm{r}(u) \leftarrow \bm{r}(u)+\Delta \bm{r}_{_1}(u)$ \;
}
\textbf{parallel} \For{each $u\in V_{_{affected}}$}
{
  $\Delta \bm{r}_{_2}(u) \leftarrow (\bm{\hat{\pi}}(u)\hspace{-0.5mm}+\hspace{-0.5mm}\alpha \bm{r}(u)\hspace{-0.5mm}-\hspace{-0.5mm}\alpha \bm{x}(u)) \hspace{-0.5mm}\cdot\hspace{-0.5mm} \frac{(d(u)-\Delta d(u))^\beta - d(u)^\beta}{d(u)^\beta} $ \;
  \For{each $v \in \{$the added neighbors of node $u\}$}
  {
    $\Delta \bm{r}_{_2}(u) \leftarrow \Delta \bm{r}_{_2}(u) + \frac{(1-\alpha)\bm{\hat{\pi}}(v)}{d(u)^\beta d(v)^{1-\beta}}$ \;
  }
  \For{each $v \in \{$the deleted neighbors of node $u\}$}
  {
    $\Delta \bm{r}_{_2}(u) \leftarrow \Delta \bm{r}_{_2}(u) - \frac{(1-\alpha)\bm{\hat{\pi}}(v)}{d(u)^\beta d(v)^{1-\beta}}$ \;
  }
  $\Delta \bm{r}_{_2}(u) \leftarrow \Delta \bm{r}_{_2}(u) / \alpha $ \;
  $\bm{r}(u) \leftarrow \bm{r}(u)+\Delta \bm{r}_{_2}(u)$ \;
}
$\bm{\hat{\pi}},\bm{r}\leftarrow Basic Propagation(G_k, \varepsilon, \alpha, \beta, \bm{\hat{\pi}}, \bm{r})$\;
return $\bm{\hat{\pi}}$, $\bm{r}$\;
\end{algorithm}

\noindent\textbf{Example.} The following toy example with k=2 demonstrates that the increment in Algorithm 3 is equivalent to the cumulative increase calculated one by one. Let $M=\{ (e_1=(u, v), InsertEdge), (e_2=(u, y), InsertEdge) \}$, we first compute the increments on node $u$ using Algorithm~\ref{alg:2} step by step. The main difference is that we update the estimate of node $u$ rather than iteratively updating its neighbors. Inserting $e_1=(u, v)$, the update at node $u$ is:
\begin{small}
\begin{align}
\label{equ:update_(u,v)}
\bm{\hat{\pi}}^{(u, v)}(u)&=\bm{\hat{\pi}}(u) \cdot \frac{(d(u)+1)^{1-\beta}}{d(u)^{1-\beta}}, \nonumber \\ 
\Delta \bm{r}_{1}^{(u, v)}(u)\eat{=-\frac{\bm{\hat{\pi}}^{(u, v)}(u)-\bm{\hat{\pi}}(u)}{\alpha}}&=\bm{\hat{\pi}}^{(u, v)}(u) \cdot \frac{d(u)^{1-\beta}-(d(u)+1)^{1-\beta}}{\alpha \cdot(d(u)+1)^{1-\beta}}, \nonumber \\
\bm{r}_{\text {temp }}^{(u, v)}(u)&=\bm{r}(u)+\Delta \bm{r}_{1}^{(u, v)}(u), \\
\Delta \bm{r}_{2}^{(u, v)}(u)&=\frac{1}{\alpha}\left(\left(\bm{\hat{\pi}}^{(u, v)}(u)+\alpha \bm{r}_{\text {temp }}^{(u, v)}(u)-\alpha \bm{x}(u)\right) \cdot \left(\frac{d(u)^{\beta}}{(d(u)+1)^{\beta}}-1\right)
\right. \nonumber\\
&+ \left. \frac{(1-\alpha) \bm{\hat{\pi}}(v)}{(d(u)+1)^{\beta} d(v)^{1-\beta}} \nonumber \right),\\
\bm{r}^{(u, v)}(u)&=\bm{r}_{\text {temp }}^{(u, v)}(u)+\Delta \bm{r}_{2}^{(u, v)}(u) \nonumber ,
\end{align}
\end{small}
where $\bm{\hat{\pi}}^{(u, v)}(u)$ and $\bm{r}^{(u, v)}(u)$ denote the estimate and residual of node $u$ after inserting edge $e_1=(u,v)$. For notational convenience, we denote the residual caused by updating estimates as $\Delta \bm{r}_{1}^{(u, v)}(u)$, $\bm{r}_{\text{temp}}^{(u, v)}(u)$ is an intermediate result, and $\Delta \bm{r}_{2}^{(u, v)}(u)$ represents the mass that triggered by the new neighbors. Then, the edge $e_2=(u,y)$ arrives, and the node $u$ is updated as follows.
\begin{small}
\begin{align}
\label{equ:update_(u,y)}
\bm{\hat{\pi}}^{(u, y)}(u)&=\bm{\hat{\pi}}^{(u, v)}(u) \cdot \frac{(d(u)+2)^{1-\beta}}{(d(u)+1)^{1-\beta}}, \nonumber \\
\Delta \bm{r}_{1}^{(u, y)}(u)&=\bm{\hat{\pi}}^{(u, y)}(u) \cdot \frac{(d(u)+1)^{1-\beta}-(d(u)+2)^{1-\beta}}{\alpha \cdot(d(u)+2)^{1-\beta}}, \nonumber \\
\bm{r}_{\text {temp }}^{(u, y)}(u)&=\bm{r}^{(u, v)}(u)+\Delta \bm{r}_{1}^{(u, y)}(u), \\
\Delta \bm{r}_{2}^{(u, y)}(u)&=\frac{1}{\alpha}\left(\left(\bm{\hat{\pi}}^{(u, y)}(u)+\alpha \bm{r}_{\text {temp }}^{(u, y)}(u)-\alpha \bm{x}(u)\right) \cdot\left(\frac{(d(u)+1)^{\beta}}{(d(u)+2)^{\beta}}-1\right) \right.\nonumber \\
&+ \left. \frac{(1-\alpha) \bm{\hat{\pi}}(y)}{(d(u)+2)^{\beta} d(y)^{1-\beta}}\right), \nonumber \\
\bm{r}^{(u, y)}(u)&=\bm{r}_{\text {temp }}^{(u, y)}(u)+\Delta \bm{r}_{2}^{(u, y)}(u) \nonumber .
\end{align}
\end{small}

Algorithm~\ref{alg:batch_update} simultaneously calculates the increment induced by the insertion of edge $e_1=(u, v)$ and edge $e_2=(u, y)$ for node $u$:
\begin{small}
\begin{align}
\bm{\hat{\pi}}^{\prime}(u)&=\bm{\hat{\pi}}(u) \cdot \frac{(d(u)+2)^{1-\beta}}{d(u)^{1-\beta}}, \nonumber \\
\Delta \bm{r}_{1}(u)& =\bm{\hat{\pi}}^{\prime}(u) \cdot \frac{d(u)^{1-\beta}-(d(u)+2)^{1-\beta}}{\alpha \cdot(d(u)+2)^{1-\beta}}, \nonumber \\
\bm{r}_{\text {temp }}(u)&=\bm{r}(u)+\Delta \bm{r}_{1}(u), \\
\Delta \bm{r}_{2}(u)&=\frac{1}{\alpha}\left(\left(\bm{\hat{\pi}}^{\prime}(u)+\alpha \bm{r}_{\text {temp }}(u)-\alpha \bm{x}(u)\right) \cdot \left(\frac{d(u)^{\beta}}{(d(u)+2)^{\beta}}-1\right) \right.\nonumber \\
&+ \left. \frac{(1-\alpha) \bm{\hat{\pi}}(v)}{(d(u)+2)^{\beta} d(v)^{1-\beta}}+\frac{(1-\alpha) \bm{\hat{\pi}}(y)}{(d(u)+2)^{\beta} d(y)^{1-\beta}}\right), \nonumber \\
\bm{r}^{\prime}(u)&=\bm{r}_{\text {temp}}(u)+\Delta \bm{r}_{2}(u) \nonumber,
\end{align}
\end{small}
where $\bm{\hat{\pi}}^{\prime}(u)$ and $\bm{r}^{\prime}(u)$ denote the estimate and residual of node $u$ after updating, and $\bm{r}_{\text {temp }}(u)$ is the intermediate result. Substituting Equation~\ref{equ:update_(u,v)} and \ref{equ:update_(u,y)}, we found that $\bm{\hat{\pi}}^{\prime}(u)=\bm{\hat{\pi}}^{(u, y)}(u)$, $\Delta \bm{r}_{1}(u)=\Delta \bm{r}_{1}^{(u, v)}(u)+\Delta \bm{r}_{1}^{(u, y)}(u)$, and $\Delta \bm{r}_{2}(u)=\Delta \bm{r}_{2}^{(u, v)}(u)+\Delta \bm{r}_{2}^{(u, y)}(u)$. It is easy to obtain $\bm{r}^{\prime}(u)=\bm{r}^{(u, y)}(u)$. Therefore, the final results of computations performed together and separately are identical.

%% file: error_analysis.tex
\section{Error Analysis}
To analyze the error bound, we first note that Algorithm~\ref{alg:1} is applied to obtain the estimate vector $\bm{\hat{\pi}}_i$ and residual vector $\bm{r}_i$ corresponding to $G_i$ after computing increments, as line 7 in Algorithm~\ref{alg:2}. This means that the error adjustment of $\bm{\hat{\pi}}_i$ is actually performed by Algorithm~\ref{alg:1}. Therefore, the dynamic propagation method can preserve the result with the same error for each time step. Given the error threshold $\varepsilon$, we use the following theorem to derive the error bound of our method.
\begin{theorem}[Error Analysis]
\vspace{-1mm}
\label{theorem:error}
Let $\bm{\pi}_i(s)$ be the propagation vector of the graph $G_i$, and $d_i(s)$ is the degree of node $s$ at time $i$. For each node $s\in V$, Algorithm~\ref{alg:2} computes estimators $\bm{\hat{\pi}}_i(s)$ such that $|\bm{\hat{\pi}}_i(s)- \bm{\pi}_i(s)|\leq\varepsilon \cdot d_i(s)^{1-\beta}$ holds for $\forall i\in \{0, 1, 2, \dots, k\}$.
\vspace{-1mm}
\end{theorem}

%% file: details_exp.tex
\section{Experimental Details}
\subsection{Details of data pre-processing}
\begin{table*}[t]
  \caption{Parameters of generating SBM datasets.}
  \label{tab:gen_sbm}
  \vspace{-3mm}
  \begin{tabular}{l|c|c|c|c|c|c} \hline
    {\bf Dataset} & {\bf \#nodes} & {\bf \#classes} & {\bf \#intra-block edges} & {\bf \#inter-block edges} & {\bf \#snapshots} & {\bf \#nodes changed per snapshot} \\ \hline
    SBM-500$K$ & 500,000 & 50 & 20 & 1 & 10 & 2500 \\
    SBM-1$M$ & 1,000,000 & 50 & 20 & 1 & 10 & 2500 \\
    SBM-5$M$ & 5,000,000 & 100 & 20 & 1 & 10 & 2500 \\
    SBM-10$M$ & 10,000,000 & 100 & 20 & 1 & 10 & 2500 \\ \hline
\end{tabular}
\vspace{-3mm}
\end{table*}
\begin{table*}[t]
  \caption{Parameters of InstantGNN.}
  \label{tab:param}
  \vspace{-3mm}
  \begin{tabular}{l|c|c|c|c|c|c|c} \hline
    {\bf Dataset} & \hspace{3mm}{$\boldsymbol{\beta}$}\hspace{3mm} & \hspace{3mm}{$\boldsymbol{\alpha}$}\hspace{3mm} & \hspace{5mm}{$\boldsymbol{\varepsilon}$}\hspace{5mm} & \hspace{5mm}{\bf learning rate}\hspace{5mm} & \hspace{5mm}{\bf batch size}\hspace{5mm} & {\bf dropout} & {\bf hidden size} \\ \hline
    Arxiv & 0.5 & 0.1 & $1e-7$ & 0.0001 & 10000 & 0.3 & 1024 \\
    Products & 0.5 & 0.1 & $1e-7$ & 0.0001 & 10000 & 0.2 & 1024 \\
    Papers100M & 0.5 & 0.2 & $1e-9$ & 0.0001 & 10000 & 0.3 & 2048 \\
    Aminer & 0.5 & 0.1 & $1e-7$ & 0.0001 & 1024 & 0.1 & 256 \\
    SBM-500$K$ & 0.5 & 0.001 & $1e-7$ & 0.01 & 1024 & 0.1 & 1024 \\
    SBM-1$M$ & 0.5 & 0.001 & $1e-8$ & 0.001 & 4096 & 0.1 & 1024 \\
    SBM-5$M$ & 0.5 & 0.001 & $1e-8$ & 0.001 & 8192 & 0.3 & 2048 \\
    SBM-10$M$ & 0.5 & 0.001 & $5e-9$ & 0.001 & 8192 & 0.3 & 2048 \\ \hline
\end{tabular}
\vspace{-3mm}
\end{table*}
{\bf OGB datasets.} According to \cite{wang2019attacking}, the absence of edges has a detrimental effect on the performance of GNNs, such as decreasing classification accuracy. In contrast, we suppose that accuracy would grow proportionately when these missing edges are gradually reinstated. We use three OGB datasets in different sizes: (1) {\bf Papers100M} is the largest publicly available dataset by far. \eat{It is a large-scale citation network with 111,059,956 nodes and 1,615,685,872 edges. Each node denotes a paper and each edge denotes the citation relationship between a pair of papers. }To imitate the graph's dynamic nature, we eliminated 40 million edges from the original dataset and reinserted them line by line later. (2) {\bf Arxiv} is also a citation network\eat{, which consists of 169,343 nodes and 1,166,243 directed edges}. It is constructed similarly to Papers100M, except that we remove about 950 thousand edges at the start. (3) {\bf Products} is a graph generated from the Amazon product co-purchasing network.\eat{ Here the nodes represent products and the edges are created between two nodes if they are purchased together. The complete data has 2,449,029 nodes and 61,859,012 edges.} We construct the dynamic graph in the same way\eat{ as Papers100M and Arxiv} except that the number of deleting edges is set to 30 million. We utilized the node features and labels provided by OGB~\cite{hu2020ogb}. \eat{According to the definition in Section 2.1, these graph is CTDGs. However, for better comparison with baselines, we obtain snapshots of the graph at fixed intervals, thus transforming them into DTDGs.}

\noindent{\bf Aminer.} In this dataset, time steps are defined as consecutive non-overlapping one-year periods. At each time step $i$, if an author had published papers collaborating with another, we create an edge between them in $G_i$, whose weight is ignored. With this definition, we obtain 1,267,755 – 4,025,865 edges, evolving from 1984 to 2014. Furthermore, we say an author belongs to a particular community if over half of his or her papers were published in corresponding journals this year. Recall that the community the author belongs to is used as the label.\eat{, which is defined by his/her publication of the year.} Therefore, if an author publishes in a different field each year, the label will change accordingly with the time step. In addition, we generate node features by utilizing the BERT-as-service\footnote{https://github.com/jina-ai/clip-as-service}\eat{~\cite{xiao2018bertservice}} to extract word vectors from the authors' descriptions.

\noindent{\bf SBM datasets.} We propose a study on synthetic graphs using a dynamical SBM. The SBM model will decide the relationship between nodes stochastically using probabilities based on the given community memberships. Given a graph $G_{i-1}$ with the label set $C_{i-1}$ at time step $i-1$, we generate the graph $G_i$ by migrating some of the nodes. Specifically, we randomly select $k$ nodes to form a subset of $V$, denoted as $\overline{V}$. For each node $s\in \overline{V}$, we assume it belongs to the community $y\in C_{i-1}$ at time step $i-1$. Then it turns to be a member of community $y^\prime \in \{C_i-y\}$ by removing the edges connected to the community $y$ and adding the connection to nodes of community $y^\prime$. 
Table~\ref{tab:gen_sbm} provides detailed parameters for generating the four SBM datasets we used in Section~\ref{sec:dynamic_label}.

\vspace{-2mm}
\subsection{More details of adaptive training}
\begin{figure}[t]
\setlength{\abovecaptionskip}{1mm}
	\begin{small}
		\centering
		\vspace{-1mm}
		\begin{tabular}{cc}
			 \hspace{-4mm} \includegraphics[height=27mm]{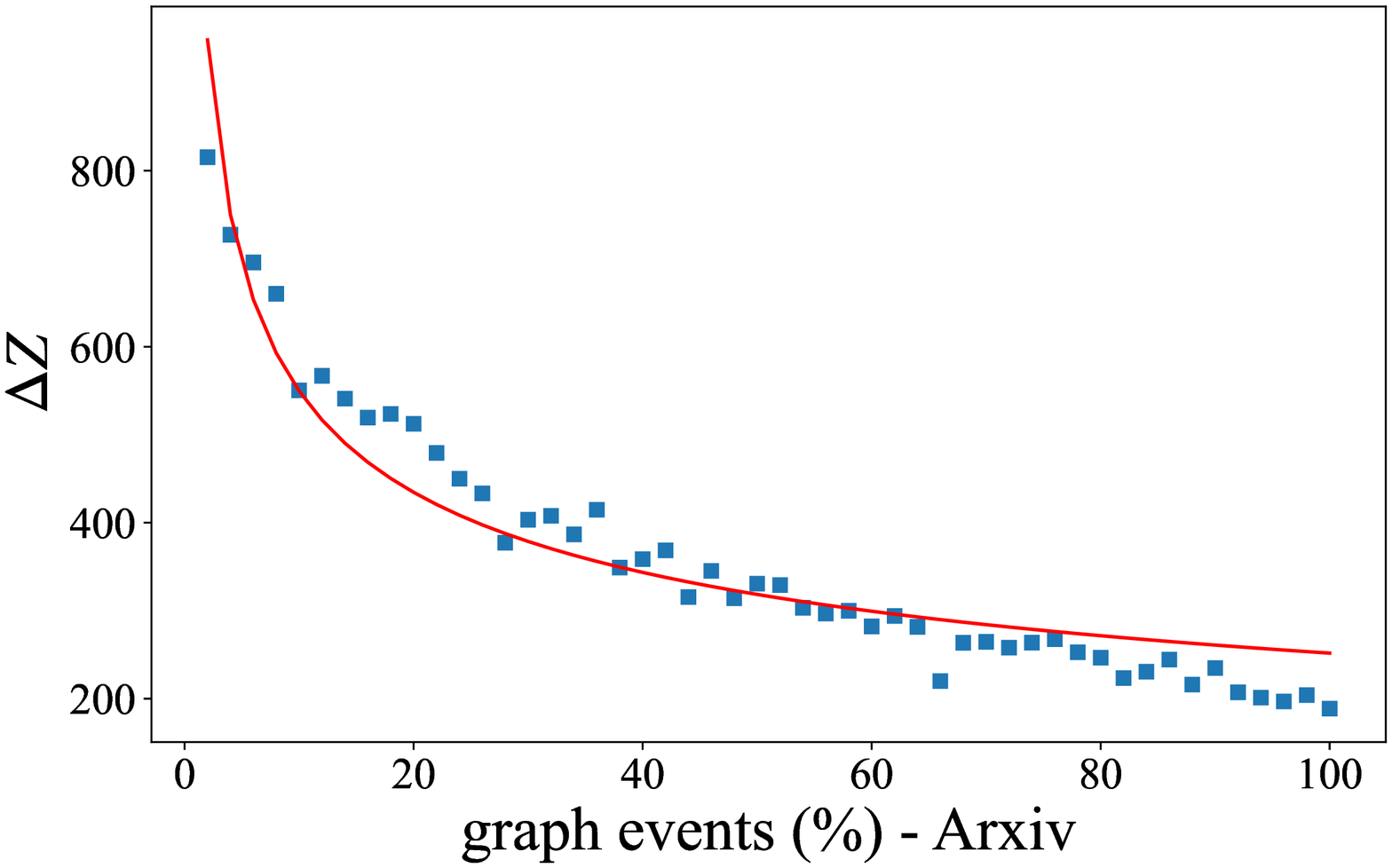} &
			 \hspace{-4mm} 
			 \includegraphics[height=27mm]{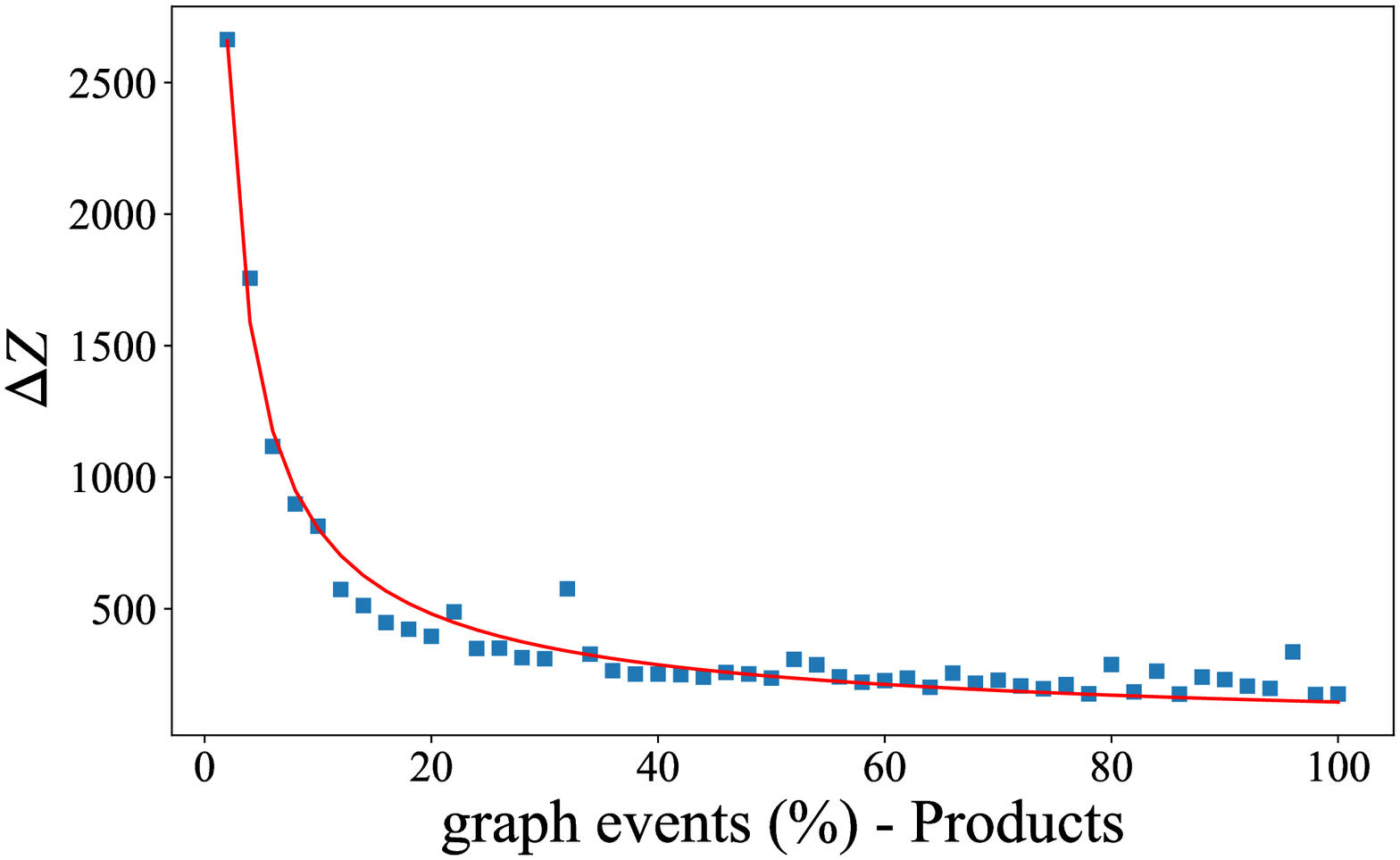}
		\end{tabular}
		\vspace{-2mm}
		\caption{Fit $\Delta \Z$ follows a power-law distribution on various datasets.}
		\label{fig:deltaZ}
		\vspace{-4mm}
	\end{small}
\end{figure}
\begin{figure}[t]
\setlength{\abovecaptionskip}{1mm}
	\begin{small}
		\centering
		\vspace{-1mm}
		\begin{tabular}{cc}
			 \hspace{-4mm} \includegraphics[height=27mm]{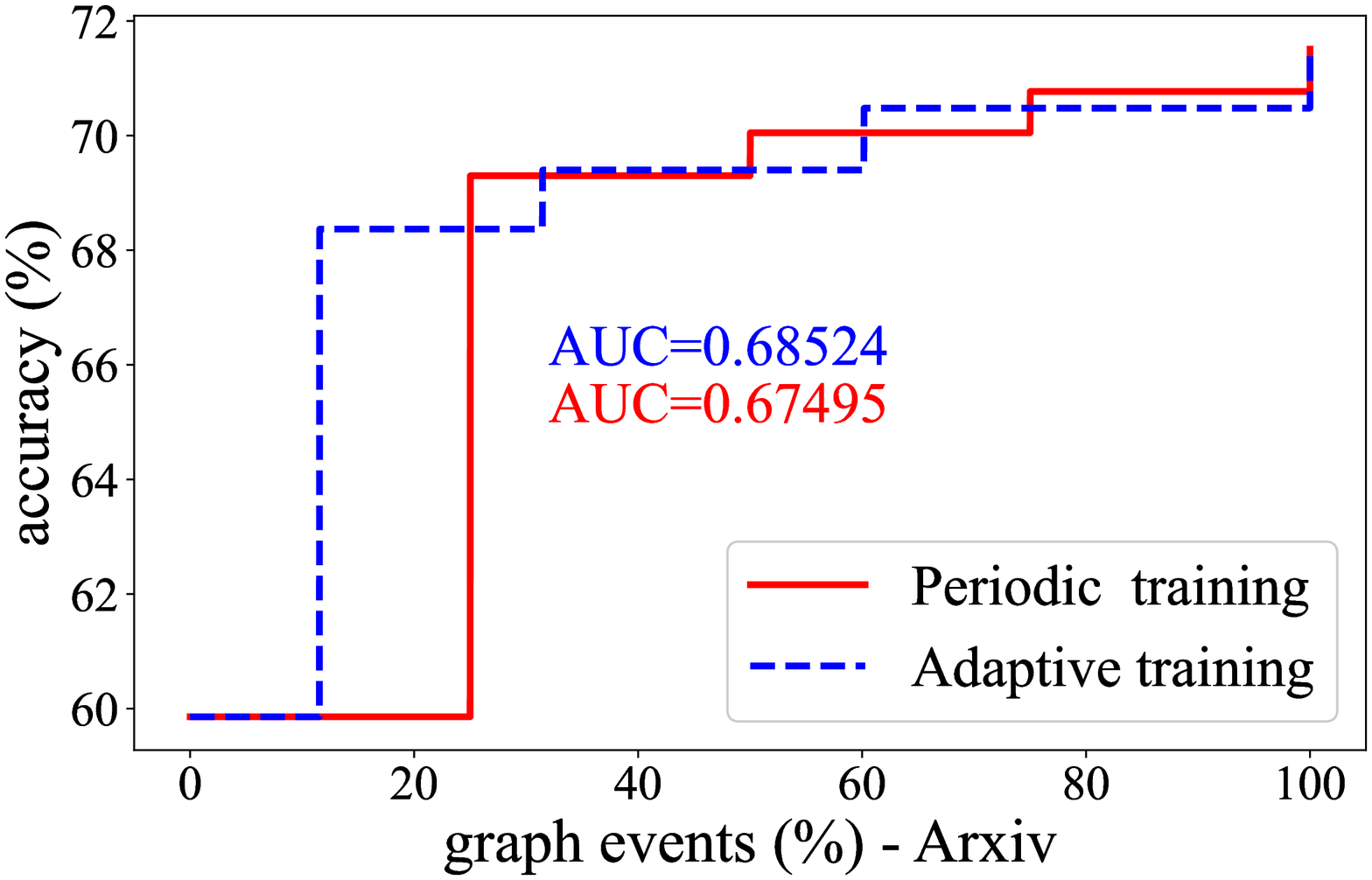} &
			 \hspace{-4mm} \includegraphics[height=27mm]{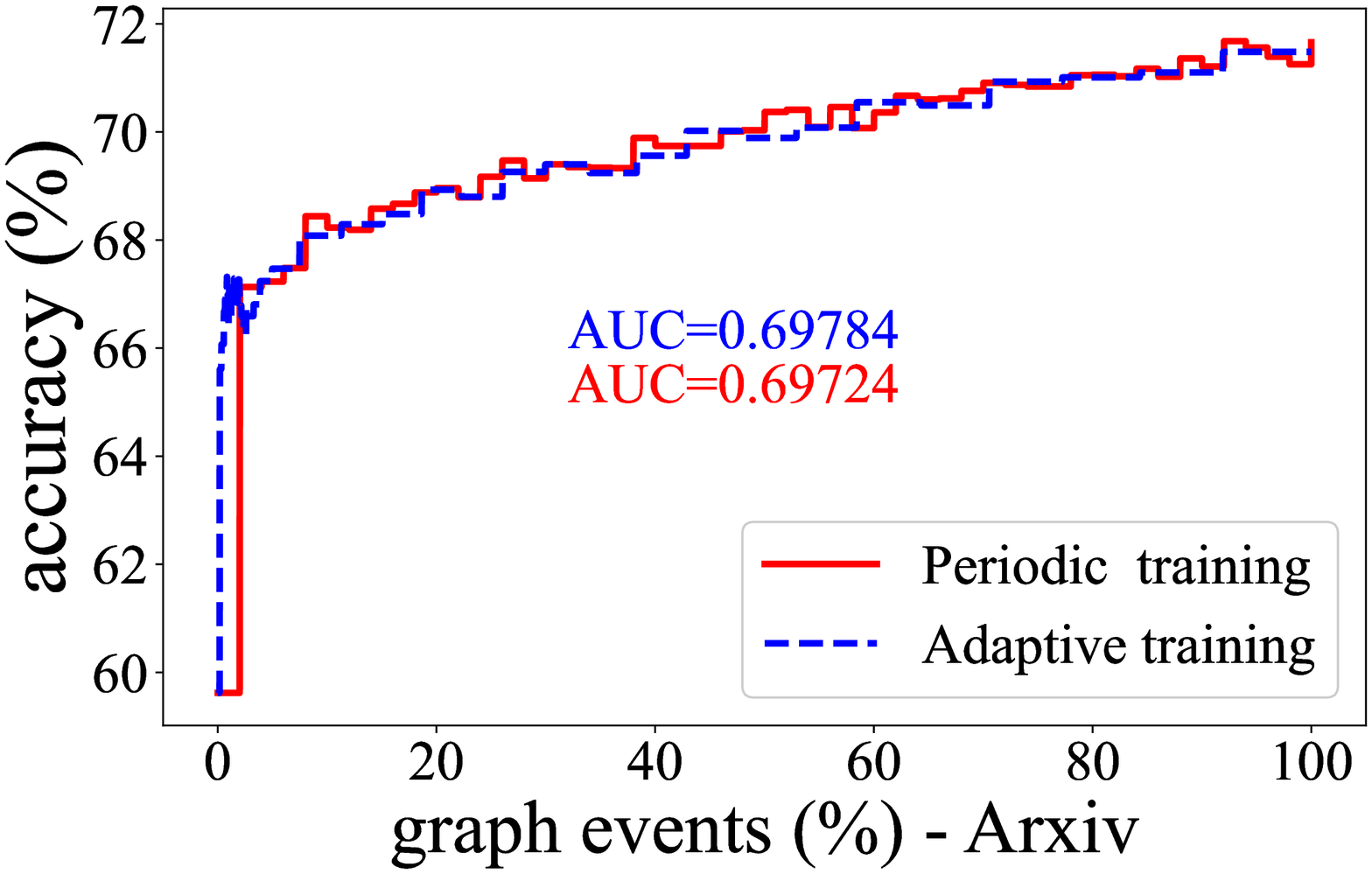} \\
		\end{tabular}
		\vspace{-2mm}
		\caption{Adaptive training on Arxiv setting $|S|\hspace{-1mm}=\hspace{-1mm}4$ and $|S|\hspace{-1mm}=\hspace{-1mm}50$.}
		\label{fig:additional_adaptive_train}
		\vspace{-2mm}
	\end{small}
\end{figure}
Firstly, we perform an empirical evaluation on Arxiv. Setting $|S|=16$, we can retrain the model 16 times, while the model is forced to retrain after $k$-th graph event, where $k$ is the total number of graph events. For \emph{Periodic training}, we retrain the model in a fixed interval, e.g. every $k/|S|$ graph events, and the accuracy curve is shown as the red line in Figure~\ref{fig:adaptive_train} (left). For \emph{Adaptive training}, we found that $\Delta \Z$ of Arxiv follows a power-law curve during the evolution. An intuitive explanation is that the graph becomes dense after a series of edges have been inserted. The changes triggered by continued insertion become smaller and smaller. Products, another insertion only dataset used in Section~\ref{sec:static_label} also shows a similar property, as shown in Figure~\ref{fig:deltaZ}.
We set $\theta\hspace{-0.5mm}=\hspace{-0.5mm}\frac{\Delta \Z}{\left \| \Z_t-\Z_0 \right\|_F }$ to determine the first $r=3$ times to retrain the model, and the remaining $(|S|-r=13)$ retraining time is determined by fitting the power-law curve. \emph{Adaptive training} achieves a better retraining gain based on the qualitative and quantitative results, as shown in Figure~\ref{fig:adaptive_train} (left).
Figure~\ref{fig:additional_adaptive_train} shows the comparison results of setting $|S|=4$ and $|S|=50$, and we can reach similar conclusions. Setting $r=3$, we set $\theta=0.05$ for $|S|=16,50$ and $\theta=0.3$ for $|S|=4$.

We also evaluate \emph{Adaptive training} on SBM-500$K$ with both inserted and deleted edges during the evolution, which better simulates real networks. We consider a special case where several nodes change their communities and recover immediately, and only a fraction of the changes are permanent. Figure~\ref{fig:adaptive_train} (right) shows the accuracy curves of setting $|S|=3$ and $\theta\hspace{-0.5mm}=\hspace{-0.5mm}\Delta \Z\hspace{-0.5mm}=\hspace{-0.5mm}1.73$. 
Note that we set $\theta$ empirically here, \eat{and Figure~\ref{} shows results of different $\theta$.
T}to maintain the fixed budget for a fair comparison, we make the model stop training after reaching the budget constraint.

\vspace{-2mm}
\subsection{Implementation details}
We implement InstantGNN in PyTorch and C++. Each layer is activated using the ReLU non-linear function. Additionally, a dropout function is used behind each layer. Table~\ref{tab:param} summarizes parameters settings of InstantGNN. For baseline methods, we employ their published codes or the codes implemented with Pytorch Geometric Temporal\eat{\cite{rozemberczki2021pytorch}}. All experiments are completed on a machine with an NVIDIA RTX A6000 GPU (48GB memory), Intel Xeon CPU (2.20 GHz) with 40 cores, and 768GB of RAM.

%% file: details_exp_more.tex
\subsection{Baselines}
The baseline methods and parameter settings are as follows. We employ their published codes or the codes implemented with Pytorch Geometric Temporal~\cite{rozemberczki2021pytorch}. 
\begin{itemize}[leftmargin = *]
  \item {\bf GCN}\cite{kipf2016semi}, one of the representative works that extend convolutional neural networks (CNNs) based on grid image data to non-Euclidean space. For Arxiv dataset, we train a full-batch GCN at each snapshot. However, it is difficult to train a full-batch GCN on Products and Papers100M, due to the memory limitation. Therefore, we use GraphSAINT\cite{zeng2019graphsaint} to sample subgraphs and then train GCN on the sampled subgraphs to achieve mini-batch at each snapshot.
  \item {\bf PPRGo}\cite{bojchevski2020pprgo}, a state-of-the-art scalable GNN that adopts PPR matrix to instead of multi-layer convolution to transfer and aggregate node information on the graph. There are three main parameters for feature propagation of PPRGo: $r_{max}$, $L$ and $k$, where $r_{max}$ is the residue threshold, $L$ is the number of hops, and $k$ denotes the number of non-zero PPR value for each training node. We vary $r_{max}$ from $10^{-5}$ to $10^{-7}$, $L$ from 2 to 10, and tune $k$ for the best performance.
  \item {\bf AGP}\cite{wang2021agp}, a general-purpose graph propagation algorithm with near-optimal time complexity. AGP decouples the feature propagation and prediction to achieve high scalability. For feature propagation, AGP has five parameters: Lapalacian parameters $a$ and $b$, weight for $i$-th layer $w_i$, error threshold $\theta$ and number of layers $L$. We set $a=b=1/2$ and $w_i=\alpha(1-\alpha)^i$ and vary parameters ($\theta$, $\alpha$, $L$) from ($10^{-5}$, 0.1, 10) to ($10^{-10}$, 0.5, 20).
  \item {\bf T-GCN}\cite{zhao2019tgcn}, a state-of-the-art temporal GNN method. It consists of gate recurrent unit (GRU) and GCN, which is applied to capture the temporal and structural information, respectively.
  \item {\bf MPNNLSTM}\cite{panagopoulos2021mpnnlstm}, a state-of-the-art temporal GNN method. MPNNLSTM uses long short-term memory (LSTM)\cite{graves2014towards} to capture the temporal information of the graph at different timesteps, and GCN is adopted to learn the graph structural information.
\end{itemize}

\vspace{-3mm}
\subsection{More discussion of results}
There are two likely interpretations for the growing trend of accuracy curves in Figure~\ref{fig:arxiv_acc_time}-\ref{fig:papers100_acc_time}, depending on the type of approach. For InstantGNN and three static methods, each node receives more information from its neighbors as the number of edges increases over time, which is beneficial in classifying. For T-GCN and MPNNLSTM, the prediction result for graph $G_t$ is based on the information learned from $G_1$ to $G_t$. Therefore, the accuracy is low at the first several snapshots for lack of information.

%% file: proof.tex
\section{Proofs} 
\label{sec:appendix}
\noindent
Let's first go over two important properties of PPR and two identical equations.
\begin{proposition}
\label{prop:1}
For any graph, given a source node $s\hspace{-0.5mm}\in\hspace{-0.5mm} V$, let $\bm{\pi}_s(t)$ be the personalized pagerank value from node $s$ to $t$, then $\sum_{t\in V}\bm{\pi}_s(t)=1$.
\end{proposition}
\begin{proposition}
For an undirected graph, let $s$ and $t$ be two nodes of the graph, then $d(s)\bm{\pi}_s(t)=d(t)\bm{\pi}_t(s)$.
\end{proposition}
\begin{invariant}
\label{invariant:1}
Assume $\bm{x}$ is the graph signal vector, $\bm{\hat{\pi}}$ and $\bm{r}$ are the estimate vector and residual vector, respectively, we have $\bm{\hat{\pi}} = (1-\alpha)\P\bm{\hat{\pi}}+\alpha (\bm{x}-\bm{r})$.
\end{invariant}
\begin{invariant}
\label{invariant:2}
Let $I$ be an identity matrix and $\bm{\pi}$ is the exact version of propagation vector, then $\bm{\pi} = \bm{\hat{\pi}} + \alpha \left[ I-(1-\alpha)\P \right]^{-1}\bm{r}$.
\end{invariant}

\subsection{Proof of Lemma~\ref{lemma1} \& Invariant~\ref{invariant:1}}
The propagation progress is set to begin with $\bm{\hat{\pi}}=\bm{0}, \bm{r}=\bm{x}$, as mentioned in Section~\ref{sec:static}. And so we plug this into Equation~\ref{equ:invariant_vector}, we obtain the initial equation as $\alpha \bm{x}=\alpha \bm{x}$. Therefore, Equation~\ref{equ:invariant_vector} is obviously true at the beginning of propagation progress. 
Assuming that the equation holds after the $\ell$-th propagation, e.g. $\bm{\hat{\pi}}^{(\ell)} + \alpha \bm{r}^{(\ell)} = \alpha \bm{x} + (1-\alpha) \P\bm{\hat{\pi}}^{(\ell)}$. Let $y=\bm{r}^{(\ell)}(u)$ be the residual of node $u$ after the $\ell$-th propagation. Following Steps 2-4 in Algorithm~\ref{alg:1}, the $(\ell\hspace{-0.5mm}+\hspace{-0.5mm}1)$-th propagation can be represented as:
\begin{equation*}
\begin{aligned}
\bm{r}^{(\ell+1)} &= \bm{r}^{(\ell)} - y\bm{e}_u + (1-\alpha)\P y\bm{e}_u \\
\bm{\hat{\pi}}^{(\ell+1)} &= \bm{\hat{\pi}}^{(\ell)}+\alpha y\bm{e}_u
\end{aligned}
\end{equation*}
where $\bm{e}_u$ is a one-hot vector with only the $u$-th element being 1 and the others being 0. Therefore, we have
\begin{equation*}
\begin{aligned}
&\alpha \bm{x} + (1\hspace{-0.5mm}-\hspace{-0.5mm}\alpha)\P\bm{\hat{\pi}}^{(\ell+1)}-\alpha\bm{r}^{(\ell+1)} \\
=& \alpha \bm{x} + (1\hspace{-0.5mm}-\hspace{-0.5mm}\alpha)\P(\bm{\hat{\pi}}^{(\ell)} + \alpha y\bm{e}_u) - \alpha \left[ \bm{r}^{(\ell)} - y\bm{e}_u + (1\hspace{-0.5mm}-\hspace{-0.5mm}\alpha)\P y\bm{e}_u \right]\\
=& \alpha \bm{x} + (1\hspace{-0.5mm}-\hspace{-0.5mm}\alpha)\P\bm{\hat{\pi}}^{(\ell)}-\alpha\bm{r}^{(\ell)} + y\bm{e}_u. \\
\end{aligned}
\end{equation*}
We substitute $\alpha \bm{x} + (1\hspace{-0.5mm}-\hspace{-0.5mm}\alpha)\P\bm{\hat{\pi}}^{(\ell)}-\alpha\bm{r}^{(\ell)}$ with $\bm{\hat{\pi}}^{(\ell)}$ and obtain a solution of the above equation as $\bm{\hat{\pi}}^{(\ell+1)}$. 
So Equation~\ref{equ:invariant_vector} also holds at the $(\ell\hspace{-0.5mm}+\hspace{-0.5mm}1)$-th propagation. And the proof is completed. 

\subsection{Proof of Invariant~\ref{invariant:2}}
From the definition and Invariant~\ref{invariant:1}, we have
\begin{equation*}
\left \{
\begin{aligned}
&\bm{\pi} = (1-\alpha) \P\bm{\pi} + \alpha \bm{x} \\
&\bm{\hat{\pi}} = (1-\alpha)\P \bm{\hat{\pi}} + \alpha (\bm{x}-\bm{r})
\end{aligned}
\right .
\end{equation*}
Then we subtract the two equations above:
\begin{equation*}
\begin{aligned}
& \bm{\pi} - \bm{\hat{\pi}} = (1-\alpha) \P (\bm{\pi} - \bm{\hat{\pi}}) + \alpha \bm{r} \\
\Leftrightarrow & \left[ I-(1-\alpha)\P \right] (\bm{\pi} - \bm{\hat{\pi}}) = \alpha \bm{r} \\
\end{aligned}
\end{equation*}
It is known that eigenvalues of $\P$ $\in (-1,1]$. Therefore, eigenvalues of $[ I-(1-\alpha)\P]$ $\in [\alpha, 2-\alpha)$, and $[ I-(1-\alpha)\P]^{-1}$ is meaningful since that $\alpha\in(0,1)$. Then we can express the relationship between $\bm{\pi}$ and $\bm{\hat{\pi}}$ as follows:
\begin{equation*}
\bm{\pi} = \bm{\hat{\pi}} + \alpha \left[ I-(1-\alpha)\P \right]^{-1} \bm{r}.
\end{equation*}

\subsection{Proof of Theorem~\ref{theorem:error}}
Recall that $\P=\D^{-\beta}\A\D^{\beta-1}$ is the propagation matrix. Let $\Q=\alpha [I-(1-\alpha)\P]^{-1}$, we can rewrite it as
\vspace{-1mm}
\begin{equation*}
    \Q=\D^{1-\beta}\cdot \sum\limits_{\ell=0}^{\infty}\alpha (1-\alpha)^\ell \left( \D^{-1}\A \right)^\ell \cdot \D^{\beta-1}.
\end{equation*}
According to the definition, the $(s,t)$-th entity of matrix $\sum_{l=0}^{\infty}\alpha (1\hspace{-0.5mm}-\hspace{-0.5mm}\alpha)^l(D^{-1}A)^\ell $ is the personalized PageRank from $s$ to $t$. Therefore, $(s,t)$-th entity of $\Q$, denoted as $q(s,t)$, is equals to $d(s)^{1-\beta}\cdot \bm{\pi}_s(t)\cdot d(t)^{\beta -1}$.\eat{ppr(s,t)} Let $d_i(s)$ be the degree of node $s$ at time $i$, $\bm{\pi}_{si}(t)$ is the personalized pagerank from $s$ to $t$ at time $i$, and $q_i(s,t)$ is equals to $d_i(s)^{1-\beta}\hspace{-0.5mm}\cdot\hspace{-0.5mm} \bm{\pi}_{si}(t)\cdot d_i(t)^{\beta -1} $. As a result, for each node $t\hspace{-0.5mm}\in\hspace{-0.5mm} V$, we can rewrite Invariant~\ref{invariant:2} using $q(s,t)$ as:
\begin{equation}
\label{equ:invariant2}
    \bm{\pi}(s) = \bm{\hat{\pi}}(s) + \sum_{t\in V}q(s,t)\cdot \bm{r}(t).
\end{equation}

According to line 7 in Algorithm~\ref{alg:2}, we invoke Algorithm~\ref{alg:1} to refresh the estimate vector and residual vector at each time step, after computing the increments caused by $Event_i$. Recall that Algorithm~\ref{alg:1} ends with $\forall s\in V$, $|\bm{r}(s)|\leq\varepsilon \cdot d_i(s)^{1-\beta}$. Therefore, we have
\begin{equation*}
\begin{aligned}
\left|\bm{\hat{\pi}}_{i}(s)-\bm{\pi}_{i}(s)\right| &=\sum_{t \in V} q_{i}(s, t) \cdot\left|\bm{r}_{i}(t)\right| \\
& \leq \sum_{t \in V} d_{i}(s)^{1-\beta} \cdot \bm{\pi}_{s i}(t) \cdot d_{i}(t)^{\beta-1} \cdot\left(\varepsilon \cdot d_{i}(t)^{1-\beta}\right) \\
&=\sum_{t \in V} d_{i}(s)^{1-\beta} \cdot \bm{\pi}_{s i}(t) \cdot \varepsilon \\
&=\varepsilon \cdot d_{i}(s)^{1-\beta}
\end{aligned}
\end{equation*}
In the last equality, we apply the property mentioned in Proposition~\ref{prop:1} that $\sum_{t \in V} \bm{\pi}_{s i}(t)=1$, and the theorem follows.

\subsection{Proof of Theorem~\ref{theorem:1}}
Let $G_0$ be the graph at time $t=0$, $M\hspace{-0.5mm}=\hspace{-0.5mm}\{Event_1,…,Event_k\}$ denotes the event sequence. Recall that $n=|V|$ is the number of nodes, which is assumed to be constant. Let $\bm{\Phi}_i(t)=\sum_{s\in V} d_i(s)^\beta q_i(s,t)$, $\bm{\Phi}_i$ is its vector form, and $\bm{\hat{\pi}}_0$ and $\bm{r}_0$ denote the estimated propagation vector and residual vector for initial graph $G_0$, obtained by running Step 1-2 of Algorithm~\ref{alg:2}. Recall that the estimates of node $s$ increases no more than $\alpha \varepsilon d(s)^{1-\beta}$ at every time node $s\in V$ push. Therefore, the total time for generating $\bm{\hat{\pi}}_0$ and $\bm{r}_0$ is:
\begin{equation*}
\begin{aligned}
& \sum\limits_{s\in V} \frac{\bm{\hat{\pi}}_0(s)\cdot d_0(s)}{\alpha \varepsilon d(s)^{1-\beta}} \\
=&\sum\limits_{s\in V} \frac{d_0(s)^\beta\cdot \left( \bm{\pi}_0(s)-\sum_{t\in V} q_0(s,t)\cdot \bm{r}_0(t) \right)}{\alpha \varepsilon} \\
=& \sum\limits_{s\in V}\frac{ d_0(s)^\beta\cdot \bm{\hat{\pi}}_0(s)}{\alpha \varepsilon} - \sum\limits_{s\in V}\frac{ d_0(s)^\beta\cdot \sum_{t\in V}q_0(s,t)\cdot \bm{r}_0(t) }{\alpha \varepsilon } \\
=& \frac{\lVert \D_0^{\beta}\cdot \bm{\pi}_0 \rVert_1}{\alpha \varepsilon} - \frac{\lVert \bm{\Phi}_0^\top \cdot \bm{r}_0 \rVert_1}{\alpha \varepsilon}
\end{aligned}
\end{equation*}
where we substitute $\bm{\hat{\pi}}_0(s)$ with $\bm{\pi}_0(s)-\sum_{t\in V} q_0(s,t)\cdot \bm{r}_0(t)$ using Invariant~\ref{invariant:2}, to express the time cost in terms of $\bm{\pi}_0$ and $\bm{r}_0$. And we will eliminate $\bm{r}_0$ naturally in the dynamic process.

\subsection{Proof of Theorem~\ref{theorem:2}}
\noindent
Note that the push process is monotonous. Hence we can control the direction of updating estimates by pushing only non-negative residuals or only negative residuals. And we decompose the time cost $T_i = T^+ + T^-$, where $T^+$ only contains time cost for pushing non-negative residuals, $T^-$ only contains time cost for pushing negative residuals. As a result, the whole procedure for getting $\bm{\hat{\pi}}_i$ can be summarized in three steps. First, the exact increment was calculated and updated to $\bm{\hat{\pi}}_{i-1}$, with $\bm{\hat{\pi}}^\prime$ being the result. Second, only non-negative estimations are pushed out, resulting in $\bm{\hat{\pi}}^{\prime\prime}$. Finally, we push negative estimates and eventually reach $\bm{\hat{\pi}}_i$ and maintain the residual vector $\bm{r}_i$ at the same time. Let $\bm{r}^\prime$ and $\bm{r}^{\prime\prime}$ be the intermediate results corresponding to $\bm{\hat{\pi}}^\prime$ and $\bm{\hat{\pi}}^{\prime\prime}$, respectively. According to Equation~\ref{equ:invariant2}, we have
\begin{equation*}
\begin{aligned}
\bm{\hat{\pi}}^\prime(s) &= \bm{\pi}(s) - \sum\limits_{t\in V}q(s,t)\cdot \bm{r}^\prime(t), \\
\bm{\hat{\pi}}^{\prime\prime}(s) &= \bm{\pi}(s) - \sum\limits_{t\in V}q(s,t)\cdot \bm{r}^{\prime\prime}(t).
\end{aligned}
\end{equation*}
We start with $T^+$. It is easy to observed that $\bm{\hat{\pi}}^{\prime\prime}(s) \geq \bm{\hat{\pi}}^\prime(s)$, since only non-negative residuals are pushed out. Then we have:
\begin{equation*}
\bm{\hat{\pi}}^{\prime\prime}(s) - \bm{\hat{\pi}}^\prime(s) = \sum\limits_{t\in V}q(s,t)\cdot (\bm{r}^{\prime}(t)- \bm{r}^{\prime\prime}(t)).
\end{equation*}
As described in \cite{zhang2016approximate}, the difference between the amount of residuals that has been pushed out and the one that are new generated is actually equal to the amount of mass that has been received into the estimates. Then the time cost for pushing non-negative estimates is
\begin{equation*}
\begin{aligned}
T^+ &= \sum\limits_{t\in V} \frac{d_i(s)\cdot (\bm{\hat{\pi}}^{\prime\prime}(s) - \bm{\hat{\pi}}^\prime(s))}{\alpha \varepsilon d(u)^{1-\beta}} \\
&= \sum\limits_{t\in V} \frac{ d_i(s)^\beta \cdot \sum\limits_{t\in V}q_i(s,t) (\bm{r}^\prime(t)- \bm{r}^{\prime\prime}(t))}{\alpha \varepsilon } \\
&= \sum\limits_{t\in V} \frac{\Phi_i(t)\cdot (\bm{r}^\prime(t)- \bm{r}^{\prime\prime}(t))}{\alpha \varepsilon } \\
&\leq \frac{\lVert \bm{\Phi}_i^\top \cdot \bm{r}^\prime \rVert_1}{\alpha \varepsilon } - \frac{\lVert \bm{\Phi}_i^\top \cdot \bm{r}^{\prime\prime} \rVert_1}{\alpha \varepsilon }
\end{aligned}
\end{equation*}

Since that only negative residuals are pushed out in the final step, we have $\bm{\hat{\pi}}^{\prime\prime}(s) \geq \bm{\hat{\pi}}_i(s)$. Then,
\begin{equation*}
\begin{aligned}
T^- &= \sum\limits_{t\in V} \frac{d_i(s)\cdot (\bm{\hat{\pi}}^{\prime\prime}(s) - \bm{\hat{\pi}}_i(s))}{\alpha \varepsilon d(u)^{1-\beta}} \\
&= \sum\limits_{t\in V} \frac{\Phi_i(t)\cdot( \bm{r}_i(t)- \bm{r}^{\prime\prime}(t))}{\alpha \varepsilon } \\ 
&\leq \frac{\lVert \bm{\Phi}_i^\top \cdot \bm{r}^{\prime\prime} \rVert_1}{\alpha \varepsilon } - \frac{\lVert \bm{\Phi}_i^\top \cdot \bm{r}_i \rVert_1}{\alpha \varepsilon }
\end{aligned}
\end{equation*}

Therefore, the total time for updating $\bm{\hat{\pi}}_{i-1}(t)$ and $\bm{r}_{i-1}(t)$ to $\bm{\hat{\pi}}_i(t)$ and $\bm{r}_i(t)$ is at most
\begin{equation*}
\vspace{-2mm}
T_i=T^+ + T^- \hspace{-1mm}\leq \frac{\lVert \bm{\Phi}_i^\top \cdot \bm{r}^\prime \rVert_1}{\alpha \varepsilon } - \frac{\lVert \bm{\Phi}_i^\top \cdot \bm{r}_i \rVert_1}{\alpha \varepsilon }.
\end{equation*}

Recall that $\bm{r}^\prime$ is the residual vector that obtained by updating the increment. Specifically, $\bm{r}^\prime=\bm{r}_{i-1}+\Delta\bm{r}_i$, where $\Delta\bm{r}_i $ is a sparse vector and we will consider it later. \eat{that only has values in the $u$, $v$, $w\in N_{i-1}(u)$ and $y\in N_{i-1}$ locations. } Since $\bm{\Phi}_i (t) \geq 0$ for each node $t\in V$, we have
\begin{equation*}
\lVert \bm{\Phi}_i^\top \cdot \bm{r}^\prime \rVert_1 \leq \lVert \bm{\Phi}_i^\top \cdot \bm{r}_{i-1} \rVert_1 + \lVert \bm{\Phi}_i^\top \cdot \bm{\Delta r}_i \rVert_1.
\end{equation*}
where
\begin{equation*}
\lVert \bm{\Phi}_i^\top \cdot \bm{r}_{i-1} \rVert_1 = \lVert (\bm{\Phi}_i^\top - \bm{\Phi}_{i-1}^\top) \cdot \bm{r}_{i-1} \rVert_1 + \lVert \bm{\Phi}_{i-1}^\top \cdot \bm{r}_{i-1} \rVert_1.
\end{equation*}
By the definition of $\bm{r}$ and the guarantee that residuals before have met the given error tolerance, for any node $t\in V$, we have $\bm{r}_{i-1}(t) \leq \varepsilon d_{i-1}(t)^{1-\beta} $. Then we have 
\begin{equation*}
\begin{aligned}
&\lVert (\bm{\Phi}_i^\top - \bm{\Phi}_{i-1}^\top) \cdot \bm{r}_{i-1} \rVert_1 \\
\leq& \sum\limits_{t\in V} \varepsilon d_{i-1}(t)^{1-\beta} \cdot [\bm{\Phi}_i(t) - \bm{\Phi}_{i-1}(t)] \\ 
=& \varepsilon\cdot \sum\limits_{t\in V} d_i(t)^{1-\beta} \sum\limits_{s\in V} d_i(s)^{\beta} d_i(s)^{1-\beta} \bm{\pi}_{si}(t) d_i(t)^{\beta-1} - \\
 \indent&\varepsilon \sum\limits_{t\in V} d_{i-1}(t)^{1-\beta} \sum\limits_{s\in V} d_{i-1}(s)^{\beta} d_{i-1}(s)^{1-\beta} \bm{\pi}_{s(i-1)}(t) d_{i-1}(t)^{\beta-1} \\
=& \varepsilon\cdot \sum\limits_{t\in V} \sum\limits_{s\in V} d_i(s)\bm{\pi}_{si}(t) - \varepsilon\cdot \sum\limits_{t\in V} \sum\limits_{s\in V} d_{i-1}(s)\bm{\pi}_{s(i-1)}(t) \\
\end{aligned}
\end{equation*}
Then, we reverse the order of the summing symbols and substitute Proposition~\ref{prop:1} to simplify $\sum_{t\in V}\bm{\pi}_{si}(t)$ and $\sum_{t\in V}\bm{\pi}_{s(i-1)}(t)$ to 1, providing the solution to the preceding equation for $\varepsilon\cdot \sum_{s\in V} (d_i(s) - d_{i-1}(s))$. Note that only the degrees of node $u$ and $v$ have changed after inserting the edge $e_i=(u,v)$. Therefore, we have 
\begin{equation*}
\left \{
\begin{aligned}
& d_i(s) - d_{i-1}(s) = 1 , \text{if } s=u \text{ or } v \\
& d_i(s) - d_{i-1}(s) = 0 , \text{otherwise}
\end{aligned}
\right.
\end{equation*}
Therefore, $\sum_{s\in V} (d_i(s) - d_{i-1}(s)) = 2$, and we have
\begin{equation*}
\lVert (\bm{\Phi}_i^\top - \bm{\Phi}_{i-1}^\top) \cdot \bm{r}_{i-1} \rVert_1 \leq 2\varepsilon.
\end{equation*}

Based on above analysis, we get an upper bound of $T_i$:
\begin{equation*}
\begin{aligned}
T_i \leq& \frac{\lVert \bm{\Phi}_i^\top \cdot \bm{r}^\prime \rVert_1}{\alpha \varepsilon } - \frac{\lVert \bm{\Phi}_i^\top \cdot \bm{r}_i \rVert_1}{\alpha \varepsilon } \\
\leq& \frac{\lVert (\bm{\Phi}_i^\top - \bm{\Phi}_{i-1}^\top) \cdot \bm{r}_{i-1} \rVert_1}{\alpha \varepsilon } + \frac{\lVert \bm{\Phi}_{i-1}^\top \cdot \bm{r}_{i-1} \rVert_1}{\alpha \varepsilon } +\\ &\frac{\lVert \bm{\Phi}_i^\top \cdot \bm{\Delta r}_i \rVert_1}{\alpha \varepsilon } - \frac{\lVert \bm{\Phi}_i^\top \cdot \bm{r}_i \rVert_1}{\alpha \varepsilon } \\
\leq& \frac{2}{\alpha}+\frac{\lVert \bm{\Phi}_i^\top \cdot \Delta\bm{r}_i \rVert_1}{\alpha\varepsilon}+\frac{\lVert \bm{\Phi}_{i-1}^\top \cdot \bm{r}_{i-1} \rVert_1}{\alpha\varepsilon } - \frac{\lVert \bm{\Phi}_{i}^\top \cdot \bm{r}_{i} \rVert_1}{\alpha\varepsilon }.
\end{aligned}
\end{equation*}

\subsection{Proof of Theorem~\ref{theorem:3}}
\noindent
Now, we consider the expected time complexity of Algorithm~\ref{alg:2}, which is calculated as the sum of the Initialization and k update times. Combining Theorem~\ref{theorem:1} and \ref{theorem:2}, we write the total time $T$ in the form as:
\begin{equation*}
\begin{aligned}
T=& T_{init}+\sum\limits_{i=1}^k T_i\\
\leq& \frac{\lVert \D_0^{\beta}\cdot \bm{\pi}_0 \rVert_1}{\alpha \varepsilon} - \frac{\lVert \bm{\Phi}_0^\top \cdot \bm{r}_0 \rVert_1}{\alpha \varepsilon} + \frac{2k}{\alpha} + \sum\limits_{i=1}^k \frac{\lVert \bm{\Phi}_i^\top \cdot \bm{\Delta r}_i \rVert_1}{\alpha\varepsilon } \\
&+ \sum\limits_{i=1}^k \left( \frac{\lVert \bm{\Phi}_{i-1}^\top \cdot \bm{r}_{i-1} \rVert_1}{\alpha\varepsilon } - \frac{\lVert \bm{\Phi}_{i}^\top \cdot \bm{r}_{i} \rVert_1}{\alpha\varepsilon } \right) \\
\leq& \frac{\lVert \D_0^{\beta}\cdot \bm{\pi}_0 \rVert_1}{\alpha \varepsilon} + \frac{2k}{
\alpha} + \sum\limits_{i=1}^k \frac{\lVert \bm{\Phi}_i^\top \cdot \Delta\bm{r}_i \rVert_1}{\alpha\varepsilon }.
\end{aligned}
\end{equation*}
where each term of $\Delta\bm{r}_i$ is the precise increment of this node that induced by inserting the edge $e_i=(u,v)$. As mentioned in Section~\ref{sec:instant2structural}, in $\Delta\bm{r}_i$, only the locations of affected nodes have meaning (e.g, $u$, $v$, $w\in N_{i-1}(u)$ and $y\in N_{i-1}$), and we have
\begin{equation}
\label{equ:delta_r}
\begin{aligned}
\lVert \bm{\Phi}_i^\top \cdot \Delta\bm{r}_i \rVert_1 &= |\bm{\Phi}_i(u)\Delta\bm{r}_i(u)| + |\bm{\Phi}_i(v)\Delta\bm{r}_i(v)| \\
&+ \sum\limits_{w\in N_{i-1}(u)}\hspace{-4mm}|\bm{\Phi}_i(w)\Delta\bm{r}_i(w)| + \sum\limits_{y\in N_{i}(v)}\hspace{-3mm}|\bm{\Phi}_i(y)\Delta\bm{r}_i(y)|.
\end{aligned}
\end{equation}

Next, we will provide upper boundaries for each of these four items to finish the proof. We start with $\bm{\Phi}_i(u)\Delta\bm{r}_i(u)$, using Equation~\ref{equ:delta_r(u)}, we have
\begin{equation*}
\begin{aligned}
\bm{\Phi}_i(u)\Delta\bm{r}_i(u) &= \bm{\Phi}_i(u) \cdot \frac{1-\alpha}{\alpha} \cdot \frac{\bm{\hat{\pi}}_{i-1}(v) }{d_{i}(u)^{\beta}d_{i-1}(v)^{1-\beta}} + \\
&\bm{\Phi}_i(u) \cdot \frac{\bm{\hat{\pi}}_{i-1}(u) + \alpha\bm{r}_{i-1}(u) - \alpha\bm{x}(u)}{\alpha}\cdot \frac{ d_{i-1}(u)^{\beta}- d_{i}(u)^{\beta}}{d_{i}(u)^{\beta}}.
\end{aligned}
\end{equation*}

Recalling the definition, we have $\bm{\Phi}_i(u)=\sum_{s\in V}d_i(s)^\beta q_i(s,u)$. 
We substitute $\bm{\hat{\pi}}_{i-1}(v)$ with $(\bm{\pi}_{i-1}(v) - \sum_{t\in V}q(v,t)\cdot \bm{r}(t))$, using Equation~\ref{equ:invariant2}, and obtain a bound for the first term of the above equation as 
$\frac{1-\alpha}{\alpha}[\bm{\pi}_{i-1}(v)+\varepsilon]$. Assuming that each element of the graph signal $\bm{x}$ is non-negative, we have
\begin{equation*}
\begin{aligned}
&\bm{\Phi}_i(u) \cdot \frac{\bm{\hat{\pi}}_{i-1}(u) + \alpha\bm{r}_{i-1}(u) - \alpha\bm{x}(u)}{\alpha}\cdot \frac{ d_{i-1}(u)^{\beta}- d_{i}(u)^{\beta}}{d_{i}(u)^{\beta}} \\
\leq& \bm{\Phi}_i(u) \cdot \frac{\bm{\hat{\pi}}_{i-1}(u) + \alpha\bm{r}_{i-1}(u)}{\alpha}\cdot \frac{ d_{i-1}(u)^{\beta}- d_{i}(u)^{\beta}}{d_{i}(u)^{\beta}} \\
\end{aligned}
\end{equation*}
Using the fact that $\frac{ d_{i-1}(u)^{\beta}- d_{i}(u)^{\beta}}{d_{i}(u)^{\beta}} \leq \frac{1}{d_i(u)}\leq\frac{1}{d_i(u)^\beta}$ and omitting some minor details, we get an upper bound as $\frac{1}{\alpha}(\bm{\pi}_{i-1}(u)+\varepsilon) + \varepsilon$. Therefore, $|\bm{\Phi}_i(u)\Delta\bm{r}_i(u)|$ is at most $\frac{1-\alpha}{\alpha}[\bm{\pi}_{i-1}(v)+\varepsilon]+\frac{1}{\alpha}(\bm{\pi}_{i-1}(u)+\varepsilon) + \varepsilon$. We deal with $|\bm{\Phi}_i(v)\Delta\bm{r}_i(v)|$ in a similar way and obtain an upper bound as $\frac{1-\alpha}{\alpha}[\bm{\pi}_{i-1}(u)+\varepsilon]+\frac{1}{\alpha}(\bm{\pi}_{i-1}(v)+\varepsilon) + \varepsilon$.

Now we handle the third term of Equation~\ref{equ:delta_r}. For each node $w\in N_{i-1}(u)$ we have
\begin{equation*}
\begin{aligned}
|\bm{\Phi}_i(w)\Delta\bm{r}_i(w)|
=&\left| \bm{\Phi}_i(w)\cdot\frac{1-\alpha}{\alpha}\cdot\frac{\bm{\hat{\pi}}_{i-1}(u)}{d_i(w)^\beta}\cdot\frac{d_{i-1}(u)^{1-\beta}-d_i(u)^{1-\beta}}{d_{i-1}(u)^{1-\beta}d_i(u)^{1-\beta}}\right| \\
=& \frac{1-\alpha}{\alpha}\cdot\frac{1}{d_{i-1}(u)^{1-\beta}d_i(u)^{1-\beta}}\cdot |\bm{\hat{\pi}}_{i-1}(u)| \\
\leq&\frac{1-\alpha}{\alpha \cdot d_i(u)}\cdot|\bm{\pi}_{i-1}(u)+\varepsilon|.
\end{aligned}
\end{equation*}
It is easy to see that 
\begin{equation*}
\begin{aligned}
\sum\limits_{w\in N_{i-1}(u)}\hspace{-4mm} |\bm{\Phi}_i(w) \Delta\bm{r}_i(w)|\leq& d_{i-1}(u)\cdot\frac{1-\alpha}{\alpha \cdot d_i(u)}\cdot|\bm{\pi}_{i-1}(u)+\varepsilon| \\
\leq& \frac{1-\alpha}{\alpha}\cdot|\bm{\pi}_{i-1}(u)+\varepsilon|.
\end{aligned}
\end{equation*}

Omitting all the minor details, we obtain an bound as $\frac{1-\alpha}{\alpha}\cdot|\bm{\pi}_{i-1}(v)+\varepsilon|$ for the fourth item of Equation~\ref{equ:delta_r}. Based on the above analysis, we have
\begin{equation*}
\lVert \bm{\Phi}_i^\top \cdot \Delta\bm{r}_i \rVert_1 \leq \frac{(3-2\alpha)\bm{\pi}_{i-1}(u)+(3-2\alpha)\bm{\pi}_{i-1}(v)+6\varepsilon-2\alpha\varepsilon}{\alpha}.
\end{equation*}

We assume that edges are arrived randomly, such that each node has an equal probability of being the endpoint of a new edge. Therefore, we have
\begin{equation*}
E[\lVert \bm{\Phi}_i^\top \cdot \Delta\bm{r}_i \rVert_1]=\frac{(6-4\alpha) \lVert \bm{\pi}_{i-1}\rVert_1}{\alpha n}+\frac{6\varepsilon}{\alpha}-2\varepsilon.
\end{equation*}
Finally, we get the upper bound of $T$ as
\begin{equation*}
\begin{aligned}
E[T]\leq& \frac{\lVert \D_0^{\beta}\cdot \bm{\pi}_0 \rVert_1}{\alpha \varepsilon} + \frac{2k}{
\alpha} + \sum\limits_{i=1}^k \frac{E[\lVert \bm{\Phi}_i^\top \cdot \Delta\bm{r}_i \rVert_1]}{\alpha\varepsilon }\\
=& \frac{\lVert \D_0^{\beta}\cdot \bm{\pi}_0 \rVert_1}{\alpha \varepsilon} + \frac{6k}{\alpha^2} + \sum\limits_{i=1}^k \frac{(6-4\alpha) \lVert \bm{\pi}_{i-1} \rVert_1}{\alpha^2 n\varepsilon}.
\end{aligned}
\end{equation*}